\pdfoutput=1

\documentclass[11pt]{article}

\usepackage{acl}

\usepackage{times}
\usepackage{latexsym}
\usepackage{graphicx}
\usepackage{amsmath}
\usepackage{multirow}
\usepackage{hyperref}
\usepackage{xcolor} 
\usepackage{booktabs}
\usepackage{algorithm}
\usepackage{algpseudocode}
\usepackage{caption}

\usepackage[T1]{fontenc}

\usepackage[utf8]{inputenc}

\usepackage{microtype}

\usepackage{inconsolata}

\usepackage{graphicx}

\title{Addressing Blind Guessing: Calibration of Selection Bias in Multiple-Choice Question Answering by Video Language Models}

\author{Olga Loginova \\
  University of Trento \\
  \texttt{olga.loginova@unitn.it}\\\And
  Oleksandr Bezrukov \\
  Gran Sasso Science Institute \\
  \texttt{oleksandr.bezrukov@gssi.it}\\\AND
  Ravi Shekhar \\
  University of Essex \\
  \texttt{r.shekhar@essex.ac.uk}\\\And
  Alexey Kravets \\
  University of Bath \\
  \texttt{ak3095@bath.ac.uk} \\
  }

\begin{document}
\maketitle
\begin{abstract}
Evaluating Video Language Models (VLMs) is a challenging task. Due to its transparency, Multiple-Choice Question Answering (MCQA) is widely used to measure the performance of these models through accuracy. However, existing MCQA benchmarks fail to capture the full reasoning capabilities of VLMs due to \textit{selection bias}, when models disproportionately favor certain answer options based on positional patterns observed during training. In this work, we conduct a comprehensive empirical analysis of several VLM architectures across major datasets designed to assess complex video-focused reasoning. We identify where the bias is most pronounced and demonstrate to what extent model responses reflect genuine understanding of video content and related questions, as opposed to reliance on arbitrary patterns or superficial cues, such as answer position. By decomposing the MCQA task and adapting fairness bias metrics to VLMs, we introduce a post-processing calibration technique \textbf{BOLD} to balance this bias. Our results show that reducing selection bias improves not only debiasing metrics but also overall model performance, including Accuracy and F1 Mean score. Our method, by suppressing "blind guessing", offers a more cost- and time-effective approach to mitigating selection bias compared to existing techniques. This study represents the first focused investigation of selection bias in video-to-text LLM-powered models.\footnote{The codes for the settings, models and calibration technique are avaialable on Github: \href{URL}{https://github.com/ologin/BOLD}}
\end{abstract}

\section{Introduction}
\label{sec:intro}

Multiple-choice question answering (MCQA) is a transparent and convenient method to assess language models' performance. Each MCQA instance presents a question and several answer options, requiring the selection of the correct one. However, large language models (LLMs) often exhibit \textit{selection bias}, i.e., a tendency to favor certain answer positions irrespective of content \cite{zheng2024large, Wang2024BeyondTA}. Such bias compromises fairness, reliability, and the true reasoning capabilities of these models. Although selection bias in LLMs has received increasing attention, it remains largely unexplored for video-language models (VLMs).

VLMs integrate textual and visual inputs by processing sequences of frames and capturing both spatial and temporal features. Recent advances align pre-trained LLMs with video encoders \cite{Bordes2024AnIT}. Although VLMs inherit many properties from LLMs, their reliance on complex visual inputs introduces unique challenges, such as higher computational overhead and the need for robust spatiotemporal reasoning. The growing role of VLM-based MCQA tasks thus calls for a tailored investigation of selection bias in VLMs.

We address this gap with the first study of video-MCQA selection bias, focusing on video-specific aspects such as dynamic unfolding of events, spatial object tracking, and camera behavior, rather than general plot understanding.

\begin{figure}[ht]
    \centering
    \includegraphics[width=0.48\textwidth]{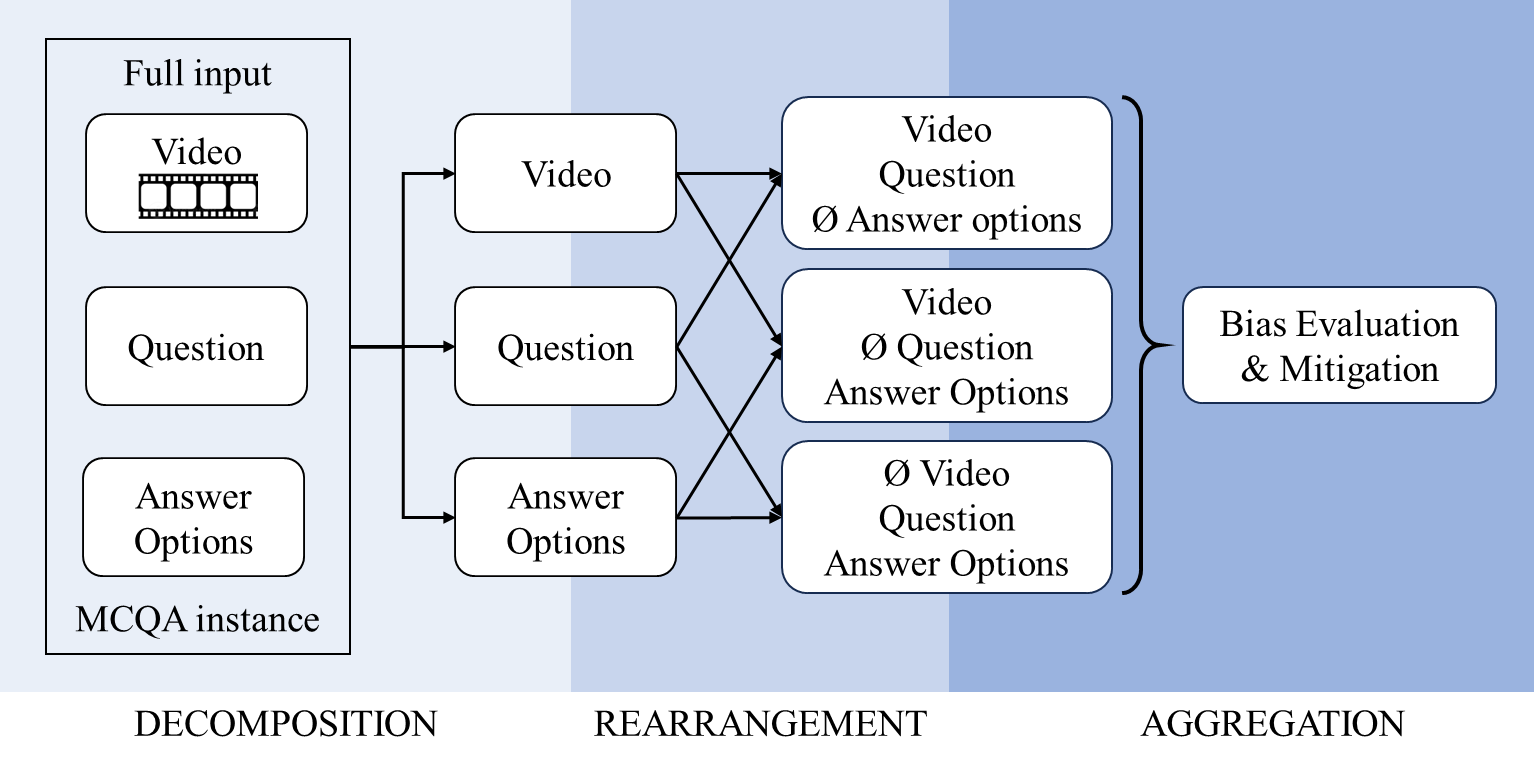}
    \caption{Decomposition approach in 3 steps: key-component decomposition, rearrangement into pairs, applying probability debiasing to the aggregated data.}
    \label{fig:decomposition}
\end{figure}

Through the empirical analysis, we locate where the model’s predictions are skewed by inherent option preferences. We find that, while each model demonstrates specific biases, accuracy-relevant settings yield option distributions similar to those in the default setting. This suggests that the models rely heavily on answer position, regardless of content. Conversely, in accuracy-irrelevant settings, the predicted distributions deviate significantly from uniformity, and this is where the bias is most apparent. Subsequently, we adapt the metrics from fairness bias studies (e.g., regarding race or age) to video-MCQA and deliver a calibration technique.

Figure~\ref{fig:decomposition} illustrates our \textit{decomposition} strategy. We find that depriving the MCQA task of its key constituent components (video, question, answer options) reveals latent bias factors. We systematically isolate the three components: we remove one component at a time and rearrange the remaining pairs, obtaining three “ill-defined” variants of the original data. These decompositions allow us to identify the underlying bias patterns across different sub-spaces of the task. We then aggregate these insights and apply a probability-based debiasing procedure that adjusts the final predictions.

By experimenting with three VLM architectures and four video-specific MCQA datasets, we show that our post-training practical solution enhances interpretability, reduces computational complexity\footnote{Traditionally, selection bias is mitigated through shuffling \cite{zheng2024large, Liu2023MMBenchIY}, where answer options are permuted multiple times to produce more impartial results. Our method requires far fewer resources compared to all the exhaustive answer-option permutations.}, and improves bias metrics. Intriguingly, while our method is designed to mitigate bias, it generally improves performance with better accuracy and F1 Mean scores.

In summary, our contributions are:

\begin{itemize}
    \item A thorough investigation of selection bias in VLM-based MCQA with 11 dataset modifications to target video-specific reasoning challenges.
    \item Adaptation of fairness metrics to assess selection bias in VLMs on video MCQA benchmarks.
    \item A novel decomposition post-processing calibration technique that leverages the rearrangement of MCQA, mitigates selection bias and improves performance.
\end{itemize}

Through the refined MCQA benchmarks, we uncover deeper reasoning issues in contemporary VLMs. Our work helps move toward more robust, fair, reliable and interpretable VLMs by bridging a gap in the study of selection bias in video understanding models and offering a cost-effective solution. 
\section{Related Works}
\label{sec:relWord}

Selection bias in LLMs has been extensively studied~\cite{Pezeshkpour2023LargeLM, Zheng2023LargeLM, Wang2024BeyondTA, Li2024CanMQ, Liusie2024TeacherStudentTF, Wang2024MyAI}. Fewer works are dedicated to image-to-text vision LLMs \cite{zong2024fool, Zhang2024DebiasingML, xue-etal-2024-strengthened}. 

\citet{zheng2024large} conducted an empirical analysis of both position and token biases and proposed PriDe, a debiasing method for LLMs that separates prior bias from unbiased results using shuffling.

We also treat the selection bias as subtraction from observed results in order to obtain the debiased outputs, but we move away from shuffling and regard bias as a vector projected onto three planes. Thus, our calibration method aligns more closely with \citet{Zhang2024DebiasingML}, who cut off the visual input 
for
the image-to-text MCQA. Differently, we operate on video-to-text, and, going further, decompose the bias vector into several projections rather than one. This way we eliminate the unimodal bias highlighted by \citet{Zhang2023UnderstandingUB}.

\citet{Wang2024BeyondTA} and \citet{Liu2023MMBenchIY} resolve the selection bias by augmenting datasets with re-ordered and additional options, increasing number of question-answer pairs in the dataset. We, in contrast, manipulate answer options, as well as videos and questions, not as a solution but as a means of analysis: we identify where the bias is most pronounced and use these modifications as a shortcut for debiasing. 

Lastly, \citet{Pezeshkpour2023LargeLM} propose calibrating LLM predictions by redistributing probabilities to empirically determined options based on the model's uncertainty. However, we find this approach insufficiently robust, as, according to our analysis, the bias depends on both the model and the dataset, and there is no consistent correlation between the model confidence and the presence of bias.
\section{Empirical Analysis}
\label{sec:expSetup}

\subsection{Datasets}
\label{sec:data}

\begin{table}[ht]
  \centering
  \caption{Summary of Question-Answer (QA) pairs, Videos, and Number of Options for each dataset.}
  \label{tab:data}
  \begin{tabular}{lccc}
    \hline
    \textbf{Dataset}      & \textbf{QA pairs} & \textbf{Videos} & \textbf{Options} \\
    \hline
    NExT-QA               & 8564              & 1000            & 5                \\
    NExT-GQA              & 4962              & 971             & 5                \\
    STAR                  & 7098              & 914             & 4                \\
    Video-MME             & 2700              & 900             & 4                \\
    Perception Test       & 7656              & 3926            & 3                \\
    \hline
  \end{tabular}
\end{table}

Our goal is to investigate VLMs' bais under different conditions. To this end, we chose datasets specifically to cover a wide variety of question types, including \textit{what}, \textit{when}, \textit{why}, \textit{how}, and \textit{yes/no} questions, that represent a broad spectrum of video-specific reasoning categories such as causal, temporal, descriptive, situational reasoning, as well as memory, abstraction, physics, and semantics.

\begin{table*}[ht]
  \centering
  \caption{Inference Accuracy (Acc) and F1 Mean Score by Models and Datasets.}
  \label{tab:inf_accuracy}
  \begin{tabular}{lcccccccc}
    \hline
    \textbf{Model}      & \multicolumn{2}{c}{\textbf{NExT-QA}} & \multicolumn{2}{c}{\textbf{STAR}} & \multicolumn{2}{c}{\textbf{Perception Test}} & \multicolumn{2}{c}{\textbf{Video-MME}} \\
                        & \textbf{Acc} & \textbf{F1 Mean} & \textbf{Acc} & \textbf{F1 Mean} & \textbf{Acc} & \textbf{F1 Mean} & \textbf{Acc} & \textbf{F1 Mean} \\
    \hline
    Video-LLaMA          & 40.85   & 40.85   & 36.59   & 31.86   & 41.59   & 37.19   & 32.67   & 28.15   \\
    Video-LLaVA          & 49.96   & 49.81   & 34.71   & 31.83   & 40.73   & 35.69   & 34.22   & 30.99   \\
    SeViLA               & 63.78   & 63.88   & 46.28   & 46.14   & 45.30   & 45.11   & 39.85   & 39.82   \\
    \hline
  \end{tabular}
\end{table*}

We selected the following four video MCQA datasets: \textbf{NExT-QA} \cite{Xiao2021NExTQANP} (and its subset \textbf{NExT-GQA} \cite{Xiao2023CanIT} with timestamp annotations to locate the answer moment), \textbf{STAR} \cite{Wu2021STARAB}, \textbf{Perception Test} \cite{Puatruaucean2023PerceptionTA}, and \textbf{Video-MME} \cite{fu2024video}. These datasets consist of videos of varying lengths, from a few seconds to over an hour. The questions are either manually crafted or generated via functional programs, and the video content range from short everyday activities to animations.

The datasets also differ in the number of answer options per question. Table~\ref{tab:data} gives the overview of the number of QA pairs, videos, and answer options for each dataset. Further details on each dataset, including QA examples, are provided in Appendix~\ref{sec:dat}.

For experimenting, we used the test splits of NExT-QA/NExT-GQA and Video-MME and the publicly available validation splits with correct answer annotations of STAR and Perception Test.

\subsection{Models}
\label{sec:model}

To explore selection bias in a broader context, we chose to examine it with respect to diverse model architectures. Despite the differences, the selected models share key characteristics: they sample frames to extract visual features, either by keyframe selection or uniform sampling, and align visual features with textual representations using pre-trained LLMs. Each model leverages pre-trained visual encoders and fine-tunes pre-trained components with video-text datasets. The models differ substantially in methods of integrating video and language modalities \cite{Bordes2024AnIT}:

\begin{figure*}[t]
\centering
  \includegraphics[width=\textwidth]{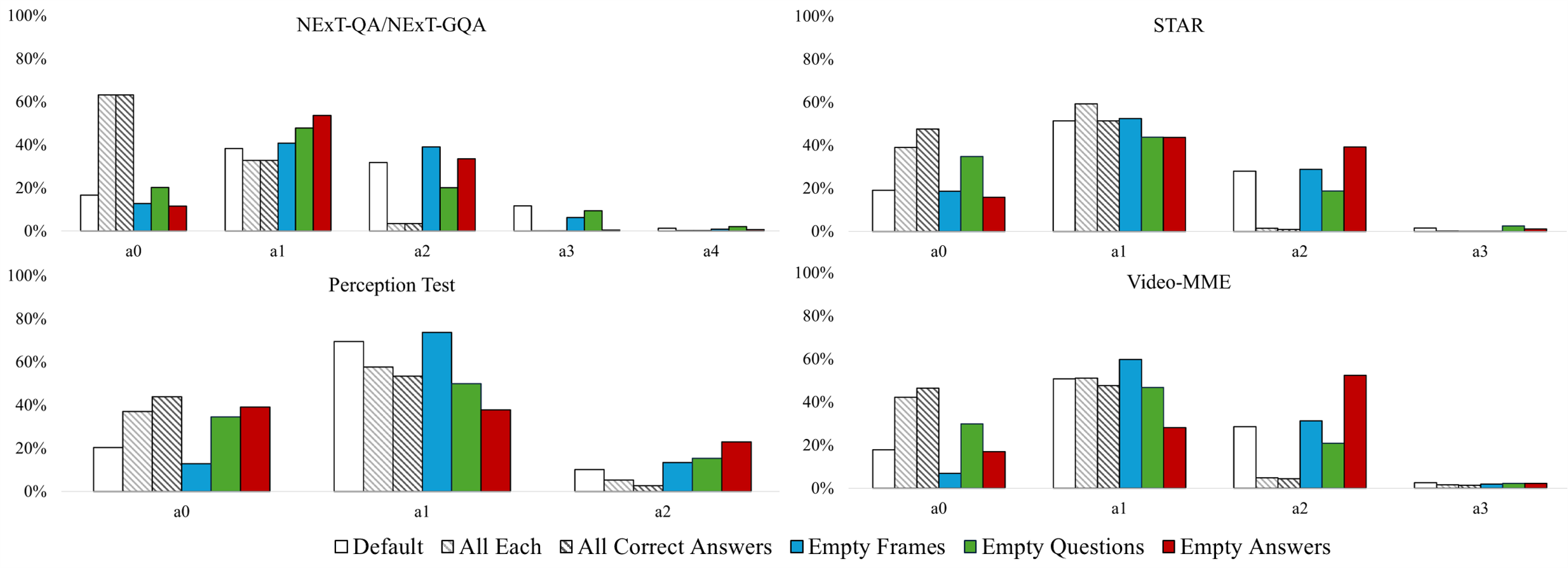}
  \caption{Video-LLaMA option distribution for accuracy-irrelevant settings across tested datasets. 
  \textit{All Each} represents the aggregated distribution under the \textit{All Identical Answers} setting. 
  }
  \label{fig:distr_llama}
\end{figure*}

\textbf{Video-LLaMA} \cite{zhang-etal-2023-video} has an additional audio channel and uses three query transformers (Q-formers) trained to align the video, audio, and language modalities. 

\textbf{Video-LLaVA} \citep{lin2023video} aligns videos and LLMs within a shared representation space and has improved multimodal instruction-following capabilities.

\textbf{SeViLA} \cite{yu2023selfchained}, specifically designed for question answering, implements a self-chaining approach with Localizer and Answerer: the first selects relevant frames, which the second uses to generate or choose answers.

Table~\ref{tab:inf_accuracy} presents the zero-shot inference results of each model on the chosen datasets. Further details on the models, their settings, and prompts can be found in Appendix~\ref{sec:app_models}.

\subsection{Experimental Settings}
\label{sec:settings}

Eleven dataset modifications are used to examine the role of each component in MCQA benchmarks: video, question, and answer options. The outmost purpose of the modifications is to identify 1) if bias is universal across models and datasets, 2) in what option(s) bias is, 3) how much each component contribute to the bias, and 4) in what settings it is the strongest with respect to each video-MCQA component.

\paragraph{Video Modifications} alter the video input: 

1. \textbf{Correct Frames}: Only frames containing the answer are given as video input. This tests if reducing extraneous visual information improves performance. The modification applies only to NExT-QA (NExT-GQA) and STAR due to the availability of timestamp annotations.

2. \textbf{Empty Frames}: Black frames are passed instead of meaningful video input. This tests the model's response to the absence of visual information within the multimodal setting. 

\paragraph{Question Modifications} adjust the questions in two ways: 

3. \textbf{Rephrased Questions}: Llama3 \cite{dubey2024llama3herdmodels} rephrases the original questions five times, then one rephrasing is randomly selected.\footnote{A manual check of 10\% of each rephrased dataset confirmed that meaning was retained, with minimal impact on accuracy due to meaning loss. Notably, very few hallucinations were spotted in the NExT-QA rephrased copy.} By changing the wording, we check if the selection bias is linked to the token bias. 

4. \textbf{Empty Questions}: Following \cite{balepur2024artifactsabductionllmsanswer}, we replace the question with an empty string to test if models can infer the question from answer choices and video.\footnote{The previous research for LLMs \cite{Balepur2024ArtifactsOA} has shown surprisingly high accuracy in this scenario. We investigate if the video modality contributes to even more meaningful predictions.}

\paragraph{Answer Option Modifications} examine the role of answer and its position in the emergence of bias:

5. \textbf{Answer Shuffling}: Options are randomly shuffled once. Previous experiments \cite{Zheng2023LargeLM,Zong2023FoolY} showed that the models are not robust against changing the option order. Regardless of accuracy, we check if the distribution of answers in each option is the same, which reveals the bias consistency.

6. \textbf{Correct Answer in Each Option}: The correct answer is placed in a fixed position; its content is swapped with another option. By changing the position of correct answer we test to what extend each option contribute to the models' choice.

7. \textbf{Correct Answer with Shuffling}: The correct answer is fixed, and the remaining options are shuffled. This fusion of the previous two approaches tests how stable a particular option choice is given that it is correct.

8. \textbf{Additional Empty Option}: A new empty option is added in the end to check for overfitting to specific options. If the added blank line is predicted, this indicates a biased distribution of predictions among a certain number of options.

9. \textbf{All Identical Answers}: All options are identical testing for deviation from uniform distribution. When the answer from each option becomes the answer of all options, the answer bias with respect to tokens is eliminated. Moreover, since there is no correct answer in the case of distractors, we can check the answer choice that happens without meaningful reasoning. 

10. \textbf{All Correct Answers}: Every option is correct. In addition to (9), where any answer is equal, in this setting any answer is valid. The expectation is to get the consistent uniform distribution.

11. \textbf{Empty Answers}: Each answer option is an empty string. With a non-existent answer, the option name and its position in the question are the only sources for choosing an answer.

To reduce token bias, we use less common option names a0, a1, a2, etc. instead of the conventional A, B, C, etc.

Modifications can also be categorized as accuracy-relevant (e.g., correct frames, shuffling, or rephrased questions) and accuracy-irrelevant (e.g., identical answers or empty parts). Further details, including accuracy results, confusion matrices and examples for each modification are provided in the Appendix \ref{sec:performance}. 

\begin{figure*}[t]
\centering
  \includegraphics[width=\textwidth]{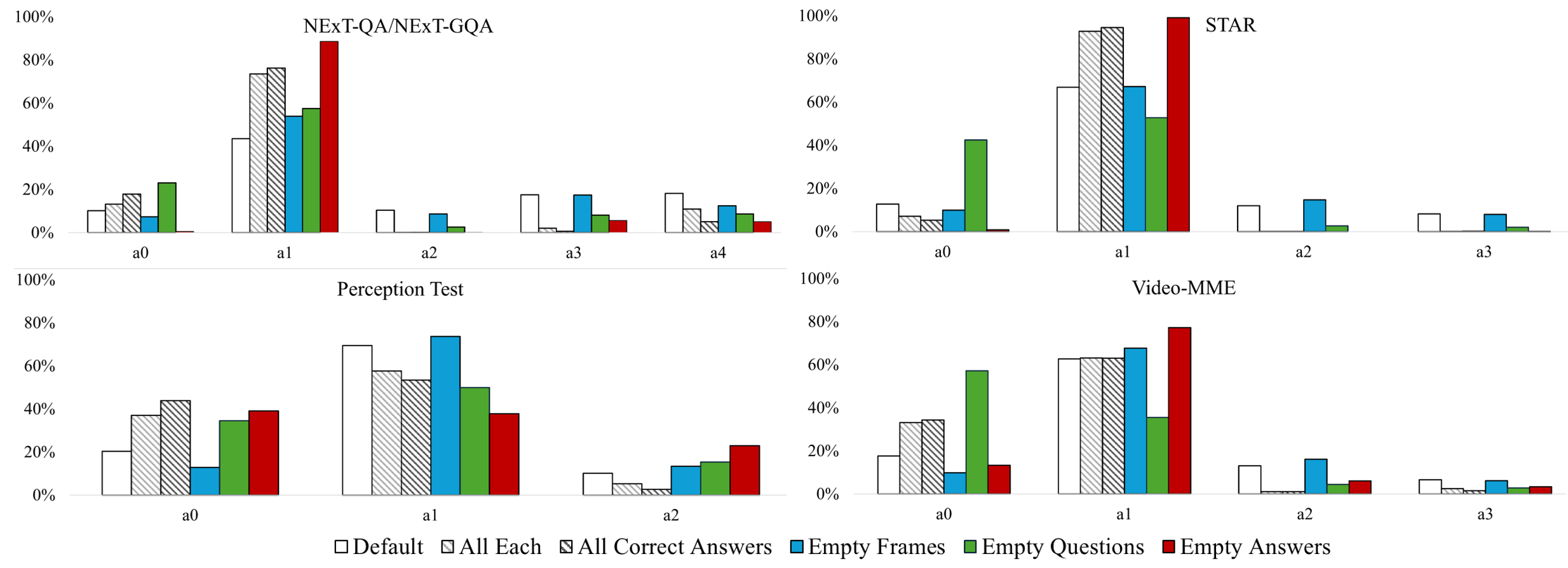}
  \caption{Video-LLaVA option distribution for accuracy-irrelevant settings across tested datasets. 
  \textit{All Each} represents the aggregated distribution under the \textit{All Identical Answers} setting. 
  }
  \label{fig:distr_llava}
\end{figure*}

\subsection{Observations}

\textbf{Accuracy-relevant} settings (Correct Frames, Rephrased Questions, and Answer Shuffling) vary in accuracy but yield almost the same option distributions as the default setting. This suggests that models rely more on answer position than on visual or textual content.

When the correct answer is placed in each option (Correct Answer in Each Option) or combined with shuffling (Correct Answer with Shuffling), the selection rate for that option increases, which indicates adequate reasoning. Moving the correct answer into less favored positions increases their selection proportion, and adding shuffling to it has minimal additional effect.

\begin{figure*}[t]
\centering
  \includegraphics[width=\textwidth]{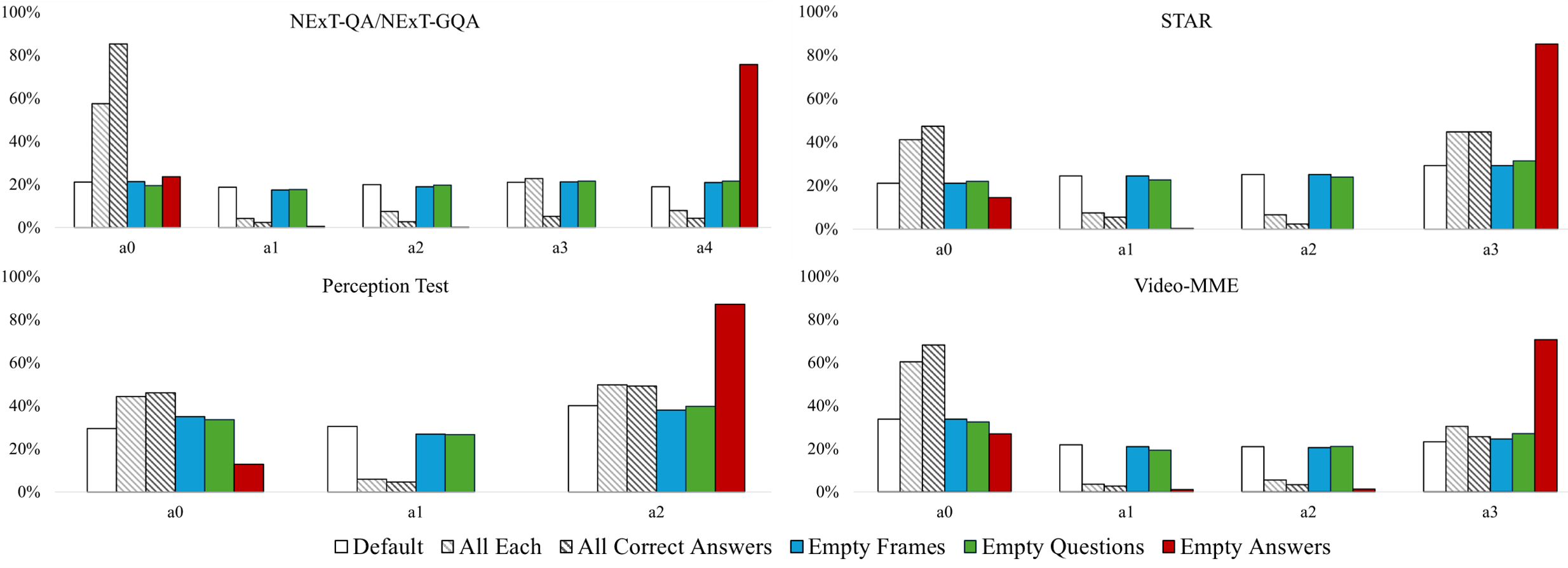}
  \caption{SeViLA option distribution for accuracy-irrelevant settings across tested datasets. 
  \textit{All Each} represents the aggregated distribution under the \textit{All Identical Answers} setting. 
  }
  \label{fig:distr_sevila}
\end{figure*}

In contrast, \textbf{accuracy-irrelevant} settings clearly expose bias. Under identical options (All Identical Answers and All Correct Answers) or incomplete conditions (Empty Frames, Empty Questions, Empty Answers), the predicted distributions deviate significantly from uniformity. Figures~\ref{fig:distr_llama}-\ref{fig:distr_sevila} highlight the patterns of the accuracy-agnostic settings as they are indicative for the study of bias, while full performance details are in Appendix~\ref{sec:performance}.

In case of All Identical Answers each-option answer showed consistently a very close distribution, so in the figures the result of All Each is obtained by averaging the outcomes across every option experiments within this setting. Tall bars concentrated in certain options indicate bias, while persistently low bars reveal neglected options. Each model exhibits pronounced biases toward specific options.

\textbf{Video-LLaMA} (Figure~\ref{fig:distr_llama}) strongly favors \textit{a1} and tends to ignore options beyond the third one. In the Perception Test (three options), it already shows bias against the third option. With identical answers, \textit{a0} sometimes competes with \textit{a1}, and when answers are empty, \textit{a2} surprisingly surpasses \textit{a1} in Video-MME. Empty settings cover the behavior of the bias to the greatest extent.

\textbf{Video-LLaVA} (Figure~\ref{fig:distr_llava}) behaves similarly but exhibits even stronger bias toward \textit{a1}, largely ignoring higher-numbered options. Unlike Video-LLaMA, it overwhelmingly favors \textit{a1} when answers are empty. With empty questions, \textit{a0} begins to compete with \textit{a1}, especially in Video-MME.

\textbf{SeViLA}\footnote{SeViLA uniquely becomes distracted when an empty option is added, suggesting it may have been trained with fixed option counts. Notably, it shows reduced accuracy for all-correct frames in STAR, which aligns with prior findings \cite{Goyal2016MakingTV, Lin2023RevisitingTR}, where "blind" language-only models were able to answer image-based questions.} demonstrates greater robustness on NExT-QA, maintaining about uniform 20\% distribution under harsh attacks. It only becomes unstable when all correct answers are identical (favoring \textit{a0}) and with empty answers (favoring the last option). Figure~\ref{fig:distr_sevila} shows slightly higher saturation for these conditions. Unlike Video-LLaMA and Video-LLaVA, which rely on multinomial probabilities, SeViLA uses the \textit{argmax} function, always selecting the first option if probabilities are equal. This explains why it often favors the first option, \textit{a0}, in identical-answer scenarios.

\section{Methodology}
\label{sec:method}

Our bias calibration approach is grounded in established guidelines for developing and validating multiple-choice test items \cite{haladyna2004developing}, which emphasize balanced, independent, and logically plausible options. This way we ensure that the datasets isolate positional bias and allow our calibration method to be applied effectively. Specifically, we require the following conditions to hold in the MCQA benchmark:

1. \textbf{Balanced options}: All answer options are presented equally without any positional or presentation bias. This is crucial and straightforward to achieve via minimal preprocessing.

2. \textbf{Independent options}: Each option for a given question is independent of others. There should be no overlapping content or logical dependencies among options that influence selection.

3. \textbf{No logically better or worse options except the correct one}: All distractors should be plausible and not trivially inferior or superior to the correct answer. This ensures that the model must rely on genuine reasoning.


\subsection{Evaluation Criteria}
\label{sec:eval}

Our framework both \textbf{detects bias} and \textbf{evaluates performance}. Bias metrics assess model's consistency in treating all answer options fairly, while performance metrics evaluate accuracy and reliability.

We monitor the standard deviations of the F1 score (F1\_std) and the recall (Recall\_std). A high F1\_std indicates inconsistent performance across options, signaling bias. A high Recall\_std suggests the model is better at identifying correct answers for certain options over others.

For option probabilities, we use the standard deviation of the symmetric Jensen-Shannon distance (JS\_std) to detect inconsistencies between the predicted and the true probability distributions. This metric, bounded between 0 and 1, is more interpretable than Kullback-Leibler (KL) divergence.

By tracking these metrics, we address group fairness, ensuring the model treats all options equitably. Large standard deviations in any metric suggest bias, which we aim to minimize.



\subsection{Debiasing Approach}

\begin{figure}[ht]
    \centering
    \includegraphics[width=0.35\textwidth]{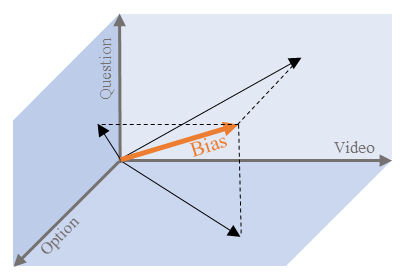}
    \caption{Bias vector and its projections on the decomposition planes.}
    \label{fig:vector}
\end{figure}


Every MCQA benchmark comprises essential components that make the answering task feasible. In the case of video-MCQA, these components include the video, the question, and the answer options. We construct decompositions where the task becomes impossible by design, as one of the key components is removed. This allows us to examine the projections of model's behavioral bias on each decomposition plane as intuitively shown in Figure \ref{fig:vector}. 
In a fair and unbiased model, projecting the dataset by omitting one of the key components would result in a uniform distribution. However, as shown in Section~\ref{sec:expSetup}, this is far from the case. In the decomposition planes, the bias is more clearly manifested, and since we expect uniformity, the bias becomes easier to detect and mitigate.

\subsubsection{Decomposition}


Let a Multiple-Choice Question Answering task $T$ be well-defined if there is one and only one correct option to answer the question based on the provided context.
Otherwise, the $T$ is ill-defined. We wish to decompose $T$ into the following independent components: video context, question, answer options. Omitting any of these components makes $T$ ill-defined. In general, there could be $n$ independent components \( C_1, C_2, \dots, C_n \). Let $P(T)$ be a well-defining property (correctness) of $T$  such that $P(T) = 1$ if $T$ is well-defined, otherwise - $P(T) = 0$. We assume that all tasks are initially well-defined.




A model $M$ takes a task $T$ as input and outputs an option ID corresponding to the correct answer for a video-based question: $M(T) = ID_{opt}$.

For an ill-defined task, a model that is both reasonable and unbiased should uniformly select any option ID. Formally, if $P(T)==0$: $M(T) = \text{RandomChoice} (\{ID_{opt}\})$.


The key insight from the analysis 
in Section \ref{sec:expSetup} is if the model has bias on a given dataset, this bias is likely to be more ubiquitous, hence evident, in the ill-defined versions of the tasks. 



Let $A$ be an attack on the well-defined task that removes one of the task component $C_i$ that makes this task defined insufficiently, i.e. $A(T) = \hat{T}$ and $P(\hat{T})=0$ while initially $P(T) = 1$. Let $A(D)$ be an attack on dataset $D$ which means applying element-wise the same attack $A$ to each $T$ from $D$. Subsequently, we consider three types of attacks that make the task ill-defined: $A_{v=0}$ for all zeroed frames in the video, $A_{q=0}$ for all questions as an empty string, and $A_{o=0}$ for all options as IDs only.

We investigate the importance of the following components of the bias $Bias_{M}$ : $Bias_{M}|_{A_{v=0}(D)}$, $Bias_{M}|_{A_{q=0}(D)}$, $Bias_{M}|_{A_{o=0}(D)}$, and how efficiently we can handle the general bias by balancing through these components.

\subsubsection{Debiasing}
The observed prediction distribution $P_{o}$ over options $\{d_i\}_{i=1}^{n}$ and conditioned on task $T$ where task $T = \text{(video, question, options)}$ can be decomposed as the prior distribution $P_{p}$ over $d_i$ and the debiased distribution $P_{d}$: 
\begin{equation}
\label{eq:equation_bias_prior_and_debiased}
P_{o}(d_i | T) = \frac{1}{Z_T} P_{p}(d_i | T) \times P_{d}(d_i | T)
\end{equation}

where $Z_{T}$ is a normalization factor.


The baseline method \textbf{BOLD} (Bias Optimisation Leveraging Decomposition) 
estimates the prior bias
as following:
\begin{equation}
\label{eq:bias_prior_decomposition}
\resizebox{0.85\hsize}{!}{$
\Tilde{P}_{p}(d_i | T) = softmax \left( \sum_{j}{P_{p}(d_i | A_j(T)} )\right)
$}
\end{equation}

First, we estimate the model bias on the dataset after attack A. We derive the $P_{p}$ knowing that $P_{d}$ is uniformly distributed from Equation ~\ref{eq:equation_bias_prior_and_debiased}: 
\begin{equation}
\label{eq:eq_attacked_prior}
\resizebox{0.75\hsize}{!}{$
    P_{p}(d_i \mid A(T)) = P_{o}(d_i \mid A(T))
$}
\end{equation}

\begin{table*}[ht]
\centering
\caption{Comparison of BOLD and Weighted\_BOLD bias mitigation approaches across models and datasets for performance and bias monitoring metrics with $k$ = 0.5. Green arrows indicate improvements: upward for Accuracy and F1\_mean, and downward for standard deviation metrics. Red arrows represent deterioration, respectively.}
\footnotesize{
\resizebox{\textwidth}{!}{
\begin{tabular}{l l p{2.2cm} r r r r r}
\hline
\multirow{2}{*}{\textbf{Model}} & \multirow{2}{*}{\textbf{Dataset}} & \multirow{2}{*}{\textbf{Configuration}} & \multicolumn{2}{c}{\textbf{Performance Metrics}} & \multicolumn{3}{c}{\textbf{Bias Monitoring Metrics}} \\ 
\cline{4-5} \cline{6-8}
& & & \textbf{Accuracy} & \textbf{F1\_mean} & \textbf{Recall\_std} & \textbf{F1\_std} & \textbf{JS\_std} \\ 
\hline
\multirow{4}{*} & NExT-QA & \textbf{BOLD} & 45.88 (\textcolor[rgb]{0,0.5,0}{↑}2.43\%) & 42.15 (\textcolor[rgb]{0,0.5,0}{↑}3.18\%) & 22.98 (\textcolor[rgb]{0,0.5,0}{↓}3.53\%) & 15.53 (\textcolor[rgb]{0,0.5,0}{↓}2.11\%) & 15.55 (\textcolor[rgb]{0,0.5,0}{↓}3.31\%)\\
& & \textbf{Weighted\_BOLD} & 45.91 (\textcolor[rgb]{0,0.5,0}{↑}2.51\%) & 42.2 (\textcolor[rgb]{0,0.5,0}{↑}3.29\%) & 22.83 (\textcolor[rgb]{0,0.5,0}{↓}4.19\%) & 15.51 (\textcolor[rgb]{0,0.5,0}{↓}2.2\%) & 15.56 (\textcolor[rgb]{0,0.5,0}{↓}3.27\%)\\
\cline{2-8}
\multirow{4}{*}{Video-LLaMA} & STAR & \textbf{BOLD} & 37.19 (\textcolor[rgb]{0,0.5,0}{↑}1.82\%) & 33.02 (\textcolor[rgb]{0,0.5,0}{↑}3.65\%) & 22.29 (\textcolor[rgb]{0,0.5,0}{↓}7.5\%) & 14.55 (\textcolor[rgb]{0,0.5,0}{↓}5.75\%) & 14.58 (\textcolor[rgb]{0,0.5,0}{↓}4.66\%)\\
& & \textbf{Weighted\_BOLD} & 37.34 (\textcolor[rgb]{0,0.5,0}{↑}2.24\%) & 33.5 (\textcolor[rgb]{0,0.5,0}{↑}5.16\%) & 21.13 (\textcolor[rgb]{0,0.5,0}{↓}12.3\%) & 14.05 (\textcolor[rgb]{0,0.5,0}{↓}8.98\%) & 14.14 (\textcolor[rgb]{0,0.5,0}{↓}7.54\%)\\
\cline{2-8}
\multirow{4}{*} & Perception Test & \textbf{BOLD} & 41.9 (\textcolor[rgb]{0,0.5,0}{↑}1.24\%) & 38.68 (\textcolor[rgb]{0,0.5,0}{↑}4.01\%) & 24.05 (\textcolor[rgb]{0,0.5,0}{↓}11.11\%) & 9.87 (\textcolor[rgb]{0,0.5,0}{↓}13.45\%) & 14.45 (\textcolor[rgb]{0,0.5,0}{↓}10.01\%)\\
& & \textbf{Weighted\_BOLD} & 42.06 (\textcolor[rgb]{0,0.5,0}{↑}1.62\%) & 39.36 (\textcolor[rgb]{0,0.5,0}{↑}5.86\%) & 21.8 (\textcolor[rgb]{0,0.5,0}{↓}19.45\%) & 9.11 (\textcolor[rgb]{0,0.5,0}{↓}20.06\%) & 13.02 (\textcolor[rgb]{0,0.5,0}{↓}18.94\%)\\
\cline{2-8}
\multirow{4}{*} & Video-MME & \textbf{BOLD} & 32.73 (\textcolor[rgb]{0,0.5,0}{↑}2.13\%) & 29.23 (\textcolor[rgb]{0,0.5,0}{↑}3.85\%) & 19.29 (\textcolor[rgb]{0,0.5,0}{↓}6.45\%) & 11.96 (\textcolor[rgb]{0,0.5,0}{↓}4.46\%) & 13.1 (\textcolor[rgb]{0,0.5,0}{↓}4.73\%)\\
& & \textbf{Weighted\_BOLD} & 32.2 (\textcolor[rgb]{0,0.5,0}{↑}0.47\%) & 28.98 (\textcolor[rgb]{0,0.5,0}{↑}2.94\%) & 17.65 (\textcolor[rgb]{0,0.5,0}{↓}14.38\%) & 11.69 (\textcolor[rgb]{0,0.5,0}{↓}6.56\%) & 12.64 (\textcolor[rgb]{0,0.5,0}{↓}8.04\%)\\
\cline{1-8}
\multirow{4}{*}{} & NExT-QA & \textbf{BOLD} & 51.81 (\textcolor[rgb]{0,0.5,0}{↑}3.72\%) & 51.71 (\textcolor[rgb]{0,0.5,0}{↑}3.82\%) & 13.27 (\textcolor[rgb]{0,0.5,0}{↓}18.63\%) & 2.67 (\textcolor[rgb]{0,0.5,0}{↓}18.42\%) & 4.58 (\textcolor[rgb]{0,0.5,0}{↓}15.06\%)\\
& & \textbf{Weighted\_BOLD} & 52.15 (\textcolor[rgb]{0,0.5,0}{↑}4.39\%) & 52.04 (\textcolor[rgb]{0,0.5,0}{↑}4.49\%) & 12.72 (\textcolor[rgb]{0,0.5,0}{↓}22.02\%) & 2.62 (\textcolor[rgb]{0,0.5,0}{↓}19.83\%) & 4.49 (\textcolor[rgb]{0,0.5,0}{↓}16.83\%)\\
\cline{2-8}
\multirow{4}{*}{Video-LLaVA} & STAR & \textbf{BOLD} & 37.21 (\textcolor[rgb]{0,0.5,0}{↑}7.01\%) & 35.76 (\textcolor[rgb]{0,0.5,0}{↑}12.37\%) & 18.28 (\textcolor[rgb]{0,0.5,0}{↓}26.85\%) & 4.97 (\textcolor[rgb]{0,0.5,0}{↓}27.15\%) & 4.54 (\textcolor[rgb]{0,0.5,0}{↓}21.39\%)\\
& & \textbf{Weighted\_BOLD} & 37.53 (\textcolor[rgb]{0,0.5,0}{↑}7.94\%) & 36.18 (\textcolor[rgb]{0,0.5,0}{↑}13.69\%) & 17.45 (\textcolor[rgb]{0,0.5,0}{↓}30.15\%) & 4.91 (\textcolor[rgb]{0,0.5,0}{↓}28.03\%) & 4.26 (\textcolor[rgb]{0,0.5,0}{↓}26.24\%)\\
\cline{2-8}
\multirow{4}{*}{} & Perception Test & \textbf{BOLD} & 41.62 (\textcolor[rgb]{0,0.5,0}{↑}2.22\%) & 38.69 (\textcolor[rgb]{0,0.5,0}{↑}8.4\%) & 21.75 (\textcolor[rgb]{0,0.5,0}{↓}20.9\%) & 10.92 (\textcolor[rgb]{0,0.5,0}{↓}25.97\%) & 3.78 (\textcolor[rgb]{0,0.5,0}{↓}28.4\%)\\
& & \textbf{Weighted\_BOLD} & 41.95 (\textcolor[rgb]{0,0.5,0}{↑}3.02\%) & 39.46 (\textcolor[rgb]{0,0.5,0}{↑}10.58\%) & 19.73 (\textcolor[rgb]{0,0.5,0}{↓}28.26\%) & 10.24 (\textcolor[rgb]{0,0.5,0}{↓}30.53\%) & 3.4 (\textcolor[rgb]{0,0.5,0}{↓}35.51\%)\\
\cline{2-8}
\multirow{4}{*}{} & Video-MME & \textbf{BOLD} & 34.7 (\textcolor[rgb]{0,0.5,0}{↑}1.19\%) & 32.79 (\textcolor[rgb]{0,0.5,0}{↑}5.81\%) & 18.19 (\textcolor[rgb]{0,0.5,0}{↓}24.51\%) & 6.16 (\textcolor[rgb]{0,0.5,0}{↓}23.63\%) & 3.84 (\textcolor[rgb]{0,0.5,0}{↓}19.93\%)\\
& & \textbf{Weighted\_BOLD} & 34.63 (\textcolor[rgb]{0,0.5,0}{↑}0.97\%) & 32.97 (\textcolor[rgb]{0,0.5,0}{↑}6.38\%) & 16.36 (\textcolor[rgb]{0,0.5,0}{↓}32.07\%) & 6.01 (\textcolor[rgb]{0,0.5,0}{↓}25.43\%) & 3.43 (\textcolor[rgb]{0,0.5,0}{↓}28.5\%)\\
\cline{1-8}
\multirow{4}{*} & NExT-QA & \textbf{BOLD} & 63.92 (\textcolor[rgb]{0,0.5,0}{↑}0.02\%) & 63.89 (\textcolor[rgb]{0,0.5,0}{↑}0.02\%) & 2.03 (\textcolor[rgb]{0,0.5,0}{↓}5.47\%) & 1.18 (\textcolor[rgb]{0,0.5,0}{↓}10.4\%) & 1.99 (\textcolor[rgb]{0,0.5,0}{↓}0.17\%)\\
& & \textbf{Weighted\_BOLD} & 63.93 (\textcolor[rgb]{0,0.5,0}{↑}0.04\%) & 63.91 (\textcolor[rgb]{0,0.5,0}{↑}0.04\%) & 1.99 (\textcolor[rgb]{0,0.5,0}{↓}7.64\%) & 1.19 (\textcolor[rgb]{0,0.5,0}{↓}9.83\%) & 1.99 (\textcolor[rgb]{0,0.5,0}{↓}0.04\%)\\
\cline{2-8}
\multirow{4}{*}{SeViLA} & STAR & \textbf{BOLD} & 46.22 (\textcolor{red}{↓}0.12\%) & 46.1 (\textcolor{red}{↓}0.08\%) & 4.13 (\textcolor[rgb]{0,0.5,0}{↓}7.46\%) & 2.26 (\textcolor[rgb]{0,0.5,0}{↓}1.78\%) & 2.1 (\textcolor[rgb]{0,0.5,0}{↓}6.64\%)\\
& & \textbf{Weighted\_BOLD} & 46.2 (\textcolor{red}{↓}0.18\%) & 46.08 (\textcolor{red}{↓}0.13\%) & 4.01 (\textcolor[rgb]{0,0.5,0}{↓}10.17\%) & 2.2 (\textcolor[rgb]{0,0.5,0}{↓}4.36\%) & 2.07 (\textcolor[rgb]{0,0.5,0}{↓}8.2\%)\\
\cline{2-8}
\multirow{4}{*} & Perception Test & \textbf{BOLD} & 45.32 (\textcolor[rgb]{0,0.5,0}{↑}0.06\%) & 45.18 (\textcolor[rgb]{0,0.5,0}{↑}0.16\%) & 5.18 (\textcolor[rgb]{0,0.5,0}{↓}16.61\%) & 2.95 (\textcolor[rgb]{0,0.5,0}{↓}2.43\%) & 1.3 (\textcolor[rgb]{0,0.5,0}{↓}18.42\%)\\
& & \textbf{Weighted\_BOLD} & 45.31 (\textcolor[rgb]{0,0.5,0}{↑}0.03\%) & 45.2 (\textcolor[rgb]{0,0.5,0}{↑}0.21\%) & 4.56 (\textcolor[rgb]{0,0.5,0}{↓}26.58\%) & 2.83 (\textcolor[rgb]{0,0.5,0}{↓}6.43\%) & 1.08 (\textcolor[rgb]{0,0.5,0}{↓}31.92\%)\\
\cline{2-8}
\multirow{4}{*} & Video-MME & \textbf{BOLD} & 40.19 (\textcolor[rgb]{0,0.5,0}{↑}0.84\%) & 40.17 (\textcolor[rgb]{0,0.5,0}{↑}0.88\%) & 4.11 (\textcolor[rgb]{0,0.5,0}{↓}12.03\%) & 1.41 (\textcolor[rgb]{0,0.5,0}{↓}16.19\%) & 0.73 (\textcolor[rgb]{0,0.5,0}{↓}12.52\%)\\
& & \textbf{Weighted\_BOLD} & 40.04 (\textcolor[rgb]{0,0.5,0}{↑}0.46\%) & 40.03 (\textcolor[rgb]{0,0.5,0}{↑}0.54\%) & 3.78 (\textcolor[rgb]{0,0.5,0}{↓}19.14\%) & 1.44 (\textcolor[rgb]{0,0.5,0}{↓}14.6\%) & 0.69 (\textcolor[rgb]{0,0.5,0}{↓}17.22\%)\\
\cline{2-8}
\hline
\end{tabular}
}
}
\label{tab:res}
\end{table*}

To estimate the bias under $A_{v=0}$, $A_{q=0}$ and $A_{o=0}$ attacks, we first compute $P_{p}(d_i | A_{v=0}(T))$, $P_{p}(d_i | A_{q=0}(T))$ and $P_{p}(d_i | A_{o=0})$, and then the prior estimation, respectively: $
\Tilde{ P}_{p}(d_i | T) = softmax [ P_{p}(d_i | A_{v=0}(T)) + P_{p}(d_i | A_{q=0}(T)) + P_{p}(d_i | A_{o=0}(T))]
$.

According to \citet{zheng2024largelanguagemodelsrobust}, 
biases related to option IDs may transfer effectively across different samples and domains. Therefore, finding the global prior with $K$ samples, we can mitigate the bias for the entire dataset with minimal efforts. 

From the dataset $D$, we take $K = k*||D||$ samples, denoted as $D_k$,  where $k$ is adjusted coefficient based on estimation budget. 
Each sample in $D_k$ is subjected to the set of attacks $A$. For each, we calculate the sample-specific prior $P_{p}(d_i | A(T))$  by Equation \ref{eq:eq_attacked_prior} 
and estimate the adjusted prior $\tilde{P}_{p}(d_i | T)$ by Equation \ref{eq:bias_prior_decomposition}. The global prior, $\tilde{P}_{\text{p}}(d_i)$, is then obtained by averaging the priors for each sample.

Ultimately, the posterior debiased through the global prior is calculated as follows: 
\begin{equation}
\resizebox{0.9\hsize}{!}{$
P_{d}(d_i | T) = softmax \left( \log P_o(d_i | T) - \log \tilde{P}_{p}(d_i) \right)
$}
\end{equation}

In the extended version, \textbf{Weighted\_BOLD}, we use weights for $P_{p}(d_i | A(T))$: 
\begin{equation}
\label{eq:bias_prior_decomposition_extended}
\resizebox{0.9\hsize}{!}{$
\Tilde{P}_{p}(d_i | T) = softmax \left( \sum_{j}{w_jP_{p}(d_i | A_j(T)} )\right)
$}
\end{equation}
This allows for a more optimal direction in the latent space of the priors, as we can prioritize certain priors by significance and refine the prior estimation $\Tilde{P}_{p}(d_i | T)$.

To gain the optimal weights, we use a 5-fold cross-validation procedure. In each fold, we optimize weights $\{w_i\}$ using the COBYLA algorithm \cite{Powell1994ADS}, subject to the constraints $0 \leq w_i \leq 1$ or \( |w_i| \leq 1 \). 

The restriction $0 \leq w_i \leq 1$ implies that the resulting debiasing vector is formed as a convex combination of the prior directions, remaining within the positive span of the original vectors. With this easy-to-interpret setup, we analyze how much each prior contributes to the final debiasing vector. Notably, a weight distribution with $w_i = 0$ enables ablation-style analysis by turning off the corresponding debiasing component. Uninformative or detrimental priors are automatically suppressed, resulting in a more streamlined and potentially more generalizable debiasing vector. When we allow the more relaxed constraint  \( |w_i| \leq 1 \), the COBYLA optimizer explores the full linear subspace spanned by the prior vectors, and the solution is no longer restricted to the positive orthant. This provides additional flexibility to correct for complex biases that may not be adequately addressed by additive combinations alone. 

While we minimize the standard deviation of Recall\_std across the options on the test set, we monitor the same bias metrics on the validation set. 
After completing the cross-validation we average the estimated global priors across the folds and apply this averaged global prior. Both algorithms, BOLD and Weighted\_BOLD, are outlined in Appendix \ref{sec:algorithms}.

\section{Results}
\label{sec:results}

We conducted tests with the $k$ sample parameter set to 0.25, 0.5, 0.75, and 1 (the entire dataset) in combination with positive and negative weights. The configuration with $k$=0.5 and \( w_i \geq 0 \) provided the optimal balance between sample size and performance. As shown in Table~\ref{tab:res} with results for positive weights and $k$=0.5,\footnote{Additional results for $k$ values of 0.25, 0.75, and 1 combined with different weights are provided in Appendix~\ref{sec:algorithms}.} the overall significant improvement in bias metrics, including \textit{Recall\_std}, indicates that our method effectively reduces option disparity. 

Weighted\_BOLD generally outperforms BOLD because the weights capture critical priors in the latent space. Thus, they prove to be effective for fine-tuning bias correction within the decomposed priors.

Both BOLD and Weighted\_BOLD consistently improve performance metrics for Video-LLaMA and Video-LLaVA. SeViLA generally shows more resilience to bias calibration, with minimal effect on performance on NExT-QA and marginal deterioration on STAR. This may be attributed to the model's training for statistically close to uniform distribution of results over a fixed options number. Despite decreases in Accuracy and F1 Mean on STAR, the reduction in \textit{Recall\_std} across all datasets for this model suggests that its debiased responses are more truthful, reflecting a fairer distribution of correct answers across options.

Video-LLaVA, which exhibits the most pronounced selection bias, benefits the most from bias reduction via prior decomposition across all datasets, achieving a maximum accuracy gain of 7.94\% on STAR with the weighted configuration.
\section{Conclusion}
\label{sec:conc}






We presented the first comprehensive analysis of selection bias in video-based MCQA tasks, using modified datasets to pinpoint how bias emerges within specific answer options and across different VLM architectures. Our findings reveal that while all models share common tendencies in answer distribution, they also possess distinct bias profiles both across and within datasets. Notably, the bias vector is consistently more pronounced when projecting onto the decomposed planes where one key MCQA component (video, question, or answer options) is removed.

Building on these insights, we proposed the BOLD calibration method using a global prior bias vector, derived from the decomposed versions of the task, to mitigate selection bias. Our technique requires only three decompositions per question and is thus resource-efficient. Our BOLD bias removal not only enhances fairness and interpretability but also improves performance metrics. Extending the method to Weighted\_BOLD refines the balance between debiasing and performance gains.

This easily scalable, post-processing, automatic, and unsupervised method is applicable to any existing VLM and allows for a sharper focus on its true reasoning capabilities. By making MCQA benchmarks more robust, transparent, and impartial, we enable a clearer assessment of the genuine reasoning cues in VLMs and pave the way for future improvements in model design and training strategies.

Furthermore, elevating bias suppression to an abstract level, where tasks are systematically decomposed and key components are removed to identify bias, we offer a framework beyond the current scope to address other bias-related challenges in machine learning models.
\section{Limitations}
\label{sec:limiit}

Our approach rests on the assumption that when models lack sufficient direct information, they rely on indirect cues, manifesting as selection bias. While we have shown that bias can be decomposed by selectively removing the video, question, or answer component, this decomposition may not be exhaustive. There could be more optimal directions in the latent space of decomposed priors. Similarly, although our method identifies pronounced bias when all answers are correct, there remain scenarios where further supervised debiasing (e.g., applying COBYLA optimization with true labels) might yield more optimal results. We aimed for an easy-to-implement, unsupervised solution without relying on ground-truth answers.

We also assume the validity of uniform distribution as a baseline: Jensen-Shannon Divergence operates under the assumption that, in the absence of prior information, the distribution should be uniform. If the test MCQA data or task suggest an uneven distribution of answer options, this may lead to underestimating the significance of certain factors.

Another limitation is that we primarily assessed our post-processing calibration on the original datasets rather than systematically verifying performance improvements across all modifications. While our current evaluation covers four challenging video datasets and three model architectures, more granular, scenario-specific testing would provide stronger evidence of robustness. However, performing comprehensive evaluations on every modification poses practical challenges. These limitations point toward future work: exploring more nuanced testing scenarios, investigating methods that do not rely on uniform reference distributions, and potentially leveraging additional computational resources or innovative techniques for even more thorough bias and performance assessments.

\section{Ethics Statement}
\label{sec:ethical}

We based our experiments on existing datasets, ensuring compliance with their respective copyrights. Our research uses open-source VLMs, which, while offering accessibility and flexibility, also carry the inherent risks associated with open-ended text generation.

\section*{Acknowledgements}
Olga Loginova thanks Amazon Alexa for supporting her research through a generous donation to Raffaella Bernardi. Ravi Shekhar has been partially funded by the UKRI and the European Union. The views and opinions expressed are, however, those of the authors only and do not necessarily reflect those of the European Union or Research Executive Agency. Neither the European Union nor the granting authority can be held responsible for them. All authors are deeply grateful to Giovanni Bonetta for his invaluable contributions and insightful consultations at earlier stages of the project.

\bibliography{custom}
\appendix
\label{sec:appendix} 

\section{Datasets}
\label{sec:dat}

\subsection{NExT-QA and NExT-GQA}

\begin{figure}[ht]
  \includegraphics[width=\columnwidth]{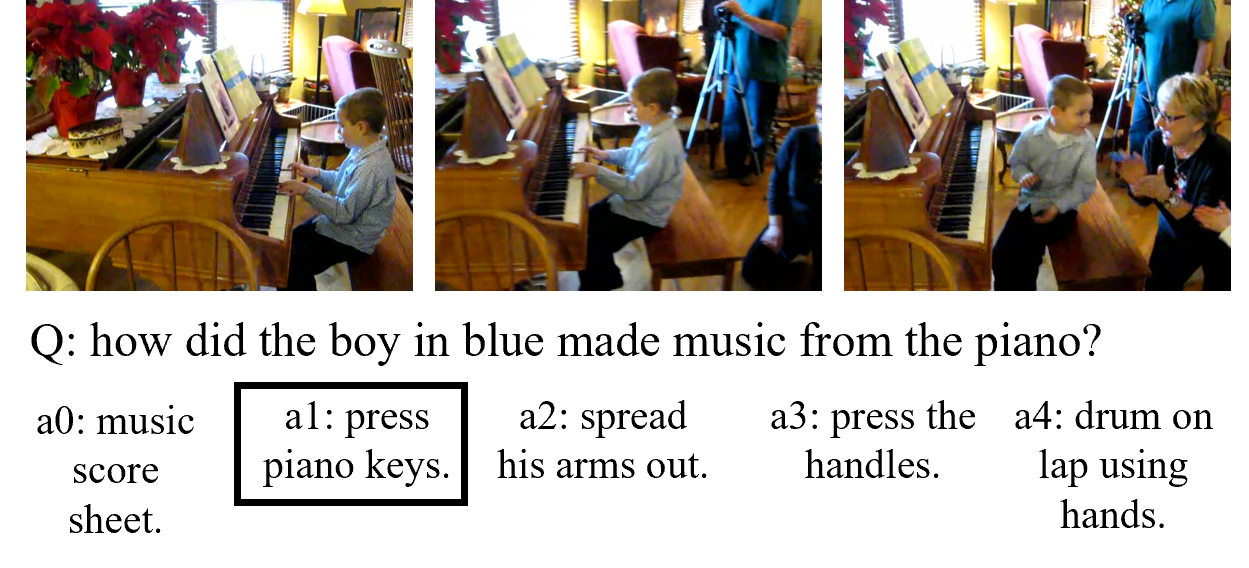}
  \caption{Example of a QA pair from NExT-QA in the Causal category. The correct answer is in the box.}
  \label{fig:nextqa_example_1}
\end{figure}

\begin{figure}[H]
  \includegraphics[width=\columnwidth]{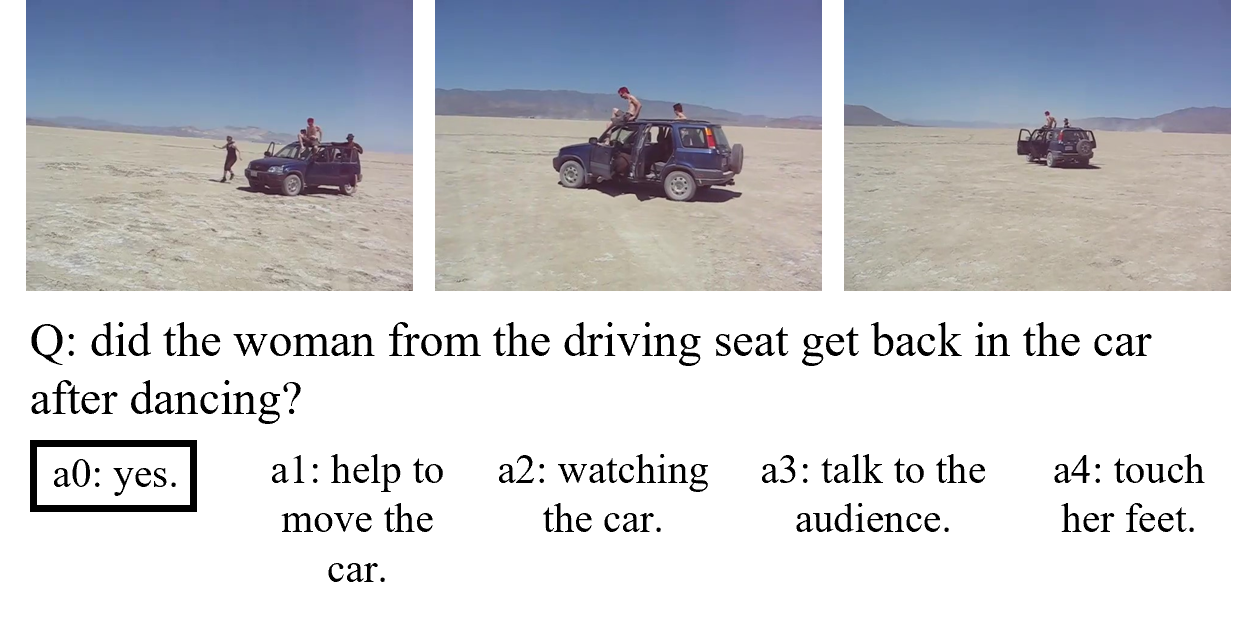}
  \caption{Example of a QA pair from NExT-QA in the Temporal category. The correct answer is in the box.}
  \label{fig:nextqa_example_2}
\end{figure}

\begin{figure}[H]
  \includegraphics[width=\columnwidth]{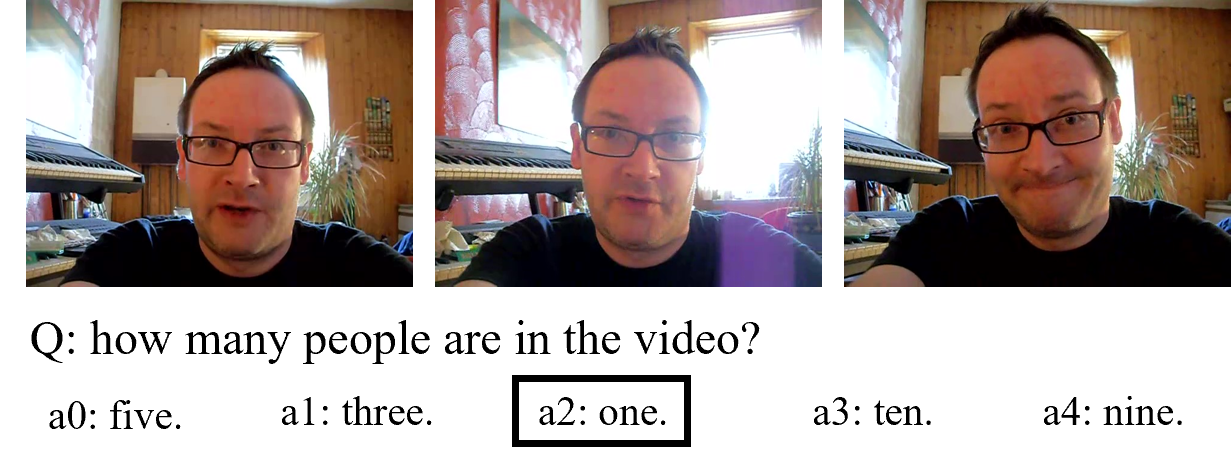}
  \caption{Example of a QA pair from NExT-QA in the Descriptive category. The correct answer is in the box.}
  \label{fig:nextqa_example_3}
\end{figure}

NExT-QA, released in 2021, is one of the first datasets designed to assess a VLM's ability to capture causal, temporal, and descriptive topics through what/when/why/how and before/after questions. It comprises videos sourced from the VidOR dataset~\cite{Shang2019AnnotatingOA}, with an average length of 35.73 seconds. The test split primarily contains causal questions (52.57\%) addressing \textit{why} and \textit{how}, temporal reasoning questions (31.03\%) about preceeding, following, or concurrent actions, and descriptive tasks (16.04\%). Videos are manually annotated with questions and correct answers. Distractors are selected from 50 candidates based on a cosine similarity of less than 0.9 to the correct answer, ensuring they are close but not identical. This approach sometimes allows questions to be meaningfully answered without viewing the video (see Figures ~\ref{fig:nextqa_example_1}-~\ref{fig:nextqa_example_3}).

Since our models were released after 2021, they might have included NExT-QA in their training or fine-tuning data.

NExT-GQA, released in 2024 as an extension of NExT-QA, focuses on evaluating the localization abilities of VLMs. It includes manually annotated timestamp labels that are key for answering the questions. Descriptive questions are excluded because they pertain to global video content or have answers available throughout the video with no specific temporal grounding. We also excluded QA pairs with multiple annotated time intervals. Consequently, the test set of NExT-GQA comprises 56.69\% causal questions and 43.31\% temporal questions.

\subsection{STAR}


The STAR dataset is derived from the Action Genome dataset~\cite{Ji2019ActionGA}, which itself is based on Charades~\cite{YuanyuanICCV2017}. Charades comprises indoor activity videos with an average length of 30 seconds. Action Genome provides annotations of objects and their relationships within these videos.

The question-answer pairs are generated using functional programs based on situation hyper-graphs: the graphs represent the extraction of abstract representations from the videos, while the functional programs provide multiple types of questions that cover different levels of difficulty in situated reasoning. There are four types of questions focusing on occurred facts, temporal order, future probability, and feasibility in specific situations:
\begin{itemize} 
\item \textbf{Sequence} (50.53\%): What did the person do before/after ...?''
\item \textbf{Interaction} (33.78\%): ``What did a person do ...?''
\item \textbf{Prediction} (8.79\%): ``What will the person do next with ...?''
\item \textbf{Feasibility} (6.9\%): ``What is the person able to do?'' or ``Which object is possible to be ...?'' 
\end{itemize}
All QA pairs are annotated with a timestamp label. 
To avoid answer distribution bias, each question type is balanced through resampling. To prevent reasoning shortcuts due to frequent action or entity co-occurrences and to ensure models cannot easily answer without genuine understanding, the compositionality of verbs and nouns is controlled.

\begin{figure}[H]
  \includegraphics[width=\columnwidth]{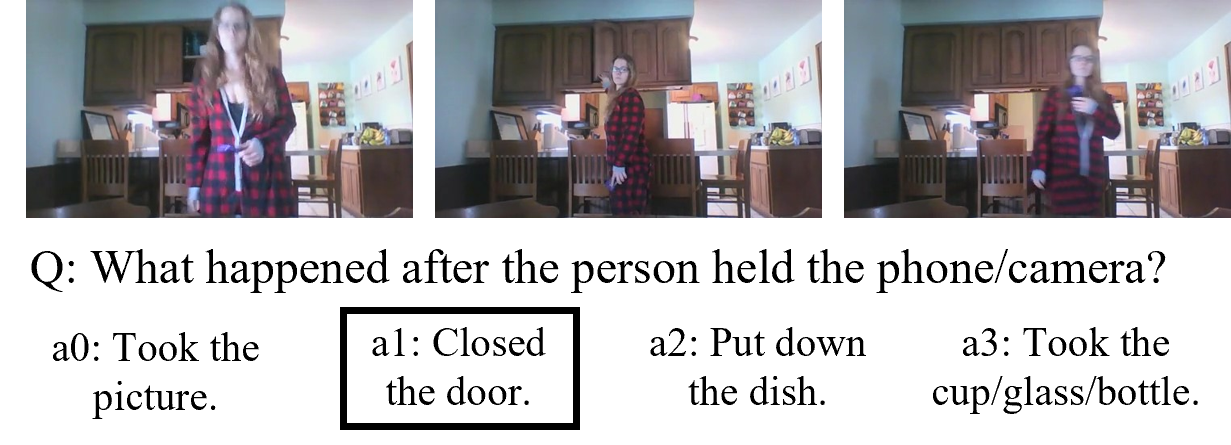}
  \caption{Example of a QA pair from STAR in the Sequence category. The correct answer is in the box.}
  \label{fig:star_example_1}
\end{figure}

\begin{figure}[H]
  \includegraphics[width=\columnwidth]{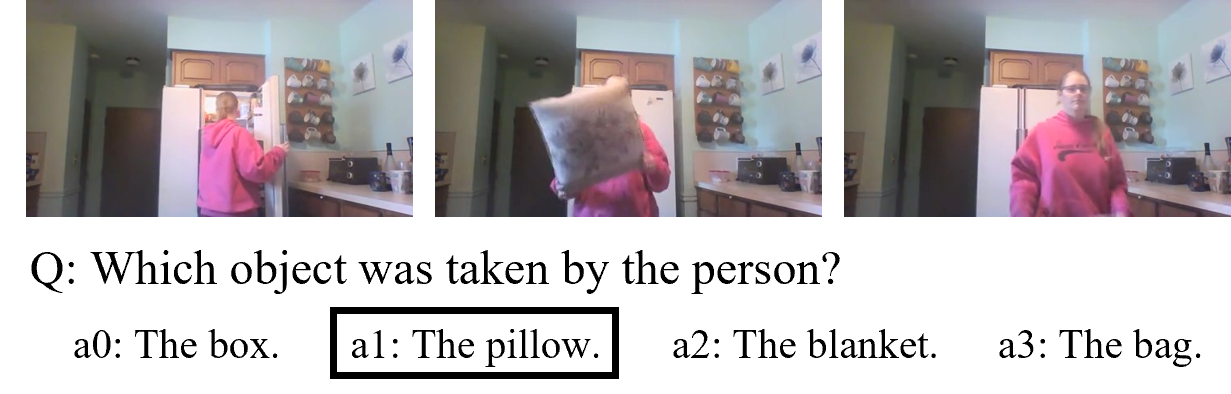}
  \caption{Example of a QA pair from STAR in the Interaction category. The correct answer is in the box.}
  \label{fig:star_example_2}
\end{figure}

\begin{figure}[H]
  \includegraphics[width=\columnwidth]{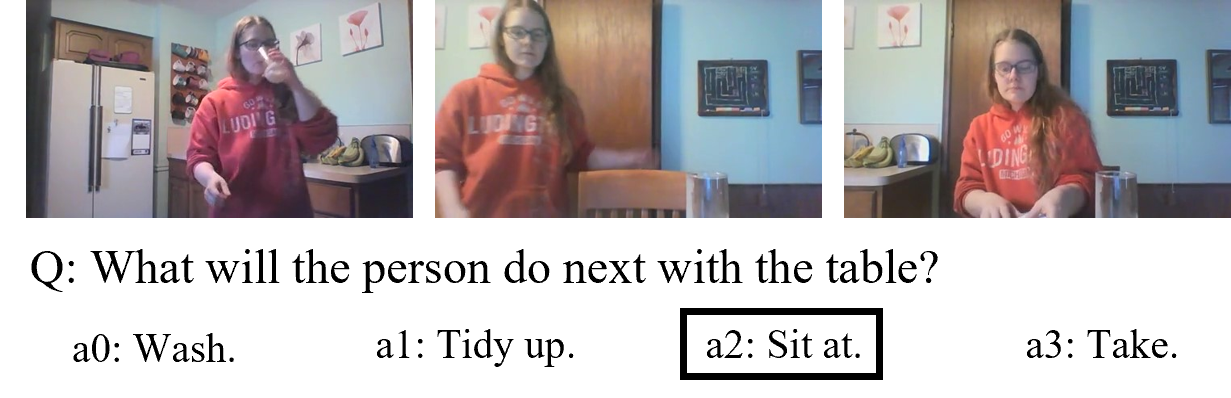}
  \caption{Example of a QA pair from STAR in the Prediction category. The correct answer is in the box.}
  \label{fig:star_example_3}
\end{figure}

\begin{figure}[H]
  \includegraphics[width=\columnwidth]{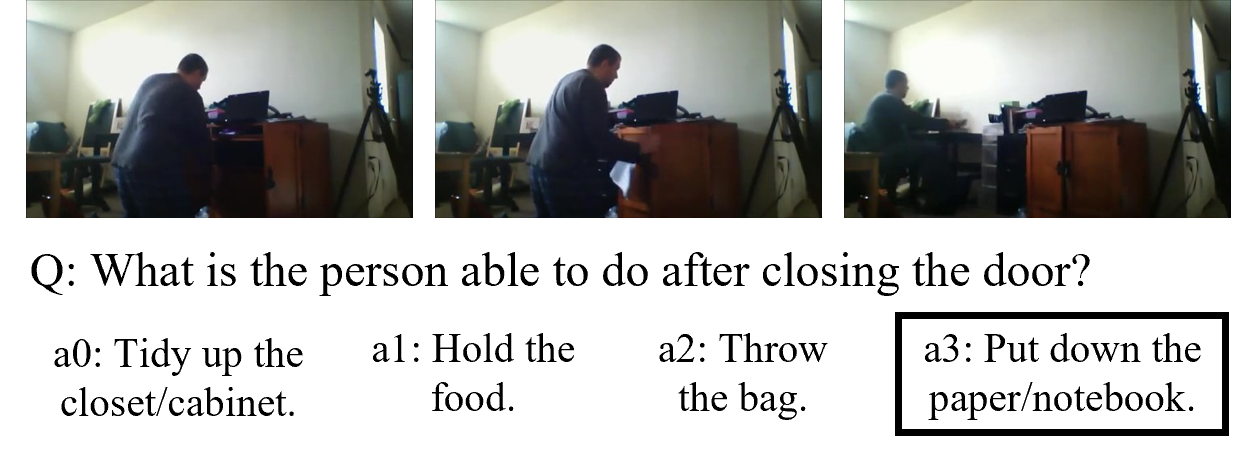}
  \caption{Example of a QA pair from STAR in the Feasibility category. The correct answer is in the box.}
  \label{fig:star_example_4}
\end{figure}

Examples are shown in Figures ~\ref{fig:star_example_1}-~\ref{fig:star_example_4}.

\subsection{Perception Test}

The Perception Test dataset consists of four categories designed to evaluate VLMs on different types of reasoning:

\begin{itemize} \item \textbf{Physics} (36.26\%) \item \textbf{Semantics} (30.54\%) \item \textbf{Abstraction} (29.28\%) \item \textbf{Memory} (3.92\%) \end{itemize}

These categories test descriptive, explanatory, predictive, and counterfactual reasoning abilities. The video scripts are based on human perception screening tests, with an average length of 23 seconds. All answers are crowd-sourced. Distractors are both manually created and automatically generated, particularly for standard closed yes/no questions.

We use a 40\% split of the correct-answer-annotated validation set to save the resources. Sampling is stratified to maintain the original balance of question types and the correct answer distribution. The split contains approximately 15\% closed questions, which poses an additional challenge for models that tend to exhibit affirmation bias~\cite{Pezeshkpour2023LargeLM}. Notably, for closed questions in this dataset, Video-LLaMA and Video-LLaVA often respond with "yes" or "no" rather than selecting an answer option.

\begin{figure}[H]
  \includegraphics[width=\columnwidth]{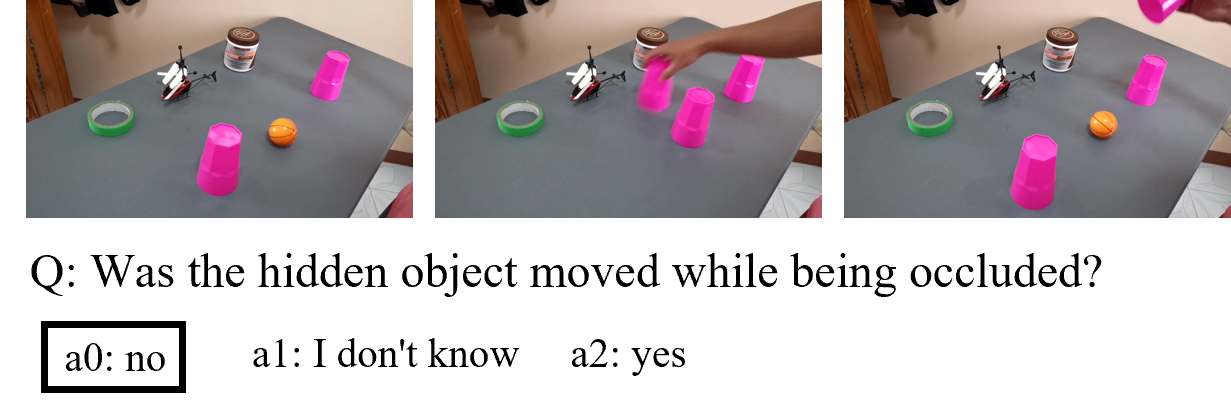}
  \caption{Example of a QA pair from Perception Test in the Descriptive Physics category. The correct answer is in the box.}
  \label{fig:pt_example_1}
\end{figure}

\begin{figure}[H]
  \includegraphics[width=\columnwidth]{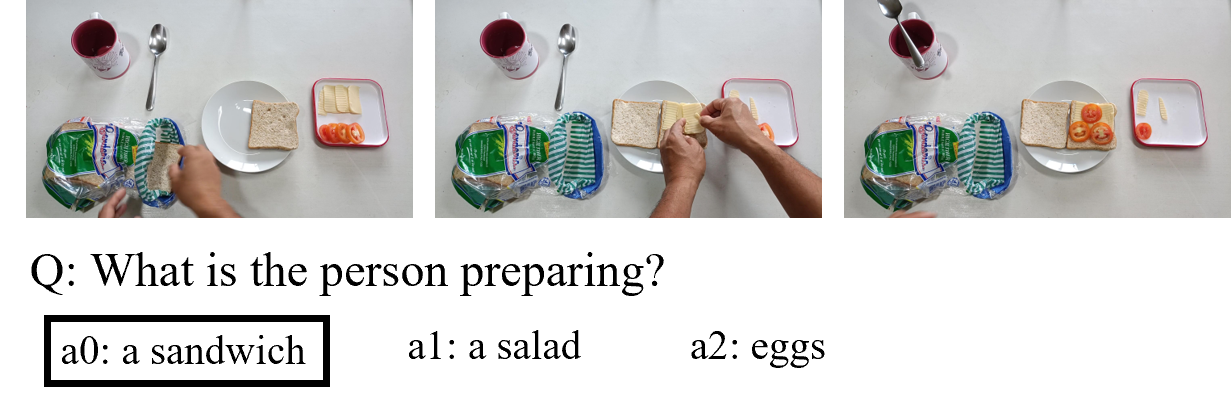}
  \caption{Example of a QA pair from Perception Test in the Descriptive Semantics category. The correct answer is in the box.}
  \label{fig:pt_example_2}
\end{figure}

\begin{figure}[H]
  \includegraphics[width=\columnwidth]{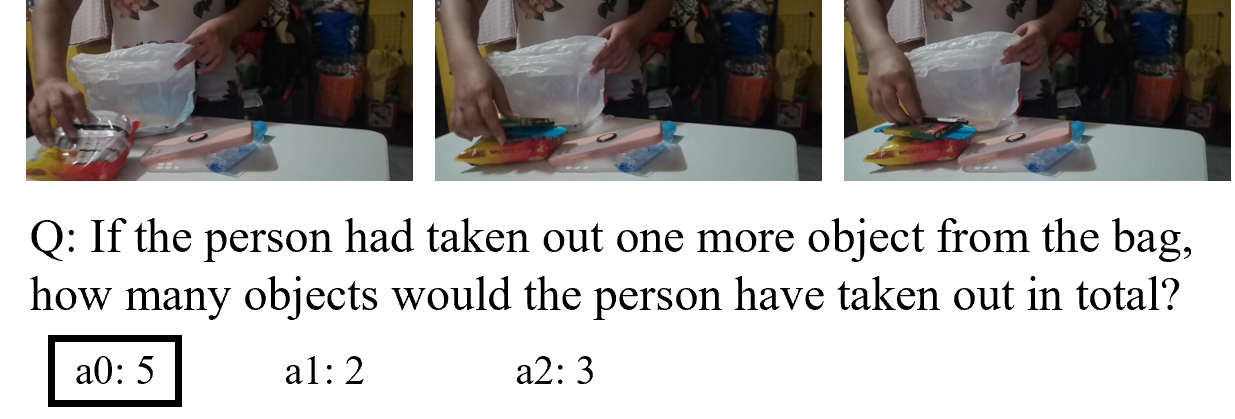}
  \caption{Example of a QA pair from Perception Test in the Counterfactual Abstraction category. The correct answer is in the box.}
  \label{fig:pt_example_3}
\end{figure}

\begin{figure}[H]
  \includegraphics[width=\columnwidth]{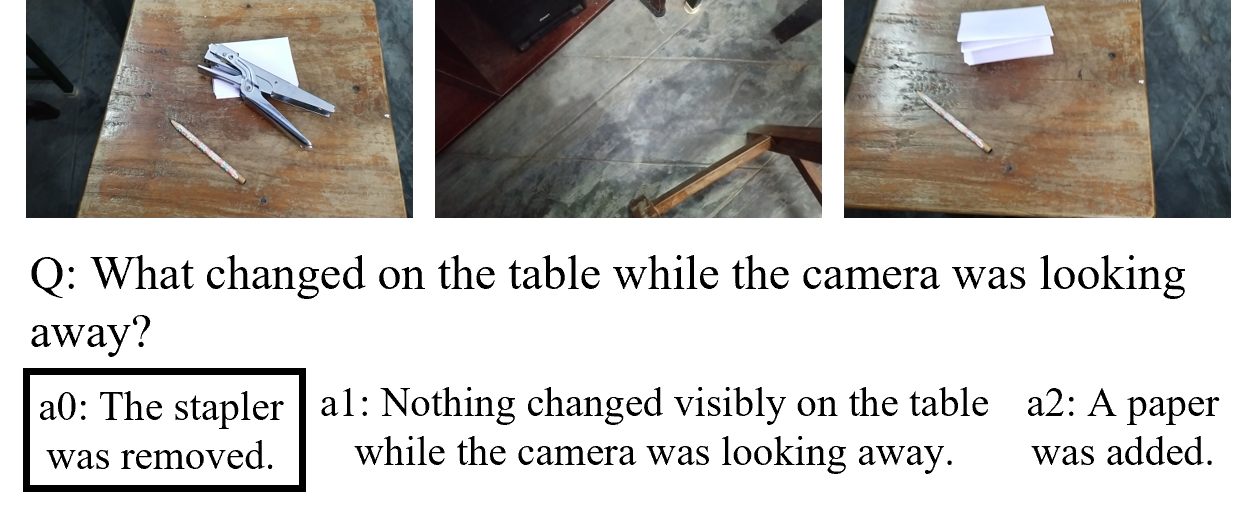}
  \caption{Example of a QA pair from Perception Test in the Explanatory Memory category. The correct answer is in the box.}
  \label{fig:pt_example_4}
\end{figure}

Examples of QA pairs from the Perception Test are shown in Figures ~\ref{fig:pt_example_1}-~\ref{fig:pt_example_4}.

\subsection{Video-MME}

Video-MME was developed to address the lack of diversity in video MCQA benchmarks. It consists of 900 YouTube videos ranging from 11 seconds to 1 hour in length. The dataset covers 6 primary visual domains with 30 subfields and 12 task types including questions on temporal and spatial perception, object and action reasoning, OCR, and counting problems. All QA pairs and distractors are manually created with three questions per video. Additionally, the dataset features videos in languages other than English. The videos additionally are supplied with subtitles. 

The released test split has an imbalanced option distribution, so we equalized it to properly assess option bias.

\begin{figure}[H]
  \includegraphics[width=\columnwidth]{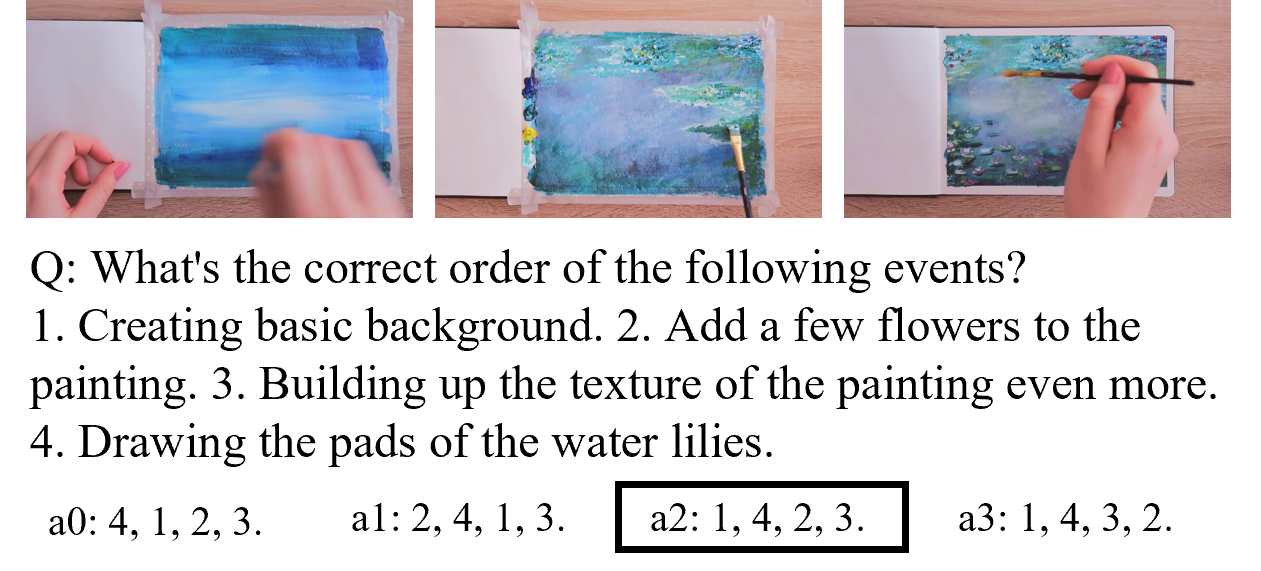}
  \caption{Example of a QA pair from Video-MME in the Temporal Reasoning category. The correct answer is in the box.}
  \label{fig:mme_example_1}
\end{figure}

\begin{figure}[H]
  \includegraphics[width=\columnwidth]{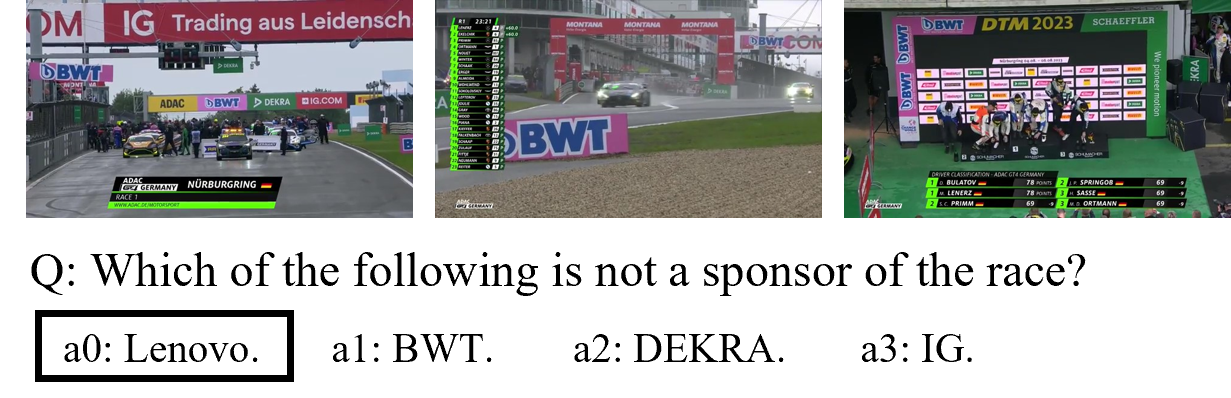}
  \caption{Example of a QA pair from Video-MME in the OCR Problems category. The correct answer is in the box.}
  \label{fig:mme_example_2}
\end{figure}

\begin{figure}[H]
  \includegraphics[width=\columnwidth]{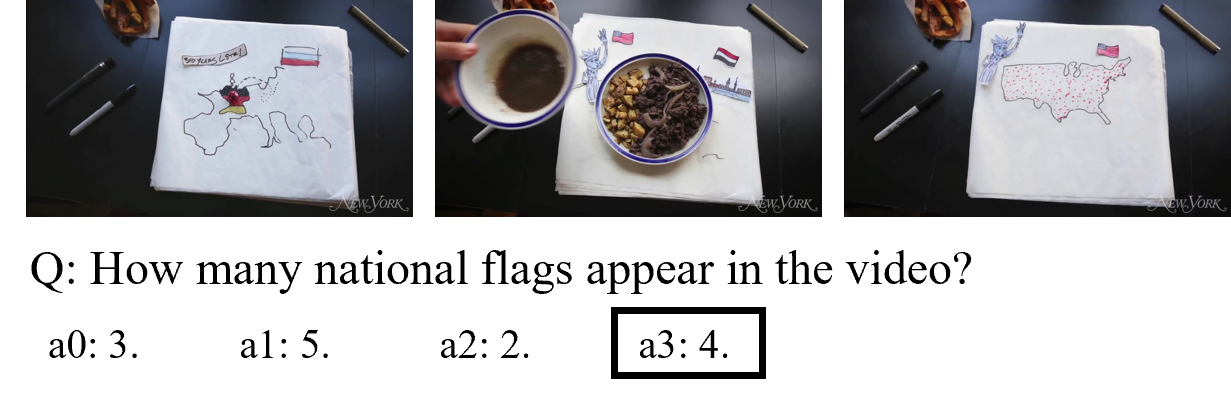}
  \caption{Example of a QA pair from Video-MME in the Counting Problem category. The correct answer is in the box.}
  \label{fig:mme_example_3}
\end{figure}

\begin{figure}[H]
  \includegraphics[width=\columnwidth]{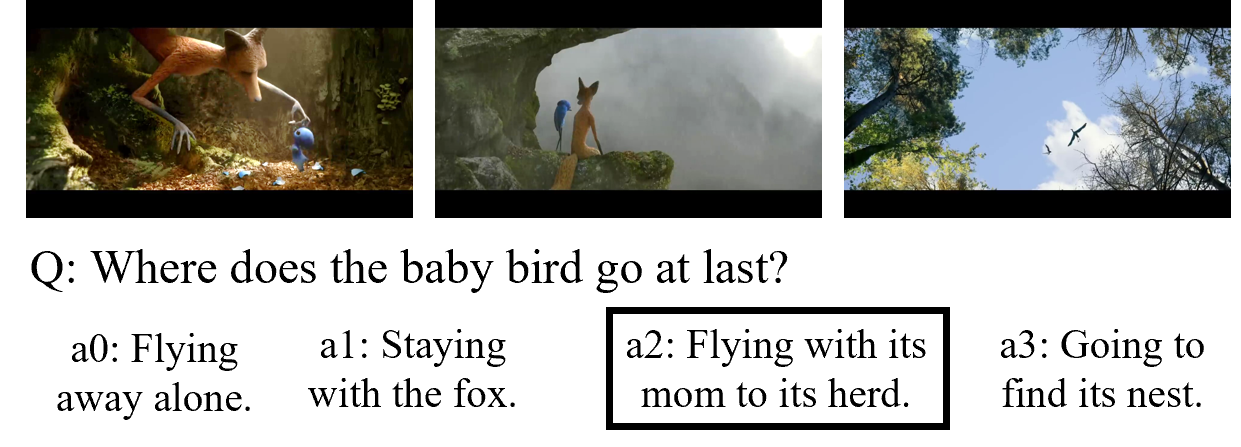}
  \caption{Example of a QA pair from Video-MME in the Spatial Perception category. The correct answer is in the box.}
  \label{fig:mme_example_4}
\end{figure}

Examples are shown in Figures ~\ref{fig:mme_example_1}-~\ref{fig:mme_example_4}.

\section{Models' Settings and Prompts}
\label{sec:app_models}

\subsection{Video-LLaMA}

Video-LLaMA aligns a BLIP-2-based visual encoder \cite{blip2} with LLaMA \cite{Touvron2023LLaMAOA}. It incorporates audio signals using Imagebind \cite{girdhar2023imagebind}. Both the video and audio Q-formers are trained on the Webvid-2M dataset \cite{Bain21} to align the modalities. LLaMA is fine-tuned with visual instructional data.

Video-LLaMA is designed as a conversational model and was likely not specifically trained on MCQA, but rather on open-ended VQA datasets to generate captions, descriptions, and explanatory answers. Consequently, throughout our experiments with different prompt options, it tends to engage in dialogue: it repeats the question, responds with the general phrase "Hello, how can I help you today?", describes the video content, or even asks follow-up questions such as "What do you think about the video?" rather than providing a direct answer from a predefined list of options or option IDs.\footnote{It is noteworthy that the model often uses the following template to explain its answer choice: "The ID of the correct answer: $a_i$. Explanation:~\ldots"} Making it elicit a clear and concise answer option is quite challenging.

To overcome this, we experimented with various prompts, temperature settings, and token limits, and developed the following strategy:

\textbf{System prompt:} 
\begin{quote}
\footnotesize
\texttt{Analyse the video, choose the correct answer in multiple choice question answering. Give the ID of the correct answer option from the given list of options. Be concise.}
\end{quote}

\textbf{User prompt:} 
\begin{quote}
\footnotesize
\texttt{\textless question\textgreater\ Options: \textless options\textgreater. The ID of the correct answer:}
\end{quote}

We set the \textit{number of beams} to 2 and the \textit{temperature} to 0.8. Additionally, we allowed the model up to 30 attempts to produce an answer that includes an answer ID. We used the 7B parameter model due to resource constraints. For answering, the model selects eight uniformly sampled frames.

\subsection{Video-LLaVA}

Video-LLaVA \citep{lin2023video} uses the pre-trained CLIP ViT-L/14 vision encoder \cite{Radford2021LearningTV} and the Vicuna language model \cite{vicuna2023}. Visual features from the encoder are projected into the same-dimensional space as the textual embeddings via a linear projector.

Video-LLaVA effectively delivers concise answers as just a single option based on a short, general prompt. However, in some cases, the model outputs as an answer option only "A" or responds with "yes" or "no" to a closed question.

For this model, we used the following prompt:

\begin{quote}
\footnotesize
\texttt{\textless video\_tag\textgreater\
Question: \textless question\textgreater\ Options: \textless options\textgreater\
The ID of the correct answer is: Respond with only the correct answer ID. ASSISTANT:}
\end{quote}

Notably, on the completely unfamiliar datasets Perception Test and Video-MME adding an empty option increases the number of responses that cannot be parsed as an answer ID (although this increase is statistically insignificant). Combined with the observation that the model tends to disregard options presented later in the prompt, this suggests that the length of the question affects the model's performance, and tokens closer to the end of the prompt have a lower probability.

The model also selects eight frames from the video by uniform sampling. To conserve computational resources, we used Video-LLaVA-7B with half-precision set by Torch's autocast. The temperature is equal to 1.

\subsection{SeViLA}

SeViLA adopts a selfchaining
approach with BLIP-2 \cite{blip2} for
keyframe localization and answering question. Its Localizer uniformly samples 32
frames and selects the 4 most relevant key ones, while the  Answerer generates responses based
on those frames. The Localizer and Answerer both
have Flan-T5 \cite{Chung2022ScalingIL} for LLM.

SeViLA does not have any unanswered or unparsable responses because it was specifically tailored for the multiple-choice VQA task: internally, it maps any answer option IDs to \{A, B, C, D, E\}. However, it is limited to observing at most six answer options. In a side experiment, we shifted all answers to options G, H, I, J, K leaving the first six options empty, and obtained identical results to our other experimental setting, \textit{Empty Answers}. This indicates that the model is completely blind to options beyond F and is likely overfitted to the options \{A, B, C, D, E\}.

\begin{figure*}[t]
    \includegraphics[width=0.32\linewidth]{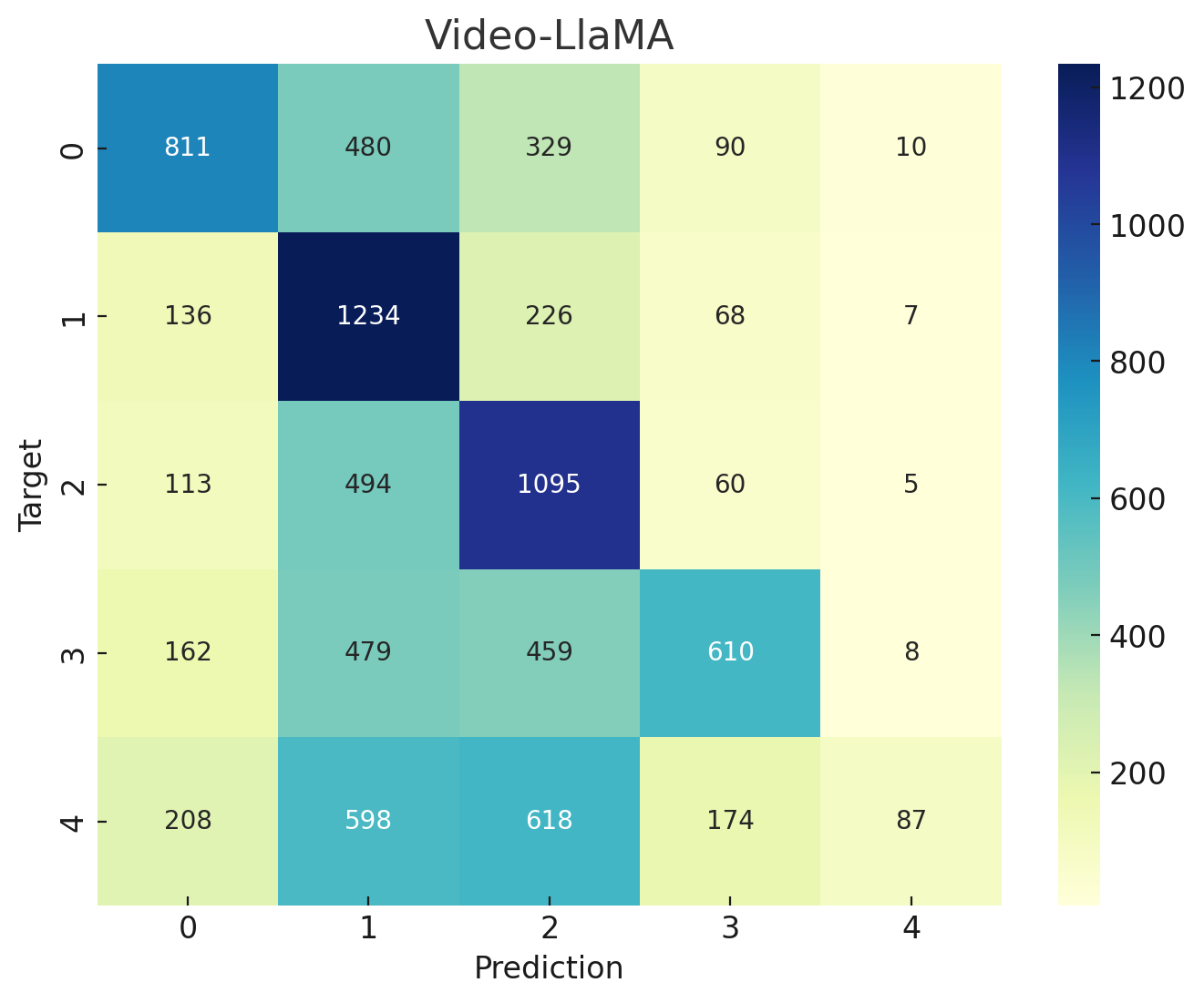}
    \hfill
    \includegraphics[width=0.32\linewidth]{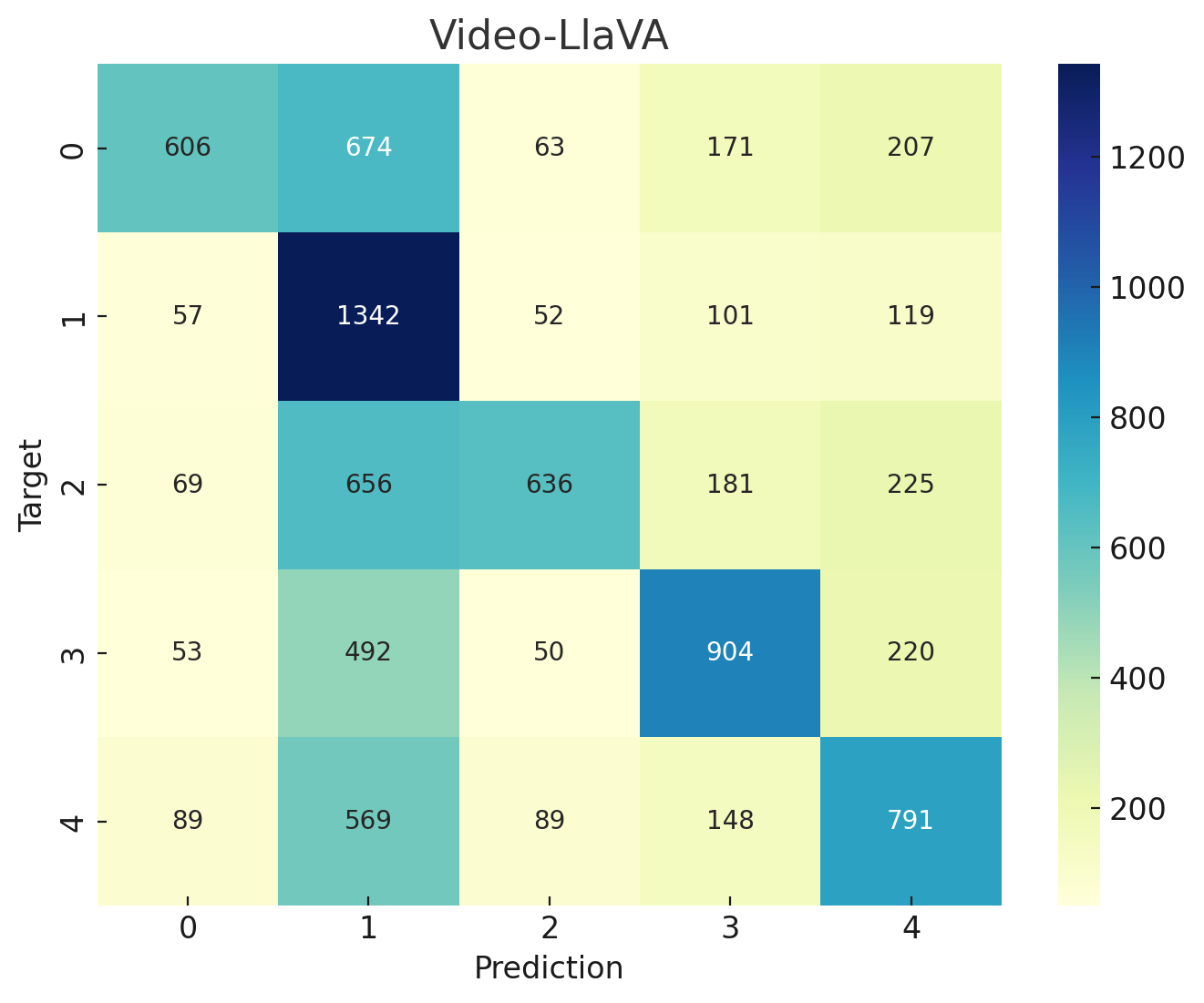}
    \hfill
    \includegraphics[width=0.32\linewidth]{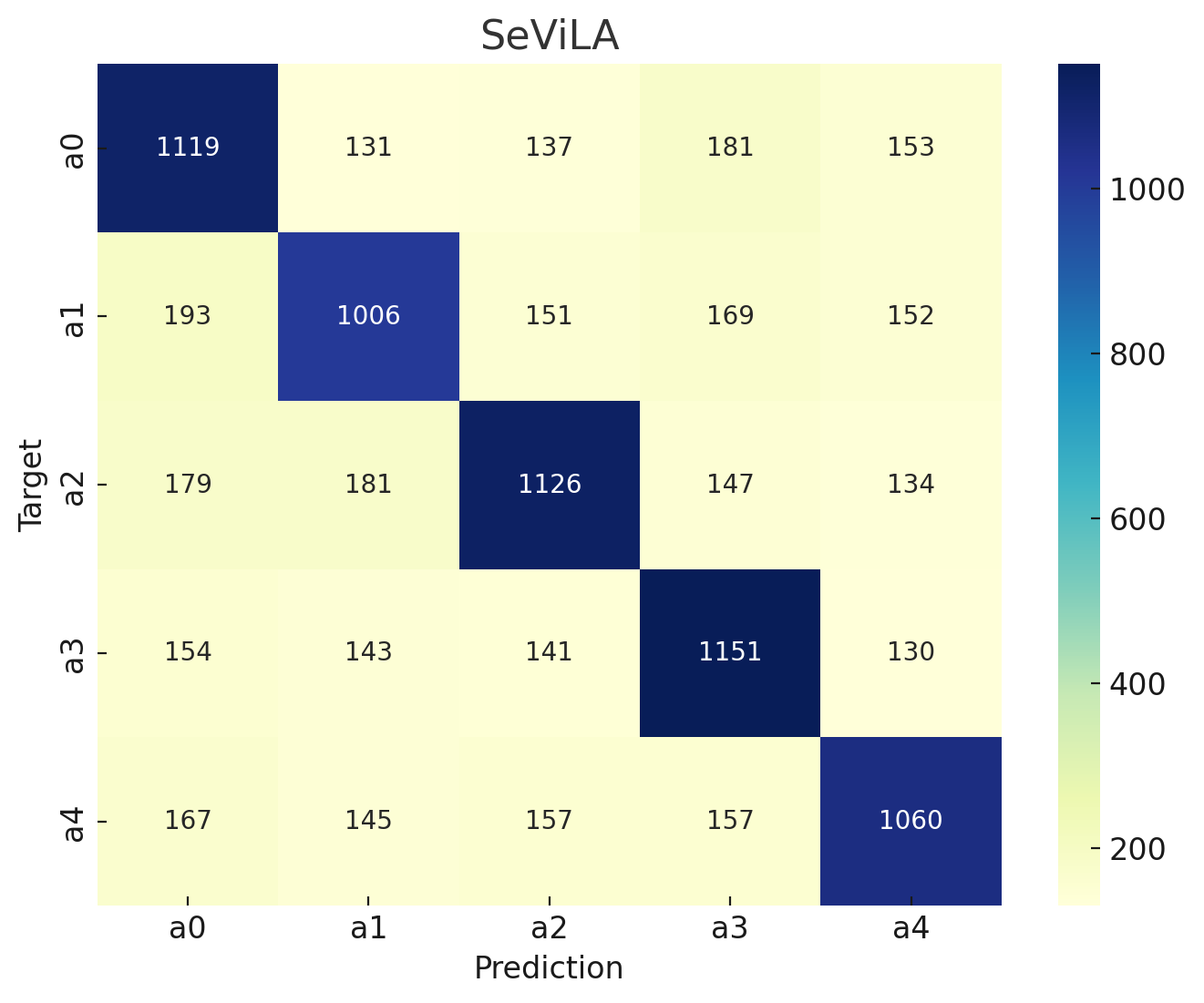}
    \hfill
    
    \caption{All Models' Confusion Matrices for NeXT-QA.}
    \label{fig:cm_nextqa}
\end{figure*}

\begin{figure*}[t]
    \includegraphics[width=0.32\linewidth]{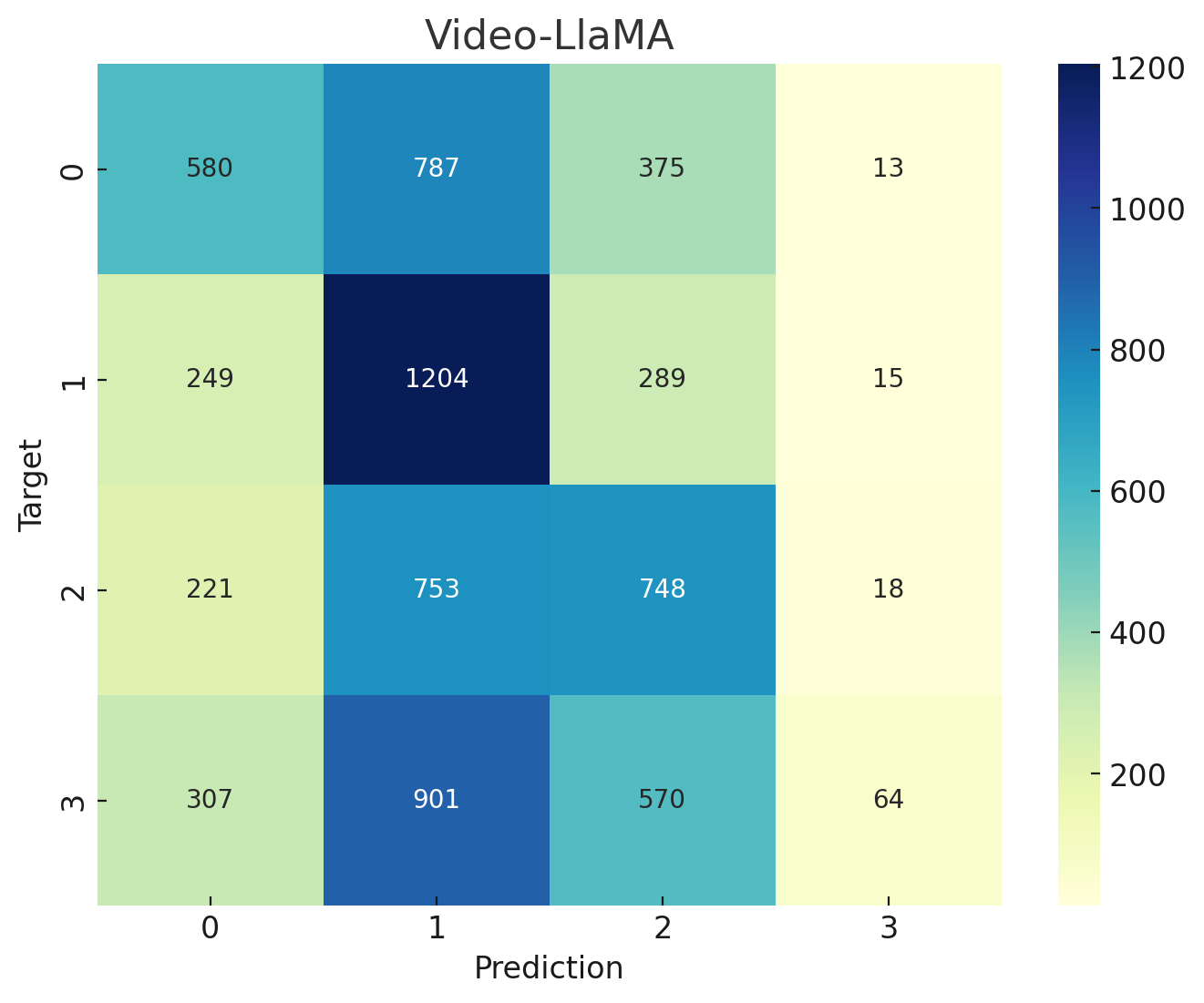}
    \hfill
    \includegraphics[width=0.32\linewidth]{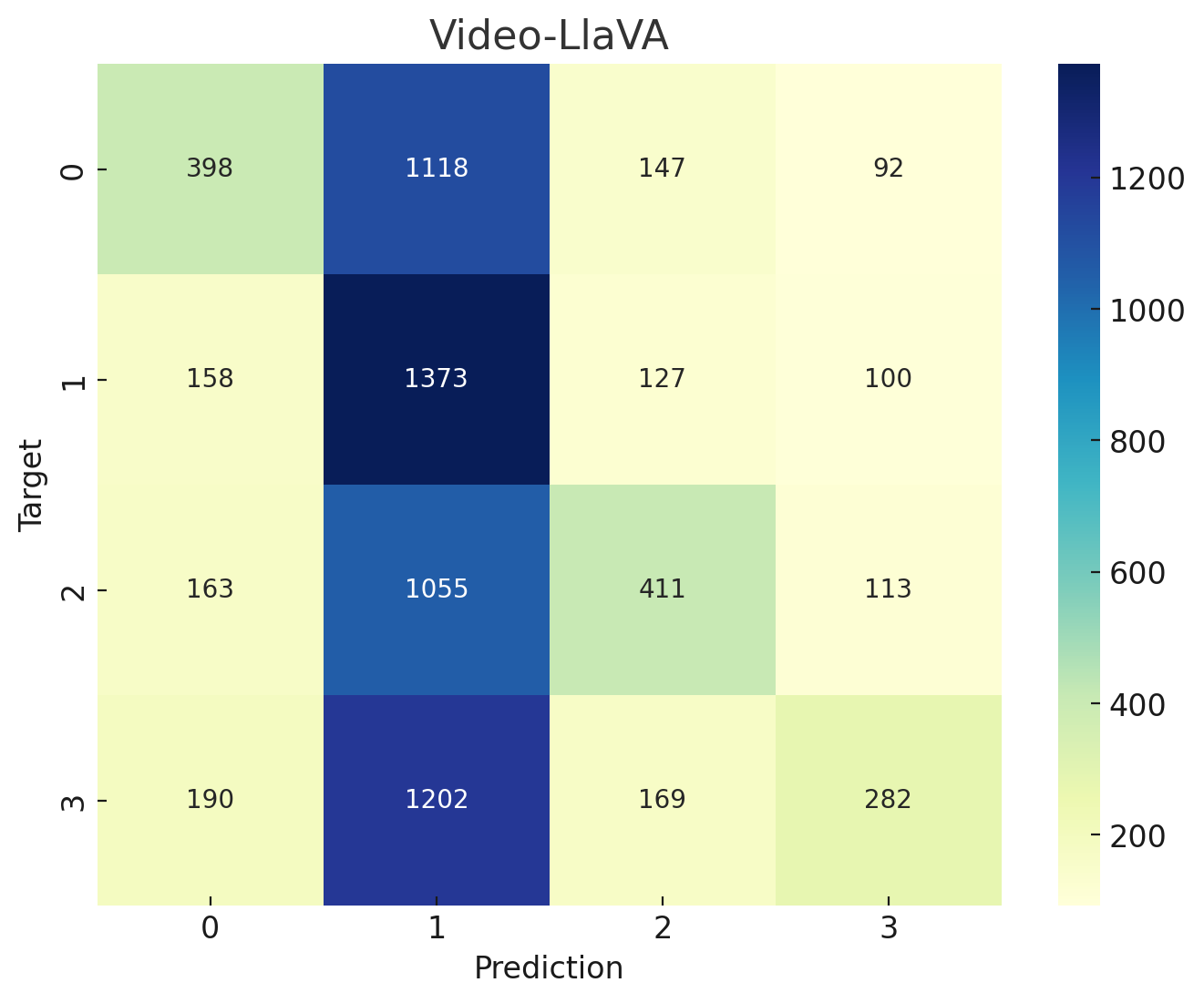}
    \hfill
    \includegraphics[width=0.32\linewidth]{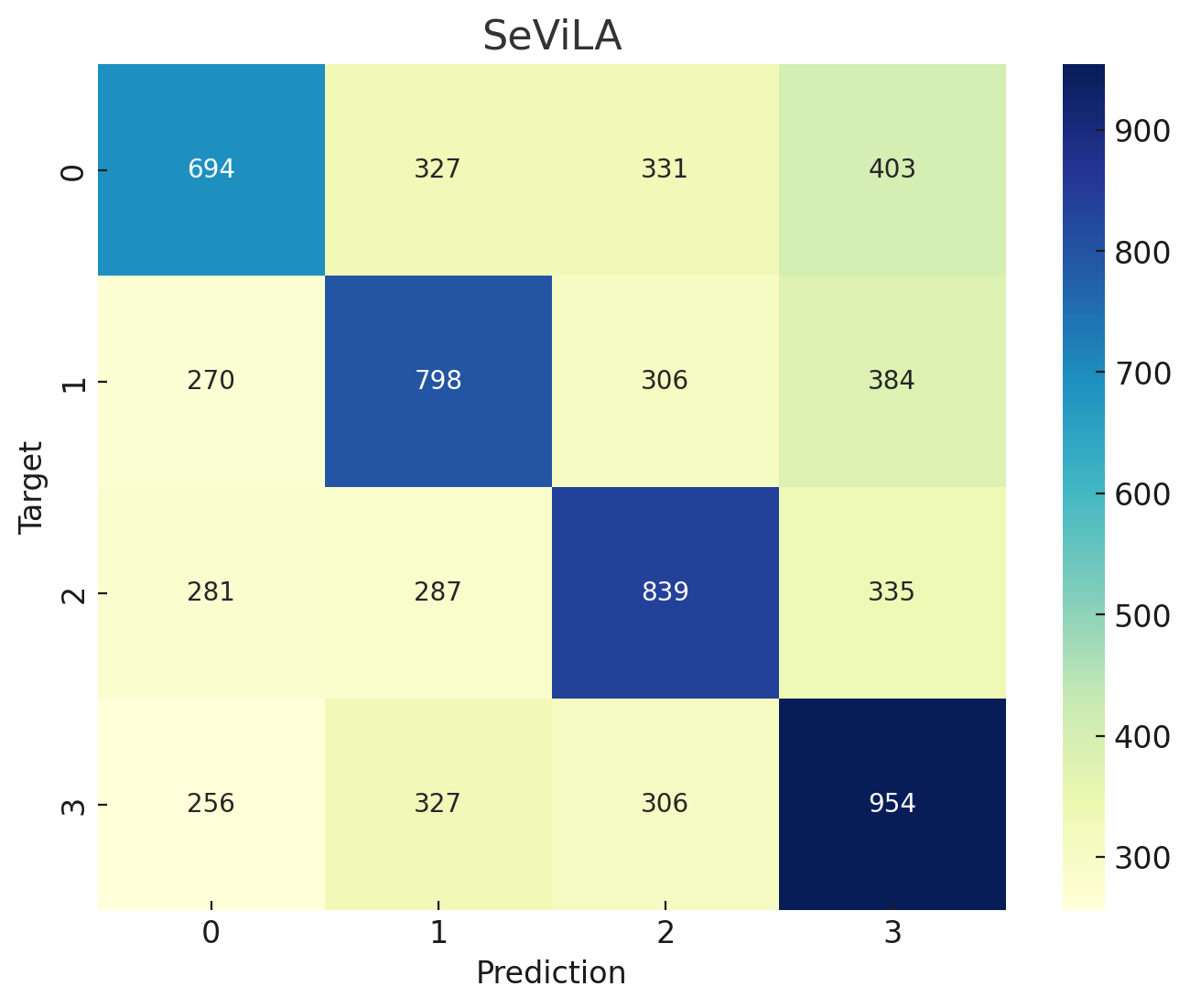}
    \hfill
    
    \caption{All Models' Confusion Matrices for STAR.}
    \label{fig:cm_star}
\end{figure*}

Furthermore, the shape of the confusion matrices for NExT-QA and STAR in Figures~\ref{fig:cm_nextqa} and~\ref{fig:cm_star} shows an unusually accurate understanding of these datasets, unlike those for Perception Test and Video-MME. Since NExT-QA and STAR were released before SeViLA, the model might have been fine-tuned on them. On the other hand, this suggests that the positional bias in this model is much less pronounced.

Another vulnerability of SeViLA is that it uses \textit{argmax} calculated on logit probabilities. However, logits without numerical stabilization can produce \texttt{NaN} values when activated via softmax. In such cases, according to \textit{argmax}, the model's answer defaults to $a_0$. Among the settings relevant for debiasing, such behavior was observed in 22 cases on NExT-QA.

For this model, we used the original prompts:
\textbf{Localizer prompt (when the model selects 32 frames):} 
\begin{quote}
\footnotesize
\texttt{Does the information within the frame provide necessary details to accurately answer the given question?}
\end{quote}

\textbf{Answerer prompt (when the model selects the answer based on 4 key frames):} 
\begin{quote}
\footnotesize
\texttt{Considering information in frames, select the correct answer from the options.}
\end{quote}

\begin{figure*}[t]
    \includegraphics[width=0.32\linewidth]{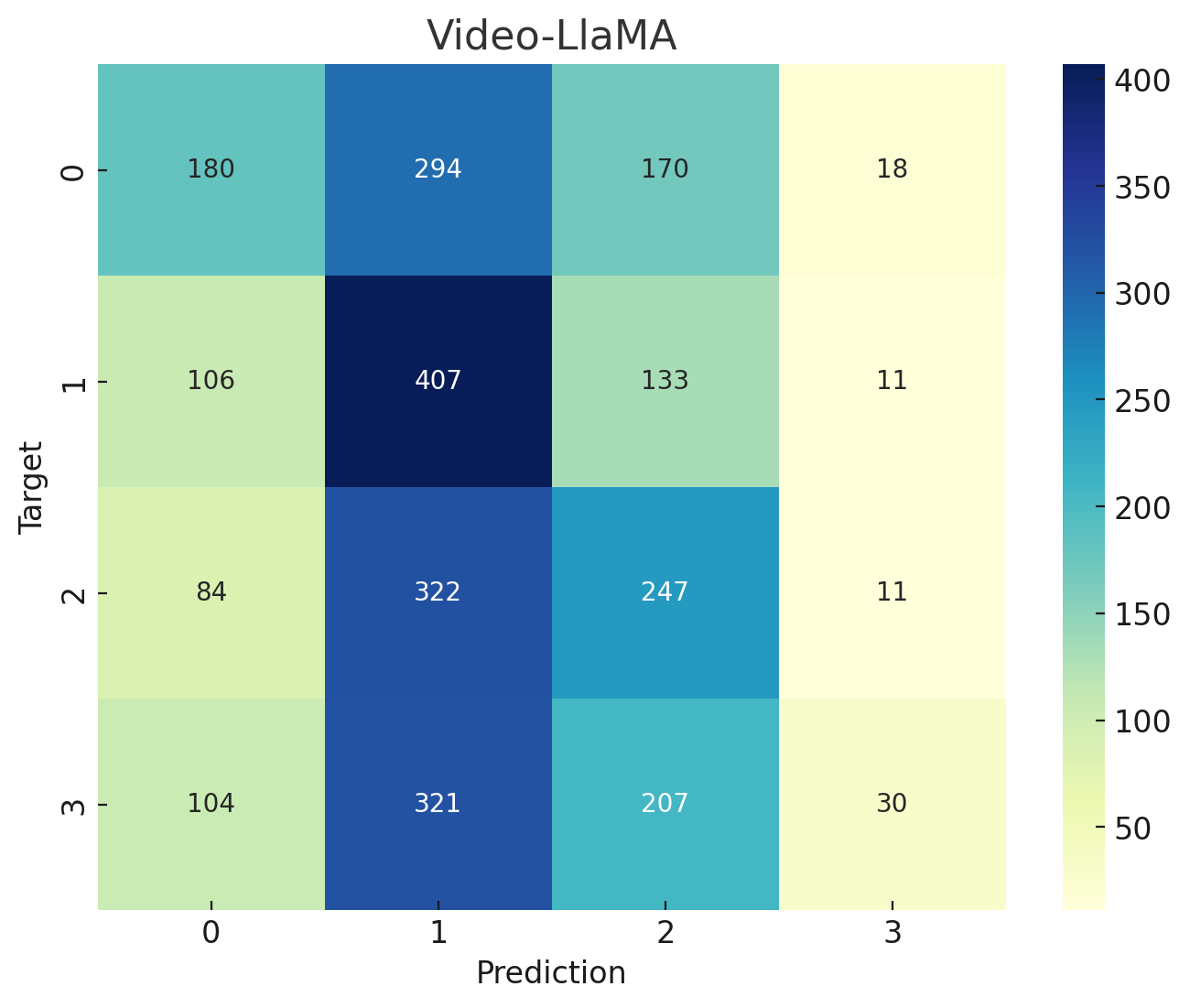}
    \hfill
    \includegraphics[width=0.32\linewidth]{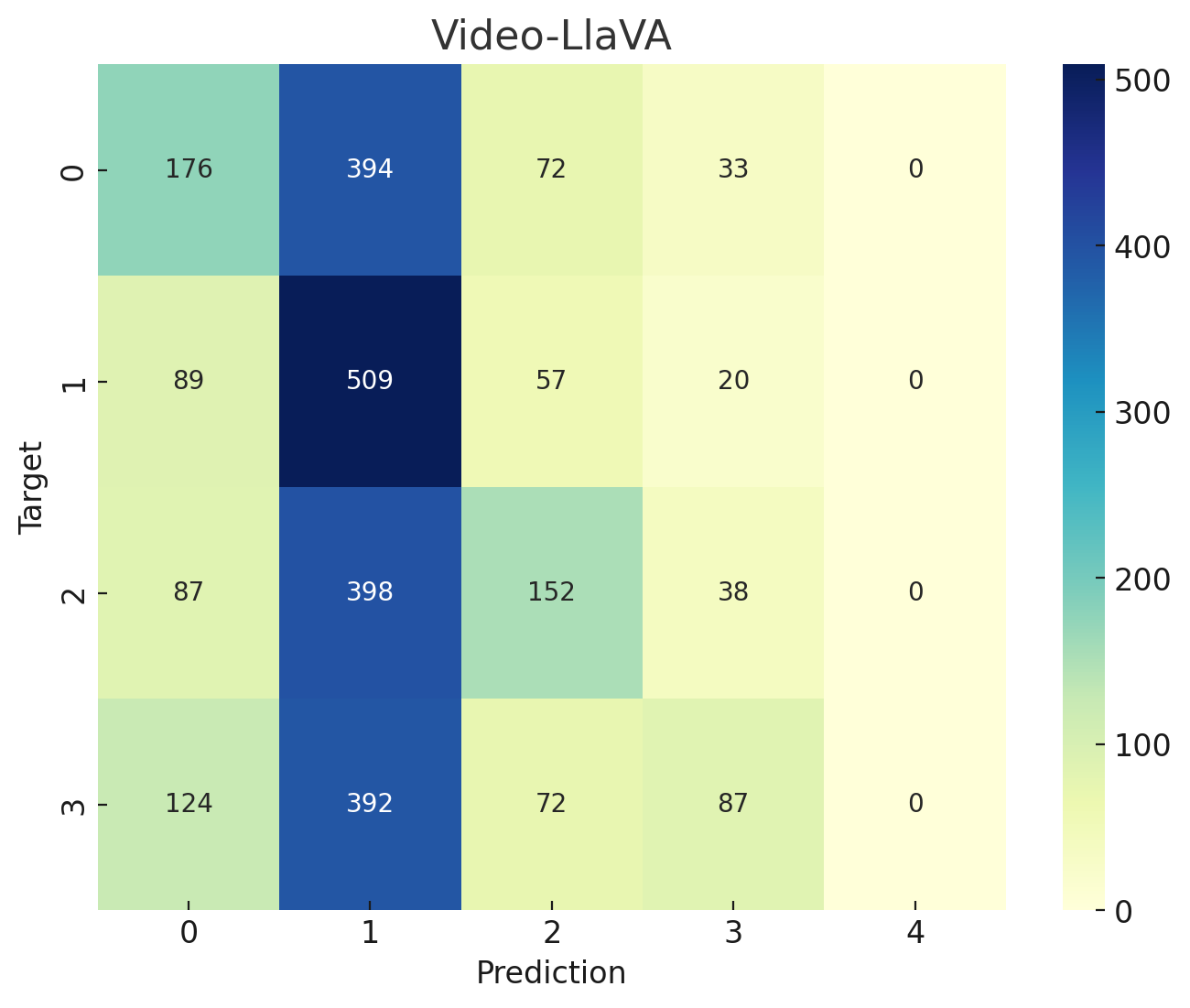}
    \hfill
    \includegraphics[width=0.32\linewidth]{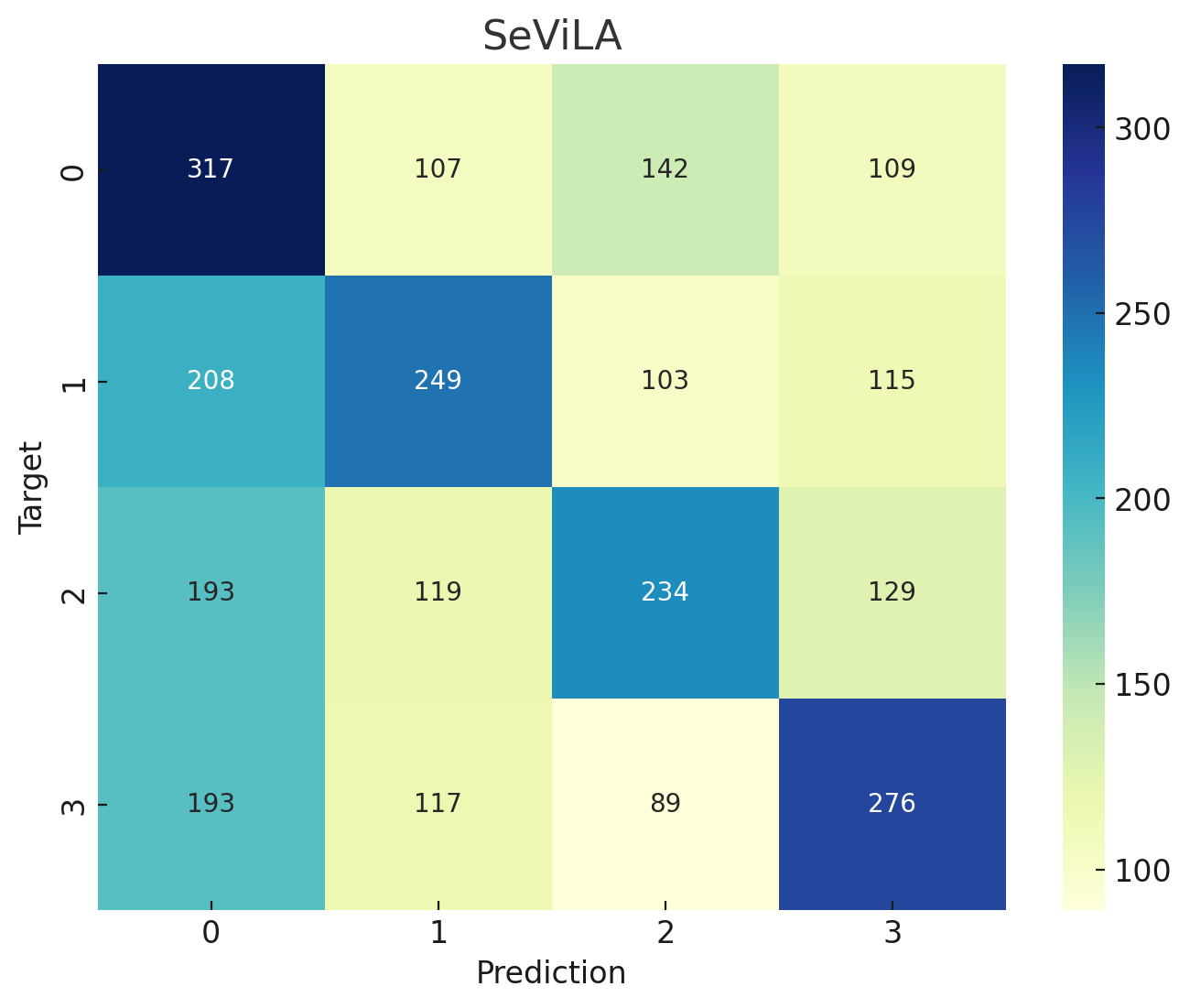}
    \hfill
    
    \caption{All Models' Confusion Matrices for Video-MME.}
    \label{fig:cm_mme}
\end{figure*}

The confusion matrices for all models in the default (unmodified) setting in Figures ~\ref{fig:cm_nextqa}–\ref{fig:cm_mme} provide a quick insight into positional bias: darker blue cells indicate a greater accumulation of answers.

\section{Models' Performance on All Settings}
\label{sec:performance}

\subsection{Experimental Settings Examples}

As the modifications to the video component do not affect the questions and answers, Table~\ref{tab:json_example} presents examples of adjustments applied only to the question-answer pairs, along with their original versions. The example provided for the Video-MME dataset illustrates analogous changes implemented across all four datasets.

\subsection{Models' Performance on the Settings}

Tables~\ref{tab:videollama_nextqa_full} through~\ref{tab:sevilavideomme} present the distribution of selections across answer options and the accuracy (where applicable) for all settings, along with the target distributions. Notably, as soon as the correct answer aligns with the model's biased option, the performance boosts.

\begin{figure*}[t]
\centering
  \includegraphics[width=\textwidth]{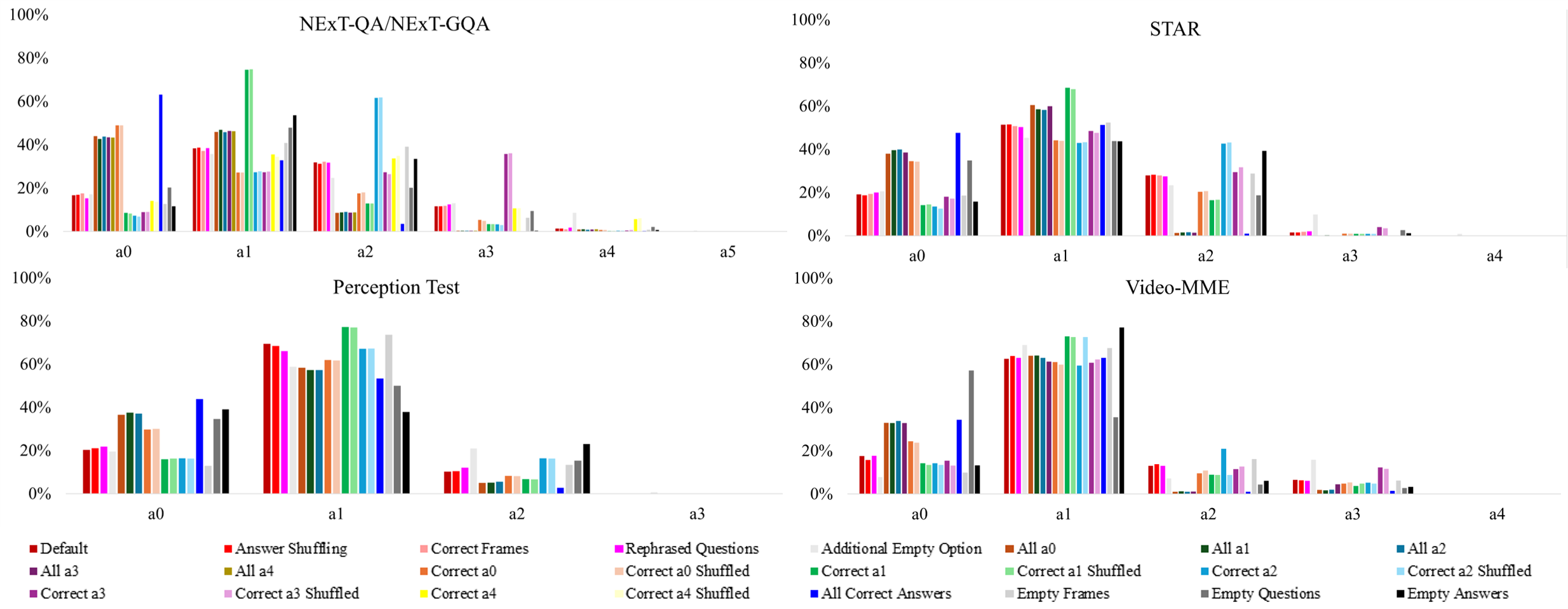}
  \caption{Video-Llama option distribution for all settings of NeXT-QA, NExT-GQA, STAR, Perception Test and Video-MME. Correct a\textsubscript{i}, Correct a\textsubscript{i}\ Shuffled and All a\textsubscript{i} represent Correct Answer, Correct Answer with Shuffling and All Correct Answers in each option, respectively.}
  \label{fig:distr_videollama_full}
\end{figure*}

\begin{figure*}[t]
\centering
  \includegraphics[width=\textwidth]{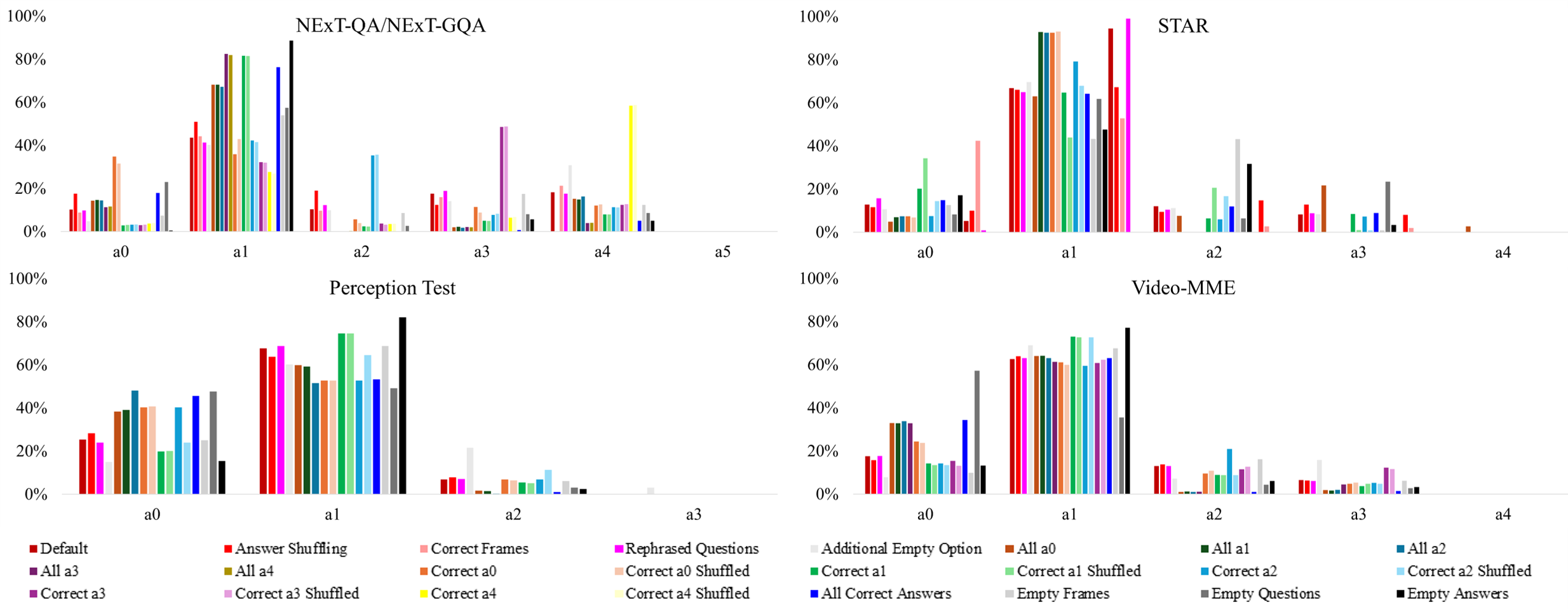}
  \caption{Video-Llava option distribution for all settings of NeXT-QA, NExT-GQA, STAR, Perception Test and Video-MME. Correct a\textsubscript{i}, Correct a\textsubscript{i}\ Shuffled and All a\textsubscript{i} represent Correct Answer, Correct Answer with Shuffling and All Correct Answers in each option, respectively.}
  \label{fig:distr_videollava_full}
\end{figure*}

\begin{figure*}[t]
\centering
  \includegraphics[width=\textwidth]{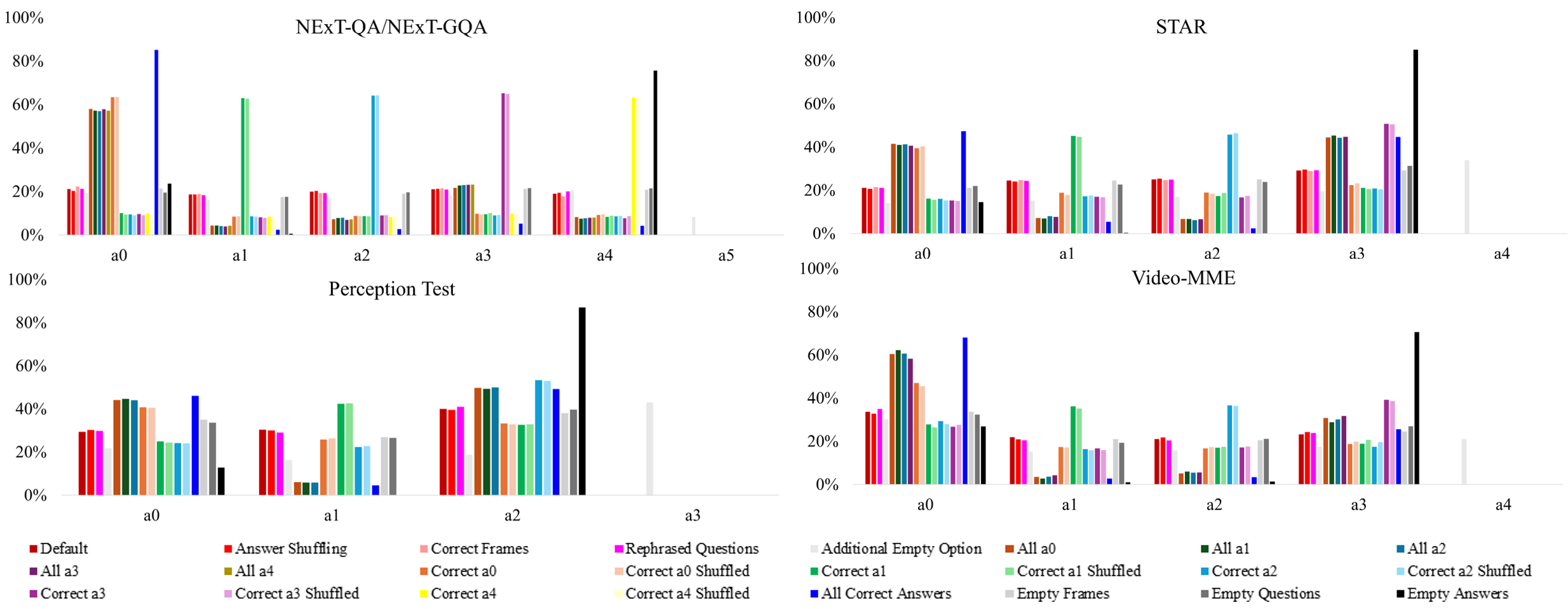}
  \caption{SeViLA option distribution for all settings of NeXT-QA, NExT-GQA, STAR, Perception Test and Video-MME. Correct a\textsubscript{i}, Correct a\textsubscript{i}\ Shuffled and All a\textsubscript{i} represent Correct Answer, Correct Answer with Shuffling and All Correct Answers in each option, respectively.}
  \label{fig:distr_sevila_full}
\end{figure*}

The full data on distribution answers in each option is given in Figures~\ref{fig:distr_videollama_full} to~\ref{fig:distr_sevila_full}. 

\section{Debiasing Algorithms and Results}
\label{sec:algorithms}

The BOLD algorithm \ref{alg:BOLD} describes estimating and adjusting for prior probabilities derived from decomposed inputs of ill-defined tasks. The Weighted\_BOLD algorithm ~\ref{alg:Weighted_BOLD} is an extension of Algorithm ~\ref{alg:BOLD}, this method introduces weighting to the prior estimation to enhance debiasing by better capturing critical priors.

We fixed the random seed to 1 and conducted a series of experiments varying the weights and the sampling parameter \( k \). All results including the default values are presented in Tables ~\ref{tab:25}–\ref{tab:100}. 

When considering only positive weights (\( w_i \geq 0 \)), we assume that the debiasing vector is oriented correctly and use the weights to fine-tune the magnitude of its decomposed constituents. Therefore, positive weights generally enhance improvements and aggravate deteriorations. Specifically, in this case, all the primary debiasing metrics improve along with \textit{Recall\_std}, albeit at different rates across models and datasets.

When we optimize a single bias metric, i. e. \textit{Recall\_std}, by allowing negative weights, we search the latent subspace for the most optimal vector to improve that specific metric. However, this optimization may not correlate with an improvement in the overall bias, as it focuses solely on that metric. Consequently, using negative weights can lead to deterioration in other bias and performance metrics. This observation underscores the importance of evaluating bias using multiple metrics.

We provide the results for both cases, when weights are constrained to \( 0 \leq w_i \leq 1 \) and when \( |w_i| \leq 1 \), as we believe that the observation of the direction of the bias vectors is crucial for understanding the overall debiasing effect.

\clearpage
\begin{table*}[ht]
\centering
\caption{Questions and answers modifications example from MME dataset. Correct answer is highlighted in bold in the ``Default'' modification type.}
\begin{tabular}{@{}p{4cm} p{5.5cm} p{6cm}@{}}
\toprule
\textbf{Modification Type} & \textbf{Question} & \textbf{Answers} \\
\midrule
Default & How many national flags appear in the video? & (a0) 3. (a1) 5. (a2) 2. \textbf{(a3) 4.} \\
Empty answers & Default question & (a0) -- (a1) -- (a2) -- (a3) -- \\
Answer shuffling & Default question & (a0) 4. (a1) 5. (a2) 2. (a3) 3. \\
Additional empty option & Default question & (a0) 3. (a1) 5. (a2) 2. (a3) 4. (a4) -- \\
Correct answer in each option & Default question & (a0) 4. (a1) 4. (a2) 4. (a3) 4. \\
All identical answers (a0) & Default question & (a0) 3. (a1) 3. (a2) 3. (a3) 3. \\
All identical answers (a1) & Default question & (a0) 5. (a1) 5. (a2) 5. (a3) 5. \\
All identical answers (a2) & Default question & (a0) 2. (a1) 2. (a2) 2. (a3) 2. \\
All identical answers (a3) & Default question & (a0) 4. (a1) 4. (a2) 4. (a3) 4. \\
Correct answer in (a0) & Default question & (a0) 4. (a1) 5. (a2) 2. (a3) 3. \\
Correct answer in (a1) & Default question & (a0) 3. (a1) 4. (a2) 2. (a3) 5. \\
Correct answer in (a2) & Default question & (a0) 3. (a1) 5. (a2) 4. (a3) 2. \\
Correct answer in (a3) & Default question & (a0) 3. (a1) 5. (a2) 2. (a3) 4. \\
Correct answer (a0) with shuffling & Default question & (a0) 4. (a1) 5. (a2) 3. (a3) 2. \\
Correct answer (a1) with shuffling & Default question & (a0) 5. (a1) 4. (a2) 2. (a3) 3. \\
Correct answer (a2) with shuffling & Default question & (a0) 3. (a1) 2. (a2) 4. (a3) 5. \\
Correct answer (a3) with shuffling & Default question & (a0) 5. (a1) 3. (a2) 2. (a3) 4. \\
Rephrased question & What is the total number of national flags that appear in the video? & (a0) 3. (a1) 5. (a2) 2. (a3) 4. \\
Empty questions & -- & (a0) 3. (a1) 5. (a2) 2. (a3) 4. \\

\bottomrule
\end{tabular}
\label{tab:json_example}
\end{table*}

\clearpage

\begin{figure*}[t]
    \includegraphics[width=0.32\linewidth]{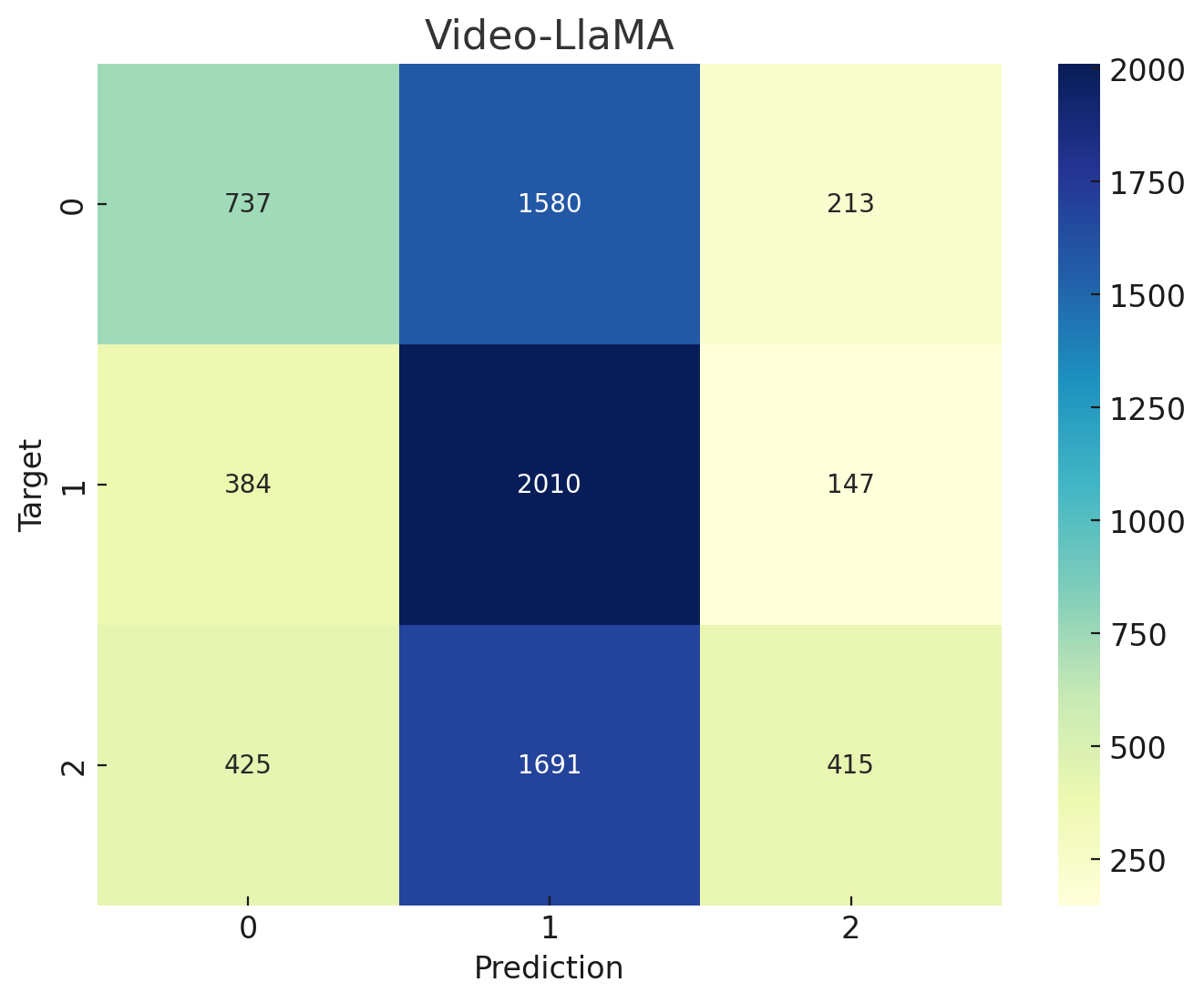}
    \hfill
    \includegraphics[width=0.32\linewidth]{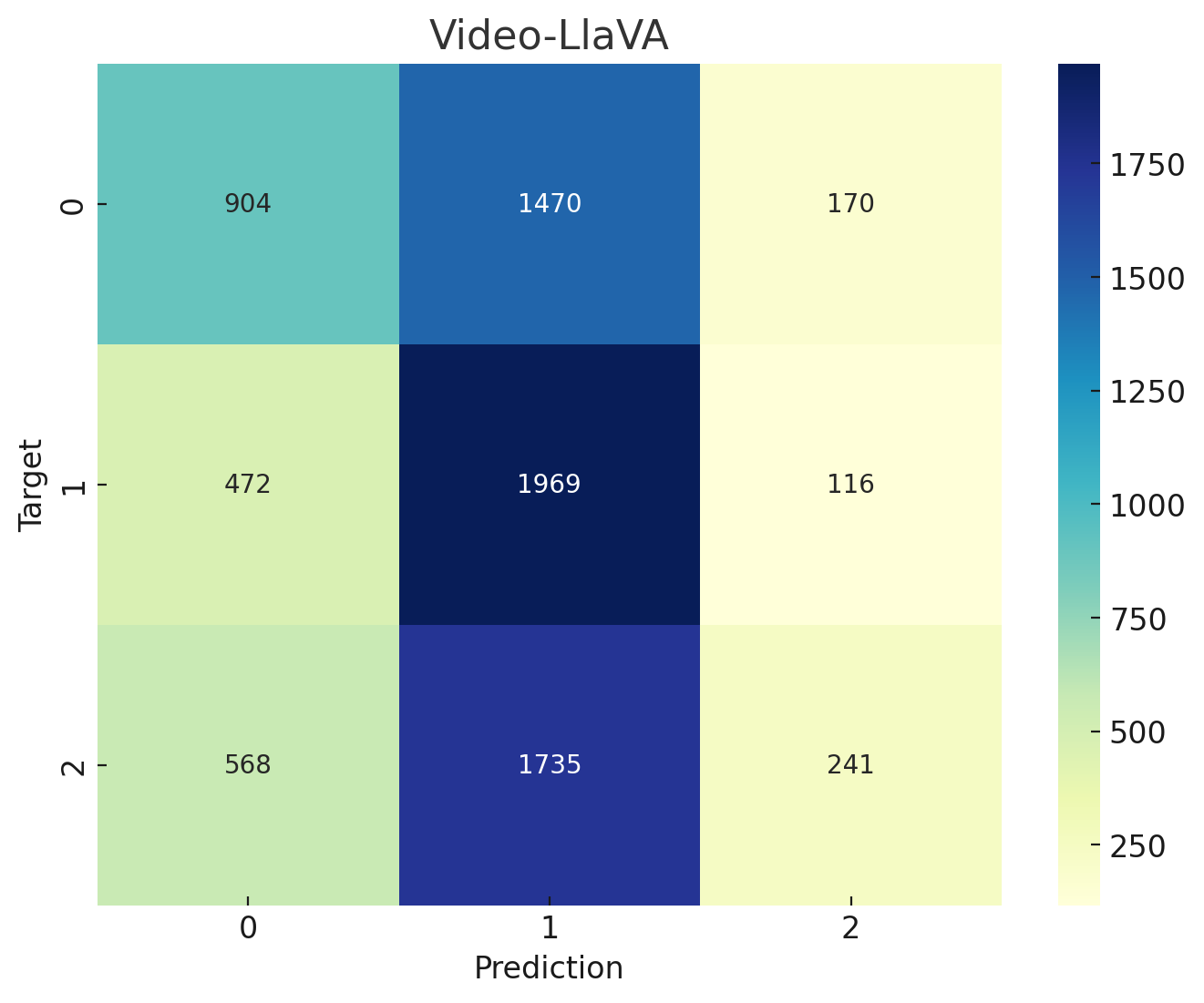}
    \hfill
    \includegraphics[width=0.32\linewidth]{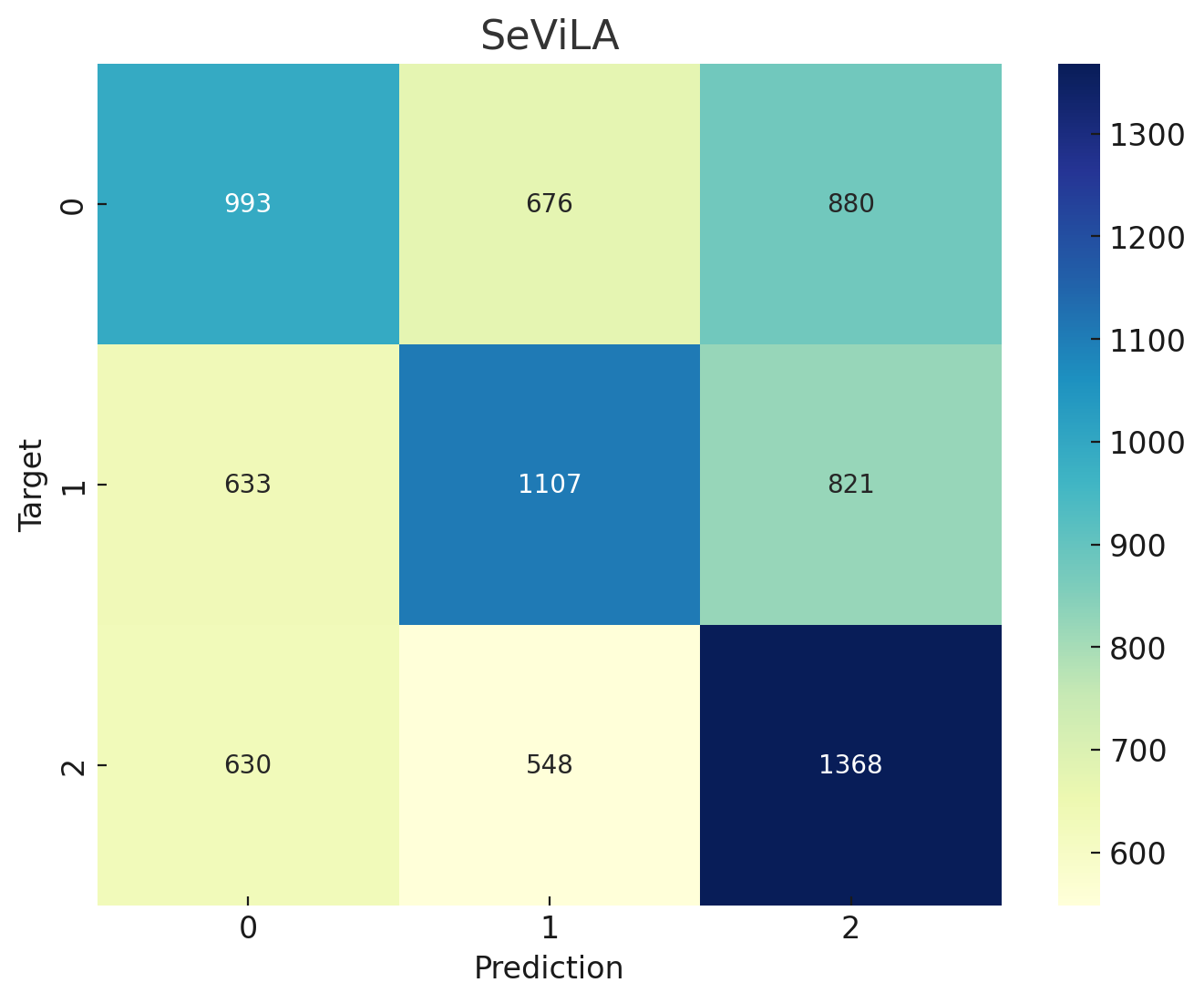}
    \hfill
    
    \caption{All Models' Confusion Matrices for Perception Test.}
    \label{fig:cm_pt}
\end{figure*}

\begin{table*}[ht]
\centering
\footnotesize
\caption{Video-LLaMA Performance on All Settings of NExT-QA (including NExT-GQA). Each cell shows the count of selections for each option (\textit{a0}--\textit{a5}), along with the percentage in parentheses. The Accuracy column represents the number of correct answers where applicable. The N/A column indicates the number of cases where the model did not output an answer option after 30 attempts.}
\label{tab:videollama_nextqa_full}
\resizebox{\textwidth}{!}{%
\begin{tabular}{l|rrrrrr|r|r}
\hline
\textbf{Setting} & \textbf{a0} & \textbf{a1} & \textbf{a2} & \textbf{a3} & \textbf{a4} & \textbf{a5} & \textbf{N/A} & \textbf{Accuracy} \\
\hline
Target & 1,721 (20.10\%) & 1,671 (19.51\%) & 1,767 (20.63\%) & 1,719 (20.07\%) & 1,686 (19.69\%) & -- & 0 (0.00\%) & -- \\
Default & 1,430 (16.70\%) & 3,285 (38.37\%) & 2,727 (31.85\%) & 1,002 (11.70\%) & 117 (1.37\%) & -- & 3 (0.04\%) & 3,837 (44.82\%) \\
Answer Shuffling & 1,443 (16.97\%) & 3,295 (38.75\%) & 2,654 (31.21\%) & 993 (11.68\%) & 119 (1.40\%) & -- & 60 (0.70\%) & 3,799 (44.67\%) \\
Correct Frames & 869 (17.61\%) & 1,836 (37.20\%) & 1,588 (32.18\%) & 587 (11.89\%) & 55 (1.11\%) & -- & 27 (0.54\%) & 2,155 (43.67\%) \\
Rephrased Questions & 1,305 (15.36\%) & 3,270 (38.50\%) & 2,704 (31.83\%) & 1,062 (12.50\%) & 153 (1.80\%) & -- & 70 (0.82\%) & 3,762 (44.29\%) \\
Additional Empty Option & 1,432 (17.17\%) & 2,987 (35.82\%) & 2,067 (24.78\%) & 1,090 (13.07\%) & 725 (8.69\%) & 39 (0.47\%) & 224 (2.62\%) & 4,086 (48.99\%) \\
All a1 & 3,537 (42.71\%) & 3,891 (46.98\%) & 737 (8.90\%) & 32 (0.39\%) & 85 (1.03\%) & -- & 282 (3.29\%) & -- \\
All a2 & 3,614 (43.82\%) & 3,780 (45.83\%) & 748 (9.07\%) & 32 (0.39\%) & 73 (0.89\%) & -- & 317 (3.70\%) & -- \\
All a3 & 3,585 (43.43\%) & 3,831 (46.41\%) & 722 (8.75\%) & 33 (0.40\%) & 83 (1.01\%) & -- & 310 (3.62\%) & -- \\
All a4 & 3,590 (43.35\%) & 3,837 (46.33\%) & 733 (8.85\%) & 34 (0.41\%) & 87 (1.05\%) & -- & 283 (3.30\%) & -- \\
Correct a0 & 4,136 (48.98\%) & 2,303 (27.27\%) & 1,489 (17.63\%) & 453 (5.36\%) & 64 (0.76\%) & -- & 119 (1.39\%) & 4,136 (48.98\%) \\
Correct a0 Shuffled & 4,136 (49.00\%) & 2,304 (27.30\%) & 1,525 (18.07\%) & 415 (4.92\%) & 61 (0.72\%) & -- & 123 (1.44\%) & 4,136 (49.00\%) \\
Correct a1 & 730 (8.61\%) & 6,323 (74.61\%) & 1,102 (13.00\%) & 292 (3.45\%) & 28 (0.33\%) & -- & 89 (1.04\%) & 6,323 (74.61\%) \\
Correct a1 Shuffled & 704 (8.30\%) & 6,348 (74.88\%) & 1,098 (12.95\%) & 297 (3.50\%) & 31 (0.37\%) & -- & 86 (1.00\%) & 6,348 (74.88\%) \\
Correct a2 & 622 (7.32\%) & 2,325 (27.35\%) & 5,240 (61.64\%) & 280 (3.29\%) & 34 (0.40\%) & -- & 63 (0.74\%) & 5,240 (61.64\%) \\
Correct a2 Shuffled & 583 (6.88\%) & 2,356 (27.81\%) & 5,240 (61.84\%) & 254 (3.00\%) & 40 (0.47\%) & -- & 91 (1.06\%) & 5,240 (61.84\%) \\
Correct a3 & 762 (8.99\%) & 2,315 (27.32\%) & 2,321 (27.39\%) & 3,029 (35.74\%) & 47 (0.55\%) & -- & 90 (1.05\%) & 3,029 (35.74\%) \\
Correct a3 Shuffled & 766 (9.07\%) & 2,335 (27.65\%) & 2,235 (26.47\%) & 3,049 (36.11\%) & 59 (0.70\%) & -- & 120 (1.40\%) & 3,049 (36.11\%) \\
Correct a4 & 1,194 (14.11\%) & 3,014 (35.63\%) & 2,857 (33.77\%) & 909 (10.74\%) & 486 (5.74\%) & -- & 104 (1.21\%) & 486 (5.74\%) \\
Correct a4 Shuffled & 1,164 (13.77\%) & 2,912 (34.45\%) & 2,949 (34.88\%) & 912 (10.79\%) & 517 (6.12\%) & -- & 110 (1.28\%) & 517 (6.12\%) \\
All Correct Answers & 5,247 (63.17\%) & 2,730 (32.87\%) & 295 (3.55\%) & 9 (0.11\%) & 25 (0.30\%) & -- & 258 (3.01\%) & -- \\
Empty Frames & 1,092 (12.77\%) & 3,494 (40.87\%) & 3,344 (39.11\%) & 544 (6.36\%) & 76 (0.89\%) & -- & 14 (0.16\%) & 2,620 (30.64\%) \\
Empty Questions & 1,686 (20.31\%) & 3,975 (47.89\%) & 1,676 (20.19\%) & 786 (9.47\%) & 178 (2.14\%) & -- & 263 (3.07\%) & 3,082 (37.13\%) \\
Empty Answers & 987 (11.66\%) & 4,543 (53.68\%) & 2,837 (33.52\%) & 37 (0.44\%) & 59 (0.70\%) & -- & 101 (1.18\%) & -- \\
\hline
\end{tabular}
}
\end{table*}

\begin{table*}[ht]
\centering
\footnotesize
\caption{Video-LLaMA Performance on All Settings of STAR. Each cell shows the count of selections for each option (\textit{a0}--\textit{a4}), along with the percentage in parentheses. The Accuracy column represents the number of correct answers where applicable. The N/A column indicates the number of cases where the model did not output an answer option after 30 attempts.}
\label{tab:videollama_star}
\resizebox{\textwidth}{!}{%
\begin{tabular}{l|rrrrr|r|r}
\hline
\textbf{Setting} & \textbf{a0} & \textbf{a1} & \textbf{a2} & \textbf{a3} & \textbf{a4} & \textbf{N/A} & \textbf{Accuracy} \\
\hline
Target & 1,755 (24.73\%) & 1,758 (24.77\%) & 1,742 (24.54\%) & 1,843 (25.97\%) & -- & 0 (0.00\%) & -- \\
Default & 1,357 (19.13\%) & 3,645 (51.38\%) & 1,982 (27.94\%) & 110 (1.55\%) & -- & 4 (0.06\%) & 2,596 (36.59\%) \\
Answer Shuffling & 1,255 (18.72\%) & 3,458 (51.57\%) & 1,893 (28.23\%) & 99 (1.48\%) & -- & 393 (5.54\%) & 2,467 (36.79\%) \\
Correct Frames & 1,322 (19.43\%) & 3,455 (50.79\%) & 1,899 (27.91\%) & 127 (1.87\%) & -- & 295 (4.16\%) & 2,524 (37.10\%) \\
Rephrased Questions & 1,376 (20.03\%) & 3,462 (50.39\%) & 1,891 (27.53\%) & 141 (2.05\%) & -- & 228 (3.21\%) & 2,520 (36.68\%) \\
Additional Empty Option & 1,314 (20.59\%) & 2,892 (45.31\%) & 1,494 (23.41\%) & 630 (9.87\%) & 52 (0.81\%) & 716 (10.09\%) & 2,528 (39.61\%) \\
All a0 & 2,484 (38.04\%) & 3,949 (60.47\%) & 88 (1.35\%) & 9 (0.14\%) & -- & 568 (8.00\%) & -- \\
All a1 & 2,695 (39.68\%) & 3,976 (58.54\%) & 104 (1.53\%) & 17 (0.25\%) & -- & 306 (4.31\%) & -- \\
All a2 & 2,711 (39.93\%) & 3,957 (58.28\%) & 113 (1.66\%) & 9 (0.13\%) & -- & 308 (4.34\%) & -- \\
All a3 & 2,512 (38.52\%) & 3,909 (59.94\%) & 94 (1.44\%) & 7 (0.11\%) & -- & 576 (8.11\%) & -- \\
Correct a0 & 2,340 (34.53\%) & 2,996 (44.21\%) & 1,378 (20.33\%) & 63 (0.93\%) & -- & 321 (4.52\%) & 2,340 (34.53\%) \\
Correct a0 Shuffled & 2,313 (34.36\%) & 2,960 (43.97\%) & 1,392 (20.68\%) & 67 (1.00\%) & -- & 366 (5.16\%) & 2,313 (34.36\%) \\
Correct a1 & 966 (14.16\%) & 4,677 (68.54\%) & 1,124 (16.47\%) & 57 (0.84\%) & -- & 274 (3.86\%) & 4,677 (68.54\%) \\
Correct a1 Shuffled & 985 (14.48\%) & 4,622 (67.93\%) & 1,136 (16.70\%) & 61 (0.90\%) & -- & 294 (4.14\%) & 4,622 (67.93\%) \\
Correct a2 & 918 (13.53\%) & 2,913 (42.93\%) & 2,898 (42.71\%) & 56 (0.83\%) & -- & 313 (4.41\%) & 2,898 (42.71\%) \\
Correct a2 Shuffled & 844 (12.57\%) & 2,908 (43.32\%) & 2,901 (43.21\%) & 60 (0.89\%) & -- & 385 (5.42\%) & 2,901 (43.21\%) \\
Correct a3 & 1,226 (18.03\%) & 3,298 (48.50\%) & 2,002 (29.44\%) & 274 (4.03\%) & -- & 298 (4.20\%) & 274 (4.03\%) \\
Correct a3 Shuffled & 1,153 (17.21\%) & 3,190 (47.60\%) & 2,127 (31.74\%) & 231 (3.45\%) & -- & 397 (5.59\%) & 231 (3.45\%) \\
All Correct Answers & 3,129 (47.60\%) & 3,376 (51.35\%) & 63 (0.96\%) & 6 (0.09\%) & -- & 524 (7.38\%) & -- \\
Empty Frames & 1,323 (18.64\%) & 3,721 (52.42\%) & 2,047 (28.84\%) & 7 (0.10\%) & -- & 0 (0.00\%) & 1,854 (26.12\%) \\
Empty Questions & 2,365 (34.84\%) & 2,973 (43.80\%) & 1,275 (18.78\%) & 175 (2.58\%) & -- & 310 (4.37\%) & 2,427 (35.75\%) \\
Empty Answers & 1,121 (15.84\%) & 3,093 (43.72\%) & 2,778 (39.27\%) & 83 (1.17\%) & -- & 23 (0.32\%) & -- \\
\hline
\end{tabular}%
}
\end{table*}

\begin{table*}[ht]
\centering
\footnotesize
\caption{Video-LLaMA Performance on All Settings of Perception Test. Each cell shows the count of selections for each option (\textit{a0}--\textit{a3}), along with the percentage in parentheses. The Accuracy column represents the number of correct answers where applicable. The N/A column indicates the number of cases where the model did not output an answer option after 30 attempts.}
\label{tab:videollama_perception}
\resizebox{\textwidth}{!}{%
\begin{tabular}{l|rrrr|r|r}
\hline
\textbf{Setting} & \textbf{a0} & \textbf{a1} & \textbf{a2} & \textbf{a3} & \textbf{N/A} & \textbf{Accuracy} \\
\hline
Target & 2,549 (33.29\%) & 2,561 (33.45\%) & 2,546 (33.25\%) & -- & 0 (0.00\%) & -- \\
Default & 1,546 (20.34\%) & 5,281 (69.47\%) & 775 (10.19\%) & -- & 54 (0.71\%) & 3,162 (41.59\%) \\
Answer Shuffling & 1,593 (21.10\%) & 5,165 (68.42\%) & 791 (10.48\%) & -- & 107 (1.40\%) & 3,146 (41.67\%) \\
Rephrased Questions & 1,652 (21.83\%) & 5,004 (66.13\%) & 911 (12.04\%) & -- & 89 (1.16\%) & 3,161 (41.77\%) \\
Additional Empty Option & 1,448 (19.56\%) & 4,356 (58.86\%) & 1,549 (20.93\%) & 48 (0.65\%) & 255 (3.33\%) & 3,004 (40.59\%) \\
All a0 & 2,768 (36.60\%) & 4,414 (58.36\%) & 381 (5.04\%) & -- & 93 (1.21\%) & -- \\
All a1 & 2,842 (37.56\%) & 4,334 (57.28\%) & 390 (5.15\%) & -- & 90 (1.18\%) & -- \\
All a2 & 2,806 (37.15\%) & 4,326 (57.28\%) & 421 (5.57\%) & -- & 103 (1.35\%) & -- \\
Correct a0 & 2,245 (29.76\%) & 4,675 (61.98\%) & 623 (8.26\%) & -- & 113 (1.48\%) & 2,245 (29.76\%) \\
Correct a0 Shuffled & 2,271 (30.12\%) & 4,654 (61.73\%) & 614 (8.14\%) & -- & 117 (1.53\%) & 2,271 (30.12\%) \\
Correct a1 & 1,204 (15.95\%) & 5,833 (77.29\%) & 510 (6.76\%) & -- & 109 (1.42\%) & 5,833 (77.29\%) \\
Correct a1 Shuffled & 1,233 (16.32\%) & 5,820 (77.06\%) & 500 (6.62\%) & -- & 103 (1.35\%) & 5,820 (77.06\%) \\
Correct a2 & 1,239 (16.40\%) & 5,077 (67.18\%) & 1,241 (16.42\%) & -- & 99 (1.29\%) & 1,241 (16.42\%) \\
Correct a2 Shuffled & 1,235 (16.36\%) & 5,082 (67.30\%) & 1,234 (16.34\%) & -- & 105 (1.37\%) & 1,234 (16.34\%) \\
All Correct Answers & 3,330 (43.87\%) & 4,053 (53.40\%) & 207 (2.73\%) & -- & 66 (0.86\%) & -- \\
Empty Frames & 990 (12.93\%) & 5,641 (73.68\%) & 1,025 (13.39\%) & -- & 0 (0.00\%) & 3,005 (39.25\%) \\
Empty Questions & 2,564 (34.61\%) & 3,707 (50.03\%) & 1,138 (15.36\%) & -- & 247 (3.23\%) & 3,021 (40.77\%) \\
Empty Answers & 2,969 (39.11\%) & 2,876 (37.88\%) & 1,747 (23.01\%) & -- & 64 (0.84\%) & -- \\
\hline
\end{tabular}%
}
\end{table*}

\begin{table*}[ht]
\centering
\footnotesize
\caption{Video-LLaMA Performance on All Settings of Video-MME. Each cell shows the count of selections for each option (\textit{a0}--\textit{a4}), along with the percentage in parentheses. The Accuracy column represents the number of correct answers where applicable. The N/A column indicates the number of cases where the model did not output an answer option after 30 attempts.}
\label{tab:videollama_videomme}
\resizebox{\textwidth}{!}{%
\begin{tabular}{l|rrrrr|r|r}
\hline
\textbf{Setting} & \textbf{a0} & \textbf{a1} & \textbf{a2} & \textbf{a3} & \textbf{a4} & \textbf{N/A} & \textbf{Accuracy} \\
\hline
Target & 675 (25.00\%) & 675 (25.00\%) & 675 (25.00\%) & 675 (25.00\%) & -- & 0 (0.00\%) & -- \\
Default & 474 (17.92\%) & 1,344 (50.81\%) & 757 (28.62\%) & 70 (2.65\%) & -- & 55 (2.04\%) & 864 (32.67\%) \\
Answer Shuffling & 451 (18.11\%) & 1,242 (49.86\%) & 733 (29.43\%) & 65 (2.61\%) & -- & 209 (7.74\%) & 785 (31.51\%) \\
Rephrased Questions & 1,367 (50.65\%) & 979 (36.27\%) & 198 (7.34\%) & 155 (5.74\%) & -- & 1 (0.04\%) & 873 (32.35\%) \\
Additional Empty Option & 391 (17.29\%) & 1,040 (45.98\%) & 624 (27.59\%) & 178 (7.87\%) & 29 (1.28\%) & 438 (16.22\%) & 746 (32.98\%) \\
All a0 & 1,042 (42.81\%) & 1,244 (51.11\%) & 120 (4.93\%) & 28 (1.15\%) & -- & 266 (9.85\%) & -- \\
All a1 & 1,027 (42.00\%) & 1,295 (52.97\%) & 91 (3.72\%) & 32 (1.31\%) & -- & 255 (9.44\%) & -- \\
All a2 & 1,061 (43.36\%) & 1,232 (50.35\%) & 116 (4.74\%) & 38 (1.55\%) & -- & 253 (9.37\%) & -- \\
All a3 & 1,001 (41.06\%) & 1,223 (50.16\%) & 149 (6.11\%) & 65 (2.67\%) & -- & 262 (9.70\%) & -- \\
Correct a0 & 678 (27.19\%) & 1,141 (45.75\%) & 624 (25.02\%) & 51 (2.04\%) & -- & 206 (7.63\%) & 678 (27.19\%) \\
Correct a0 Shuffled & 699 (27.90\%) & 1,085 (43.31\%) & 664 (26.51\%) & 57 (2.28\%) & -- & 195 (7.22\%) & 699 (27.90\%) \\
Correct a1 & 376 (14.98\%) & 1,532 (61.04\%) & 547 (21.79\%) & 55 (2.19\%) & -- & 190 (7.04\%) & 1,532 (61.04\%) \\
Correct a1 Shuffled & 344 (13.78\%) & 1,519 (60.83\%) & 584 (23.39\%) & 50 (2.00\%) & -- & 203 (7.52\%) & 1,519 (60.83\%) \\
Correct a2 & 366 (14.63\%) & 1,147 (45.86\%) & 948 (37.90\%) & 40 (1.60\%) & -- & 199 (7.37\%) & 948 (37.90\%) \\
Correct a2 Shuffled & 355 (14.27\%) & 1,144 (45.98\%) & 942 (37.86\%) & 47 (1.89\%) & -- & 212 (7.85\%) & 942 (37.86\%) \\
Correct a3 & 408 (16.47\%) & 1,178 (47.56\%) & 777 (31.37\%) & 114 (4.60\%) & -- & 223 (8.26\%) & 114 (4.60\%) \\
Correct a3 Shuffled & 386 (15.55\%) & 1,199 (48.31\%) & 786 (31.67\%) & 111 (4.47\%) & -- & 218 (8.07\%) & 111 (4.47\%) \\
All Correct Answers & 1,195 (46.46\%) & 1,226 (47.67\%) & 114 (4.43\%) & 37 (1.44\%) & -- & 128 (4.74\%) & -- \\
Empty Frames & 188 (6.98\%) & 1,612 (59.81\%) & 843 (31.28\%) & 52 (1.93\%) & -- & 5 (0.19\%) & 776 (28.79\%) \\
Empty Questions & 753 (29.89\%) & 1,180 (46.84\%) & 527 (20.92\%) & 59 (2.34\%) & -- & 181 (6.70\%) & 798 (31.68\%) \\
Empty Answers & 440 (17.07\%) & 725 (28.13\%) & 1,352 (52.46\%) & 60 (2.33\%) & -- & 123 (4.56\%) & -- \\
\hline
\end{tabular}%
}
\end{table*}

\begin{table*}[ht]
\centering
\footnotesize
\caption{Video-LLaVA Performance on All the Settings of NExT-QA (including NExT-GQA). Each cell shows the count of selections for each option (\textit{a0}--\textit{a5}) along with the percentage in parentheses. Accuracy represents the number of correct answers where applicable.}
\label{tab:videollama_nextqa_all}
\resizebox{\textwidth}{!}{%
\begin{tabular}{l|rrrrrr|r|r}
\hline
\textbf{Setting} & \textbf{a0} & \textbf{a1} & \textbf{a2} & \textbf{a3} & \textbf{a4} & \textbf{a5} & \textbf{N/A} & \textbf{Accuracy} \\
\hline
Target & 1,721 (20.10\%) & 1,671 (19.51\%) & 1,767 (20.63\%) & 1,719 (20.07\%) & 1,686 (19.69\%) & -- & 0 (0.00\%) & -- \\
Default & 874 (10.21\%) & 3,733 (43.59\%) & 890 (10.39\%) & 1,505 (17.57\%) & 1,562 (18.24\%) & -- & 0 (0.00\%) & 4,279 (49.96\%) \\
Answer Shuffling & 1,507 (17.60\%) & 4,366 (50.98\%) & 1,630 (19.03\%) & 1,061 (12.39\%) & 0 (0.00\%) & -- & 0 (0.00\%) & 3,653 (42.66\%) \\
Correct Frames & 440 (8.87\%) & 2,194 (44.22\%) & 482 (9.71\%) & 791 (15.94\%) & 1,055 (21.26\%) & -- & 0 (0.00\%) & 2,407 (48.51\%) \\
Rephrased Questions & 836 (9.79\%) & 3,531 (41.36\%) & 1,054 (12.34\%) & 1,613 (18.89\%) & 1,504 (17.62\%) & -- & 26 (0.30\%) & 4,287 (50.21\%) \\
Additional Empty Option & 418 (4.88\%) & 3,449 (40.27\%) & 847 (9.89\%) & 1,208 (14.11\%) & 2,641 (30.84\%) & 1 (0.01\%) & 0 (0.00\%) & 4,264 (49.79\%) \\
All a0 & 1,232 (14.39\%) & 5,839 (68.18\%) & 9 (0.11\%) & 175 (2.04\%) & 1,309 (15.28\%) & -- & 0 (0.00\%) & -- \\
All a1 & 1,254 (14.64\%) & 5,839 (68.18\%) & 9 (0.11\%) & 187 (2.18\%) & 1,275 (14.89\%) & -- & 0 (0.00\%) & -- \\
All a2 & 1,238 (14.46\%) & 5,755 (67.21\%) & 12 (0.14\%) & 158 (1.85\%) & 1,400 (16.35\%) & -- & 1 (0.01\%) & -- \\
All a3 & 966 (11.28\%) & 7,063 (82.47\%) & 14 (0.16\%) & 178 (2.08\%) & 343 (4.01\%) & -- & 0 (0.00\%) & -- \\
All a4 & 996 (11.63\%) & 7,021 (81.98\%) & 20 (0.23\%) & 176 (2.06\%) & 351 (4.10\%) & -- & 0 (0.00\%) & -- \\
Correct a0 & 2,966 (34.87\%) & 3,054 (35.91\%) & 484 (5.69\%) & 975 (11.46\%) & 1,026 (12.06\%) & -- & 59 (0.69\%) & 2,966 (34.87\%) \\
Correct a0 Shuffled & 2,702 (31.55\%) & 3,680 (42.97\%) & 341 (3.98\%) & 759 (8.86\%) & 1,082 (12.63\%) & -- & 0 (0.00\%) & 2,702 (31.55\%) \\
Correct a1 & 251 (2.93\%) & 6,993 (81.66\%) & 209 (2.44\%) & 431 (5.03\%) & 680 (7.94\%) & -- & 0 (0.00\%) & 6,993 (81.66\%) \\
Correct a1 Shuffled & 270 (3.15\%) & 6,981 (81.52\%) & 203 (2.37\%) & 424 (4.95\%) & 686 (8.01\%) & -- & 0 (0.00\%) & 6,981 (81.52\%) \\
Correct a2 & 274 (3.20\%) & 3,619 (42.26\%) & 3,032 (35.41\%) & 669 (7.81\%) & 969 (11.32\%) & -- & 1 (0.01\%) & 3,032 (35.41\%) \\
Correct a2 Shuffled & 282 (3.29\%) & 3,561 (41.58\%) & 3,056 (35.68\%) & 707 (8.26\%) & 958 (11.19\%) & -- & 0 (0.00\%) & 3,056 (35.68\%) \\
Correct a3 & 258 (3.01\%) & 2,765 (32.29\%) & 319 (3.73\%) & 4,161 (48.59\%) & 1,060 (12.38\%) & -- & 1 (0.01\%) & 4,161 (48.59\%) \\
Correct a3 Shuffled & 278 (3.25\%) & 2,746 (32.06\%) & 271 (3.16\%) & 4,182 (48.83\%) & 1,087 (12.69\%) & -- & 0 (0.00\%) & 4,182 (48.83\%) \\
Correct a4 & 324 (3.78\%) & 2,374 (27.72\%) & 300 (3.50\%) & 557 (6.50\%) & 5,009 (58.49\%) & -- & 0 (0.00\%) & 5,009 (58.49\%) \\
Correct a4 Shuffled & 336 (3.92\%) & 2,305 (26.91\%) & 313 (3.65\%) & 579 (6.76\%) & 5,031 (58.75\%) & -- & 0 (0.00\%) & 5,031 (58.75\%) \\
All Correct Answers & 1,536 (17.94\%) & 6,531 (76.27\%) & 4 (0.05\%) & 59 (0.69\%) & 433 (5.06\%) & -- & 1 (0.01\%) & -- \\
Empty Frames & 634 (7.40\%) & 4,625 (54.01\%) & 740 (8.64\%) & 1,499 (17.50\%) & 1,066 (12.45\%) & -- & 0 (0.00\%) & 3,025 (35.32\%) \\
Empty Questions & 1,976 (23.07\%) & 4,925 (57.51\%) & 228 (2.66\%) & 696 (8.13\%) & 739 (8.63\%) & -- & 0 (0.00\%) & 2,979 (34.79\%) \\
Empty Answers & 45 (0.53\%) & 7,590 (88.63\%) & 12 (0.14\%) & 484 (5.65\%) & 433 (5.06\%) & -- & 0 (0.00\%) & -- \\
\hline
\end{tabular}
}
\end{table*}

\begin{table*}[ht]
\centering
\footnotesize
\caption{Video-LLaVA Performance on All the Settings of STAR. Each cell shows the count of selections for each option (\textit{a0}--\textit{a4}) along with the percentage in parentheses. Accuracy represents the number of correct answers where applicable.}
\label{tab:videollava_star}
\resizebox{\textwidth}{!}{%
\begin{tabular}{l|rrrrr|r|r}
\hline
\textbf{Setting} & \textbf{a0} & \textbf{a1} & \textbf{a2} & \textbf{a3} & \textbf{a4} & \textbf{N/A} & \textbf{Accuracy} \\
\hline
Target & 1,755 (24.73\%) & 1,758 (24.77\%) & 1,742 (24.54\%) & 1,843 (25.97\%) & -- & 0 (0.00\%) & -- \\
Default & 909 (12.81\%) & 4,748 (66.89\%) & 854 (12.03\%) & 587 (8.27\%) & -- & 0 (0.00\%) & 2,464 (34.71\%) \\
Answer Shuffling & 825 (11.62\%) & 4,688 (66.05\%) & 675 (9.51\%) & 910 (12.82\%) & -- & 0 (0.00\%) & 2,521 (35.52\%) \\
Correct Frames & 1,115 (15.71\%) & 4,611 (64.96\%) & 743 (10.47\%) & 629 (8.86\%) & -- & 0 (0.00\%) & 1,881 (26.50\%) \\
Rephrased Questions & 759 (10.70\%) & 4,945 (69.70\%) & 799 (11.26\%) & 592 (8.34\%) & -- & 3 (0.04\%) & 2,455 (34.60\%) \\
Additional Empty Option & 346 (4.87\%) & 4,472 (63.00\%) & 545 (7.68\%) & 1,541 (21.71\%) & 194 (2.73\%) & 0 (0.00\%) & 2,490 (35.08\%) \\
All a0 & 499 (7.03\%) & 6,596 (92.93\%) & 1 (0.01\%) & 2 (0.03\%) & -- & 0 (0.00\%) & -- \\
All a1 & 525 (7.40\%) & 6,573 (92.60\%) & 0 (0.00\%) & 0 (0.00\%) & -- & 0 (0.00\%) & -- \\
All a2 & 523 (7.37\%) & 6,571 (92.58\%) & 1 (0.01\%) & 3 (0.04\%) & -- & 0 (0.00\%) & -- \\
All a3 & 485 (6.83\%) & 6,610 (93.12\%) & 2 (0.03\%) & 1 (0.01\%) & -- & 0 (0.00\%) & -- \\
Correct a0 & 1,438 (20.26\%) & 4,598 (64.78\%) & 459 (6.47\%) & 603 (8.50\%) & -- & 0 (0.00\%) & 1,438 (20.26\%) \\
Correct a0 Shuffled & 1,424 (20.06\%) & 4,578 (64.51\%) & 453 (6.38\%) & 642 (9.05\%) & -- & 1 (0.01\%) & 1,424 (20.06\%) \\
Correct a1 & 533 (7.51\%) & 5,621 (79.19\%) & 426 (6.00\%) & 518 (7.30\%) & -- & 0 (0.00\%) & 5,621 (79.19\%) \\
Correct a1 Shuffled & 566 (7.98\%) & 5,560 (78.34\%) & 449 (6.33\%) & 522 (7.36\%) & -- & 1 (0.01\%) & 5,560 (78.34\%) \\
Correct a2 & 1,055 (14.86\%) & 4,561 (64.26\%) & 849 (11.96\%) & 633 (8.92\%) & -- & 0 (0.00\%) & 849 (11.96\%) \\
Correct a2 Shuffled & 586 (8.26\%) & 4,486 (63.20\%) & 1,381 (19.46\%) & 645 (9.09\%) & -- & 0 (0.00\%) & 1,381 (19.46\%) \\
Correct a3 & 588 (8.28\%) & 4,387 (61.81\%) & 458 (6.45\%) & 1,665 (23.46\%) & -- & 0 (0.00\%) & 1,665 (23.46\%) \\
Correct a3 Shuffled & 1,061 (14.95\%) & 4,503 (63.44\%) & 741 (10.44\%) & 793 (11.17\%) & -- & 0 (0.00\%) & 793 (11.17\%) \\
All Correct Answers & 374 (5.27\%) & 6,710 (94.53\%) & 1 (0.01\%) & 13 (0.18\%) & -- & 0 (0.00\%) & -- \\
Empty Frames & 709 (9.99\%) & 4,772 (67.23\%) & 1,048 (14.76\%) & 569 (8.02\%) & -- & 0 (0.00\%) & 1,891 (26.64\%) \\
Empty Questions & 3,013 (42.45\%) & 3,748 (52.80\%) & 194 (2.73\%) & 143 (2.01\%) & -- & 0 (0.00\%) & 2,119 (29.85\%) \\
Empty Answers & 63 (0.89\%) & 7,033 (99.08\%) & 0 (0.00\%) & 2 (0.03\%) & -- & 0 (0.00\%) & -- \\
\hline
\end{tabular}%
}
\end{table*}

\begin{table*}[ht]
\centering
\footnotesize
\caption{Video-LLaVA Performance on All the Settings of Perception Test. Each cell shows the count of selections for each option (\textit{a0}--\textit{a3}) along with the percentage in parentheses. Accuracy represents the number of correct answers where applicable.}
\label{tab:videollava_perception}
\resizebox{\textwidth}{!}{%
\begin{tabular}{l|rrrr|r|r}
\hline
\textbf{Setting} & \textbf{a0} & \textbf{a1} & \textbf{a2} & \textbf{a3} & \textbf{N/A} & \textbf{Accuracy} \\
\hline
Target & 2,549 (33.29\%) & 2,561 (33.45\%) & 2,546 (33.25\%) & -- & 0 (0.00\%) & -- \\
Default & 1,944 (25.43\%) & 5,174 (67.68\%) & 527 (6.89\%) & -- & 11 (0.14\%) & 3,114 (40.73\%) \\
Answer Shuffling & 2,164 (28.28\%) & 4,887 (63.86\%) & 602 (7.87\%) & -- & 3 (0.04\%) & 3,259 (42.58\%) \\
Rephrased Questions & 1,836 (24.04\%) & 5,253 (68.79\%) & 547 (7.16\%) & -- & 20 (0.26\%) & 3,154 (41.30\%) \\
Additional Empty Option & 1,133 (15.01\%) & 4,548 (60.24\%) & 1,634 (21.64\%) & 235 (3.11\%) & 106 (1.38\%) & 3,233 (42.82\%) \\
All a0 & 2,942 (38.43\%) & 4,586 (59.90\%) & 128 (1.67\%) & -- & 0 (0.00\%) & -- \\
All a1 & 3,001 (39.20\%) & 4,538 (59.27\%) & 117 (1.53\%) & -- & 0 (0.00\%) & -- \\
All a2 & 3,684 (48.13\%) & 3,951 (51.61\%) & 20 (0.26\%) & -- & 1 (0.01\%) & -- \\
Correct a0 & 3,086 (40.32\%) & 4,040 (52.79\%) & 527 (6.89\%) & -- & 3 (0.04\%) & 3,086 (40.32\%) \\
Correct a0 Shuffled & 3,121 (40.78\%) & 4,038 (52.76\%) & 494 (6.45\%) & -- & 3 (0.04\%) & 3,121 (40.78\%) \\
Correct a1 & 1,524 (19.91\%) & 5,712 (74.62\%) & 419 (5.47\%) & -- & 1 (0.01\%) & 5,712 (74.62\%) \\
Correct a1 Shuffled & 1,542 (20.15\%) & 5,712 (74.63\%) & 400 (5.23\%) & -- & 2 (0.03\%) & 5,712 (74.63\%) \\
Correct a2 & 3,085 (40.32\%) & 4,040 (52.80\%) & 527 (6.89\%) & -- & 4 (0.05\%) & 527 (6.89\%) \\
Correct a2 Shuffled & 1,839 (24.03\%) & 4,944 (64.60\%) & 870 (11.37\%) & -- & 3 (0.04\%) & 870 (11.37\%) \\
All Correct Answers & 3,493 (45.64\%) & 4,078 (53.28\%) & 83 (1.08\%) & -- & 2 (0.03\%) & -- \\
Empty Frames & 1,920 (25.08\%) & 5,265 (68.77\%) & 471 (6.15\%) & -- & 0 (0.00\%) & 2,957 (38.62\%) \\
Empty Questions & 3,649 (47.66\%) & 3,767 (49.20\%) & 240 (3.13\%) & -- & 0 (0.00\%) & 3,007 (39.28\%) \\
Empty Answers & 1,183 (15.45\%) & 6,284 (82.09\%) & 188 (2.46\%) & -- & 1 (0.01\%) & -- \\
\hline
\end{tabular}%
}
\end{table*}

\begin{table*}[ht]
\centering
\footnotesize
\caption{Video-LLaVA Performance on All the Settings of Video-MME. Each cell shows the count of selections for each option (\textit{a0}--\textit{a4}) along with the percentage in parentheses. Accuracy represents the number of correct answers where applicable.}
\label{tab:videollava_videomme}
\resizebox{\textwidth}{!}{%
\begin{tabular}{l|rrrrr|r|r}
\hline
\textbf{Setting} & \textbf{a0} & \textbf{a1} & \textbf{a2} & \textbf{a3} & \textbf{a4} & \textbf{N/A} & \textbf{Accuracy} \\
\hline
Target & 675 (25.00\%) & 675 (25.00\%) & 675 (25.00\%) & 675 (25.00\%) & -- & 0 (0.00\%) & -- \\
Default & 476 (17.63\%) & 1,693 (62.70\%) & 353 (13.07\%) & 178 (6.59\%) & -- & 0 (0.00\%) & 924 (34.22\%) \\
Answer Shuffling & 427 (15.81\%) & 1,728 (64.00\%) & 373 (13.81\%) & 172 (6.37\%) & -- & 0 (0.00\%) & 841 (31.15\%) \\
Rephrased Questions & 478 (17.70\%) & 1,705 (63.15\%) & 352 (13.04\%) & 165 (6.11\%) & -- & 0 (0.00\%) & 911 (33.74\%) \\
Additional Empty Option & 208 (7.84\%) & 1,833 (69.12\%) & 191 (7.20\%) & 420 (15.84\%) & 0 (0.00\%) & 48 (1.78\%) & 909 (34.28\%) \\
All a0 & 891 (33.00\%) & 1,730 (64.07\%) & 27 (1.00\%) & 52 (1.93\%) & -- & 0 (0.00\%) & -- \\
All a1 & 888 (32.90\%) & 1,733 (64.21\%) & 33 (1.22\%) & 45 (1.67\%) & -- & 1 (0.04\%) & -- \\
All a2 & 914 (33.86\%) & 1,703 (63.10\%) & 27 (1.00\%) & 55 (2.04\%) & -- & 1 (0.04\%) & -- \\
All a3 & 888 (32.90\%) & 1,657 (61.39\%) & 32 (1.19\%) & 122 (4.52\%) & -- & 1 (0.04\%) & -- \\
Correct a0 & 660 (24.44\%) & 1,651 (61.15\%) & 258 (9.56\%) & 131 (4.85\%) & -- & 0 (0.00\%) & 660 (24.44\%) \\
Correct a0 Shuffled & 642 (23.78\%) & 1,620 (60.00\%) & 293 (10.85\%) & 145 (5.37\%) & -- & 0 (0.00\%) & 642 (23.78\%) \\
Correct a1 & 384 (14.22\%) & 1,974 (73.11\%) & 241 (8.93\%) & 101 (3.74\%) & -- & 0 (0.00\%) & 1,974 (73.11\%) \\
Correct a1 Shuffled & 364 (13.48\%) & 1,964 (72.74\%) & 240 (8.89\%) & 132 (4.89\%) & -- & 0 (0.00\%) & 1,964 (72.74\%) \\
Correct a2 & 386 (14.30\%) & 1,606 (59.48\%) & 566 (20.96\%) & 142 (5.26\%) & -- & 0 (0.00\%) & 566 (20.96\%) \\
Correct a2 Shuffled & 352 (13.04\%) & 1,623 (60.11\%) & 589 (21.81\%) & 136 (5.04\%) & -- & 0 (0.00\%) & 589 (21.81\%) \\
Correct a3 & 416 (15.41\%) & 1,642 (60.81\%) & 311 (11.52\%) & 331 (12.26\%) & -- & 0 (0.00\%) & 331 (12.26\%) \\
Correct a3 Shuffled & 357 (13.23\%) & 1,682 (62.32\%) & 345 (12.78\%) & 315 (11.67\%) & -- & 1 (0.04\%) & 315 (11.67\%) \\
All Correct Answers & 928 (34.38\%) & 1,702 (63.06\%) & 29 (1.07\%) & 40 (1.48\%) & -- & 1 (0.04\%) & -- \\
Empty Frames & 267 (9.89\%) & 1,828 (67.70\%) & 437 (16.19\%) & 168 (6.22\%) & -- & 0 (0.00\%) & 764 (28.30\%) \\
Empty Questions & 1,545 (57.24\%) & 960 (35.57\%) & 119 (4.41\%) & 75 (2.78\%) & -- & 1 (0.04\%) & 790 (29.27\%) \\
Empty Answers & 359 (13.32\%) & 2,082 (77.23\%) & 165 (6.12\%) & 90 (3.34\%) & -- & 4 (0.15\%) & -- \\
\hline
\end{tabular}%
}
\end{table*}

\begin{table*}[ht]
\centering
\footnotesize
\caption{SeViLA Performance on All the Settings of NExT-QA (including NExT-GQA). Each cell shows the count of selections for each option (\textit{a0}--\textit{a5}) along with the percentage in parentheses. Accuracy represents the number of correct answers where applicable.}
\label{tab:sevilanextqa}
\resizebox{\textwidth}{!}{%
\begin{tabular}{l|rrrrrr|r}
\hline
\textbf{Setting} & \textbf{a0} & \textbf{a1} & \textbf{a2} & \textbf{a3} & \textbf{a4} & \textbf{a5} & \textbf{Accuracy} \\
\hline
Target & 1,721 (20.10\%) & 1,671 (19.51\%) & 1,767 (20.63\%) & 1,719 (20.07\%) & 1,686 (19.69\%) & -- & -- \\
Default & 1,812 (21.16\%) & 1,606 (18.75\%) & 1,712 (19.99\%) & 1,805 (21.08\%) & 1,629 (19.02\%) & -- & 5,462 (63.78\%) \\
Answer Shuffling & 1,740 (20.32\%) & 1,598 (18.66\%) & 1,737 (20.28\%) & 1,824 (21.30\%) & 1,665 (19.44\%) & -- & 5,481 (64.00\%) \\
Correct Frames & 1,108 (22.33\%) & 939 (18.92\%) & 961 (19.37\%) & 1,069 (21.54\%) & 885 (17.84\%) & -- & 2,932 (59.09\%) \\
Rephrased Questions & 1,825 (21.31\%) & 1,571 (18.34\%) & 1,657 (19.35\%) & 1,793 (20.94\%) & 1,718 (20.06\%) & -- & 5,160 (60.25\%) \\
Additional Empty Option & 1,669 (19.49\%) & 1,401 (16.36\%) & 1,457 (17.01\%) & 1,563 (18.25\%) & 1,758 (20.53\%) & 716 (8.36\%) & 5,054 (59.01\%) \\
All a0 & 4,972 (58.06\%) & 381 (4.45\%) & 631 (7.37\%) & 1,865 (21.78\%) & 715 (8.35\%) & -- & -- \\
All a1 & 4,906 (57.29\%) & 381 (4.45\%) & 675 (7.88\%) & 1,954 (22.82\%) & 648 (7.57\%) & -- & -- \\
All a2 & 4,887 (57.06\%) & 359 (4.19\%) & 681 (7.95\%) & 1,968 (22.98\%) & 669 (7.81\%) & -- & -- \\
All a3 & 4,961 (57.93\%) & 337 (3.94\%) & 601 (7.02\%) & 1,981 (23.13\%) & 684 (7.99\%) & -- & -- \\
All a4 & 4,904 (57.26\%) & 370 (4.32\%) & 615 (7.18\%) & 1,987 (23.20\%) & 688 (8.03\%) & -- & -- \\
Correct a0 & 5,434 (63.45\%) & 732 (8.55\%) & 761 (8.89\%) & 844 (9.86\%) & 793 (9.26\%) & -- & 5,434 (63.45\%) \\
Correct a0 Shuffled & 5,443 (63.56\%) & 736 (8.59\%) & 745 (8.70\%) & 814 (9.50\%) & 826 (9.65\%) & -- & 5,443 (63.56\%) \\
Correct a1 & 870 (10.16\%) & 5,399 (63.04\%) & 752 (8.78\%) & 825 (9.63\%) & 718 (8.38\%) & -- & 5,399 (63.04\%) \\
Correct a1 Shuffled & 809 (9.45\%) & 5,379 (62.81\%) & 740 (8.64\%) & 866 (10.11\%) & 770 (8.99\%) & -- & 5,379 (62.81\%) \\
Correct a2 & 813 (9.49\%) & 741 (8.65\%) & 5,495 (64.16\%) & 772 (9.01\%) & 743 (8.68\%) & -- & 5,495 (64.16\%) \\
Correct a2 Shuffled & 777 (9.07\%) & 734 (8.57\%) & 5,504 (64.27\%) & 796 (9.29\%) & 753 (8.79\%) & -- & 5,504 (64.27\%) \\
Correct a3 & 827 (9.66\%) & 704 (8.22\%) & 778 (9.08\%) & 5,591 (65.28\%) & 664 (7.75\%) & -- & 5,591 (65.28\%) \\
Correct a3 Shuffled & 782 (9.13\%) & 684 (7.99\%) & 783 (9.14\%) & 5,566 (64.99\%) & 749 (8.75\%) & -- & 5,566 (64.99\%) \\
Correct a4 & 860 (10.04\%) & 730 (8.52\%) & 708 (8.27\%) & 853 (9.96\%) & 5,413 (63.21\%) & -- & 5,413 (63.21\%) \\
Correct a4 Shuffled & 810 (9.46\%) & 727 (8.49\%) & 760 (8.87\%) & 831 (9.70\%) & 5,436 (63.48\%) & -- & 5,436 (63.48\%) \\
All Correct Answers & 7,300 (85.24\%) & 209 (2.44\%) & 237 (2.77\%) & 448 (5.23\%) & 370 (4.32\%) & -- & -- \\
Empty Frames & 1,833 (21.40\%) & 1,495 (17.46\%) & 1,626 (18.99\%) & 1,820 (21.25\%) & 1,790 (20.90\%) & -- & 3,921 (45.78\%) \\
Empty Questions & 1,671 (19.51\%) & 1,513 (17.67\%) & 1,685 (19.68\%) & 1,851 (21.61\%) & 1,844 (21.53\%) & -- & 4,485 (52.37\%) \\
Empty Answers & 2,026 (23.66\%) & 52 (0.61\%) & 5 (0.06\%) & 0 (0.00\%) & 6,481 (75.68\%) & -- & -- \\
\hline
\end{tabular}%
}
\end{table*}

\begin{table*}[ht]
\centering
\footnotesize
\caption{SeViLA Performance on All the Settings of STAR. Each cell shows the count of selections for each option (\textit{a0}--\textit{a4}) along with the percentage in parentheses. Accuracy represents the number of correct answers where applicable.}
\label{tab:sevilastar}
\resizebox{\textwidth}{!}{%
\begin{tabular}{l|rrrrr|r}
\hline
\textbf{Setting} & \textbf{a0} & \textbf{a1} & \textbf{a2} & \textbf{a3} & \textbf{a4} & \textbf{Accuracy} \\
\hline
Target & 1,755 (24.73\%) & 1,758 (24.77\%) & 1,742 (24.54\%) & 1,843 (25.97\%) & -- & -- \\
Default & 1,501 (21.15\%) & 1,739 (24.50\%) & 1,782 (25.11\%) & 2,076 (29.25\%) & -- & 3,285 (46.28\%) \\
Answer Shuffling & 1,473 (20.75\%) & 1,713 (24.13\%) & 1,807 (25.46\%) & 2,105 (29.66\%) & -- & 3,260 (45.93\%) \\
Correct Frames & 1,526 (21.50\%) & 1,757 (24.75\%) & 1,759 (24.78\%) & 2,056 (28.97\%) & -- & 3,248 (45.76\%) \\
Rephrased Questions & 1,507 (21.23\%) & 1,736 (24.46\%) & 1,771 (24.95\%) & 2,084 (29.36\%) & -- & 3,153 (44.42\%) \\
Additional Empty Option & 1,003 (14.13\%) & 1,077 (15.17\%) & 1,214 (17.10\%) & 1,395 (19.65\%) & 2,409 (33.94\%) & 2,351 (33.12\%) \\
All a0 & 2,951 (41.58\%) & 510 (7.19\%) & 480 (6.76\%) & 3,157 (44.48\%) & -- & -- \\
All a1 & 2,909 (40.98\%) & 492 (6.93\%) & 480 (6.76\%) & 3,217 (45.32\%) & -- & -- \\
All a2 & 2,935 (41.35\%) & 572 (8.06\%) & 442 (6.23\%) & 3,149 (44.36\%) & -- & -- \\
All a3 & 2,892 (40.74\%) & 550 (7.75\%) & 472 (6.65\%) & 3,184 (44.86\%) & -- & -- \\
Correct a0 & 2,806 (39.53\%) & 1,342 (18.91\%) & 1,351 (19.03\%) & 1,599 (22.53\%) & -- & 2,806 (39.53\%) \\
Correct a0 Shuffled & 2,864 (40.35\%) & 1,274 (17.95\%) & 1,312 (18.48\%) & 1,648 (23.22\%) & -- & 2,864 (40.35\%) \\
Correct a1 & 1,149 (16.19\%) & 3,216 (45.31\%) & 1,231 (17.34\%) & 1,502 (21.16\%) & -- & 3,216 (45.31\%) \\
Correct a1 Shuffled & 1,113 (15.68\%) & 3,181 (44.82\%) & 1,335 (18.81\%) & 1,469 (20.70\%) & -- & 3,181 (44.82\%) \\
Correct a2 & 1,141 (16.07\%) & 1,223 (17.23\%) & 3,253 (45.83\%) & 1,481 (20.87\%) & -- & 3,253 (45.83\%) \\
Correct a2 Shuffled & 1,090 (15.36\%) & 1,252 (17.64\%) & 3,301 (46.51\%) & 1,455 (20.50\%) & -- & 3,301 (46.51\%) \\
Correct a3 & 1,088 (15.33\%) & 1,215 (17.12\%) & 1,189 (16.75\%) & 3,606 (50.80\%) & -- & 3,606 (50.80\%) \\
Correct a3 Shuffled & 1,074 (15.13\%) & 1,194 (16.82\%) & 1,242 (17.50\%) & 3,588 (50.55\%) & -- & 3,588 (50.55\%) \\
All Correct Answers & 3,363 (47.38\%) & 391 (5.51\%) & 171 (2.41\%) & 3,173 (44.70\%) & -- & -- \\
Empty Frames & 1,501 (21.15\%) & 1,739 (24.50\%) & 1,782 (25.11\%) & 2,076 (29.25\%) & -- & 3,285 (46.28\%) \\
Empty Questions & 1,563 (22.02\%) & 1,608 (22.65\%) & 1,700 (23.95\%) & 2,227 (31.38\%) & -- & 3,496 (49.25\%) \\
Empty Answers & 1,031 (14.53\%) & 25 (0.35\%) & 0 (0.00\%) & 6,042 (85.12\%) & -- & -- \\
\hline
\end{tabular}%
}
\label{tab:sevila_star_perf}
\end{table*}

\begin{table*}[ht]
\centering
\footnotesize
\caption{SeViLA Performance on All the Settings of Perception Test. Each cell shows the count of selections for each option (\textit{a0}--\textit{a3}) along with the percentage in parentheses. Accuracy represents the number of correct answers where applicable.}
\label{tab:sevilaperception}
\resizebox{\textwidth}{!}{%
\begin{tabular}{l|rrrr|r}
\hline
\textbf{Setting} & \textbf{a0} & \textbf{a1} & \textbf{a2} & \textbf{a3} & \textbf{Accuracy} \\
\hline
Target & 2,549 (33.29\%) & 2,561 (33.45\%) & 2,546 (33.25\%) & -- & -- \\
Default & 2,256 (29.47\%) & 2,331 (30.45\%) & 3,069 (40.09\%) & -- & 3,468 (45.30\%) \\
Answer Shuffling & 2,323 (30.34\%) & 2,300 (30.04\%) & 3,033 (39.62\%) & -- & 3,519 (45.96\%) \\
Rephrased Questions & 2,286 (29.86\%) & 2,231 (29.14\%) & 3,139 (41.00\%) & -- & 3,369 (44.00\%) \\
Additional Empty Option & 1,674 (21.87\%) & 1,246 (16.27\%) & 1,436 (18.76\%) & 3,300 (43.10\%) & 2,223 (29.04\%) \\
All a0 & 3,380 (44.15\%) & 464 (6.06\%) & 3,812 (49.79\%) & -- & -- \\
All a1 & 3,427 (44.76\%) & 449 (5.86\%) & 3,780 (49.37\%) & -- & -- \\
All a2 & 3,377 (44.11\%) & 449 (5.86\%) & 3,830 (50.03\%) & -- & -- \\
Correct a0 & 3,128 (40.86\%) & 1,978 (25.84\%) & 2,550 (33.31\%) & -- & 3,128 (40.86\%) \\
Correct a0 Shuffled & 3,112 (40.65\%) & 2,022 (26.41\%) & 2,522 (32.94\%) & -- & 3,112 (40.65\%) \\
Correct a1 & 1,909 (24.93\%) & 3,249 (42.44\%) & 2,498 (32.63\%) & -- & 3,249 (42.44\%) \\
Correct a1 Shuffled & 1,869 (24.41\%) & 3,268 (42.69\%) & 2,519 (32.90\%) & -- & 3,268 (42.69\%) \\
Correct a2 & 1,858 (24.27\%) & 1,714 (22.39\%) & 4,084 (53.34\%) & -- & 4,084 (53.34\%) \\
Correct a2 Shuffled & 1,846 (24.11\%) & 1,750 (22.86\%) & 4,060 (53.03\%) & -- & 4,060 (53.03\%) \\
All Correct Answers & 3,531 (46.12\%) & 353 (4.61\%) & 3,772 (49.27\%) & -- & -- \\
Empty Frames & 2,684 (35.06\%) & 2,058 (26.88\%) & 2,914 (38.06\%) & -- & 2,766 (36.13\%) \\
Empty Questions & 2,574 (33.62\%) & 2,040 (26.65\%) & 3,042 (39.73\%) & -- & 3,139 (41.00\%) \\
Empty Answers & 985 (12.87\%) & 0 (0.00\%) & 6,671 (87.13\%) & -- & -- \\
\hline
\end{tabular}%
}
\end{table*}

\begin{table*}[ht]
\centering
\footnotesize
\caption{SeViLA Performance on All the Settings of Video-MME. Each cell shows the count of selections for each option (\textit{a0}--\textit{a4}) along with the percentage in parentheses. Accuracy represents the number of correct answers where applicable.}
\label{tab:sevilavideomme}
\resizebox{\textwidth}{!}{%
\begin{tabular}{l|rrrrr|r}
\hline
\textbf{Setting} & \textbf{a0} & \textbf{a1} & \textbf{a2} & \textbf{a3} & \textbf{a4} & \textbf{Accuracy} \\
\hline
Target & 675 (25.00\%) & 675 (25.00\%) & 675 (25.00\%) & 675 (25.00\%) & -- & -- \\
Default & 911 (33.74\%) & 592 (21.93\%) & 568 (21.04\%) & 629 (23.30\%) & -- & 1,076 (39.85\%) \\
Answer Shuffling & 887 (32.85\%) & 566 (20.96\%) & 591 (21.89\%) & 656 (24.30\%) & -- & 1,041 (38.56\%) \\
Rephrased Questions & 947 (35.07\%) & 556 (20.59\%) & 552 (20.44\%) & 645 (23.89\%) & -- & 1,041 (38.56\%) \\
Additional Empty Option & 817 (30.26\%) & 413 (15.30\%) & 428 (15.85\%) & 472 (17.48\%) & 570 (21.11\%) & 875 (32.41\%) \\
All a0 & 1,631 (60.41\%) & 97 (3.59\%) & 140 (5.19\%) & 832 (30.81\%) & -- & -- \\
All a1 & 1,681 (62.26\%) & 76 (2.81\%) & 164 (6.07\%) & 779 (28.85\%) & -- & -- \\
All a2 & 1,637 (60.63\%) & 100 (3.70\%) & 148 (5.48\%) & 815 (30.19\%) & -- & -- \\
All a3 & 1,573 (58.26\%) & 117 (4.33\%) & 153 (5.67\%) & 857 (31.74\%) & -- & -- \\
Correct a0 & 1,271 (47.07\%) & 469 (17.37\%) & 451 (16.70\%) & 509 (18.85\%) & -- & 1,271 (47.07\%) \\
Correct a0 Shuffled & 1,229 (45.52\%) & 464 (17.19\%) & 469 (17.37\%) & 538 (19.93\%) & -- & 1,229 (45.52\%) \\
Correct a1 & 752 (27.85\%) & 977 (36.19\%) & 460 (17.04\%) & 511 (18.93\%) & -- & 977 (36.19\%) \\
Correct a1 Shuffled & 714 (26.44\%) & 951 (35.22\%) & 474 (17.56\%) & 561 (20.78\%) & -- & 951 (35.22\%) \\
Correct a2 & 793 (29.37\%) & 443 (16.41\%) & 990 (36.67\%) & 474 (17.56\%) & -- & 990 (36.67\%) \\
Correct a2 Shuffled & 755 (27.96\%) & 430 (15.93\%) & 984 (36.44\%) & 531 (19.67\%) & -- & 984 (36.44\%) \\
Correct a3 & 725 (26.85\%) & 451 (16.70\%) & 463 (17.15\%) & 1,061 (39.30\%) & -- & 1,061 (39.30\%) \\
Correct a3 Shuffled & 746 (27.63\%) & 433 (16.04\%) & 475 (17.59\%) & 1,046 (38.74\%) & -- & 1,046 (38.74\%) \\
All Correct Answers & 1,840 (68.15\%) & 75 (2.78\%) & 93 (3.44\%) & 692 (25.63\%) & -- & -- \\
Empty Frames & 912 (33.78\%) & 569 (21.07\%) & 555 (20.56\%) & 664 (24.59\%) & -- & 773 (28.63\%) \\
Empty Questions & 876 (32.44\%) & 523 (19.37\%) & 571 (21.15\%) & 730 (27.04\%) & -- & 1,020 (37.78\%) \\
Empty Answers & 727 (26.93\%) & 29 (1.07\%) & 37 (1.37\%) & 1,907 (70.63\%) & -- & -- \\
\hline
\end{tabular}%
}
\end{table*}


\begin{algorithm*}[t]
\caption{BOLD: Debiasing with Prior Estimation by Decomposition}
\label{alg:BOLD}
\begin{algorithmic}[1]
\Require Video-Language model $M$, dataset $\mathcal{D} = \{T_i\}$, number of options $n$, estimation budget $K$
\Ensure Model predictions $\mathcal{Y}$
\State Initialize prediction set $\mathcal{Y} \leftarrow \emptyset$ and prior set $\mathcal{P} \leftarrow \emptyset$ \Comment{Initialization}
\State Sample estimation set $\mathcal{D}_k$ of size $K$ from $\mathcal{D}$, set $\mathcal{D}_r \leftarrow \mathcal{D} \setminus \mathcal{D}_k$
\For{$T \in \mathcal{D}_k$}
    \State Apply attacks to get observed predictions:
    \Statex \quad $P_{o}(d_i \mid A|_{v=0}(T))$, $P_{o}(d_i \mid A|_{q=0}(T))$, $P_{o}(d_i \mid A|_{\text{opts}=0}(T))$
    \State Compute decomposed priors using Equation~\ref{eq:eq_attacked_prior}:
    \Statex \quad $P_{\text{pri}}(d_i \mid A|_{v=0}(T))$, $P_{\text{pri}}(d_i \mid A|_{q=0}(T))$, $P_{\text{pri}}(d_i \mid A|_{\text{opts}=0}(T))$
    \State Estimate sample-specific prior $\tilde{P}_{\text{prior}}(d_i \mid T)$ using Equation~\ref{eq:bias_prior_decomposition}; add to $\mathcal{P}$
\EndFor
\State Estimate global prior $\tilde{P}_{\text{prior}}(d_i)$ by averaging $\mathcal{P}$ \Comment{Prior Estimation}
\For{$T \in \mathcal{D}$}
    \State Debias model prediction using the global prior:
    \Statex \quad $P_{\text{deb}}(d_i \mid T) = \operatorname{softmax} \left( \log P_o(d_i \mid T) - \log \tilde{P}_{\text{prior}}(d_i) \right)$
    \State Add the predicted answer to $\mathcal{Y}$
\EndFor
\State \Return $\mathcal{Y}$
\end{algorithmic}
\end{algorithm*}

\begin{algorithm*}[t]
\caption{Weighted\_BOLD: Debiasing with Prior Estimation by Decomposition with Weights}
\label{alg:Weighted_BOLD}
\begin{algorithmic}[1]
\Require Model $M$, dataset $\mathcal{D} = \{T_i\}$, estimation budget $K$, number of folds $F = 5$
\Ensure Debiased model predictions $\mathcal{Y}$
\State Initialize prediction set $\mathcal{Y} \leftarrow \emptyset$ and prior set $\mathcal{P} \leftarrow \emptyset$ \Comment{Initialization}
\State Sample estimation set $\mathcal{D}_k$ of size $K$ from $\mathcal{D}$; set remaining samples $\mathcal{D}_r \leftarrow \mathcal{D} \setminus \mathcal{D}_k$
\For{each fold $f = 1$ to $F$} \Comment{Cross-validation}
    \State Split $\mathcal{D}_k$ into test set $\mathcal{D}_{\text{test}}^f$ and validation set $\mathcal{D}_{\text{val}}^f$
    \For{$T \in \mathcal{D}_{\text{test}}^f$}
        \State Apply attacks to obtain observed predictions: $P_{o}(d_i \mid A(T))$
        \State Compute decomposed priors for each attack using Equation~\ref{eq:eq_attacked_prior}: $P_{\text{pri}}(d_i \mid A(T))$
        \State Estimate sample-specific prior $\tilde{P}_{\text{pri}}(d_i \mid T)$ using Equation~\ref{eq:bias_prior_decomposition_extended}; add to $\mathcal{P}^f$
    \EndFor
    \State Optimize weights $\{w_i\}$ using COBYLA, subject to constraints $0 \leq w_i \leq 1$ or \( |w_i| \leq 1 \), to minimize \textit{recall\_std} bias on $\mathcal{D}_{\text{test}}^f$
\EndFor
\State Compute global prior by averaging over folds:
\[
\tilde{P}_{\text{prior}}(d_i) = \frac{1}{F} \sum_{f=1}^{F} \tilde{P}_{\text{pri}}^f(d_i)
\]
\Comment{Global Prior Estimation}
\For{$T \in \mathcal{D}$} \Comment{Debiasing}
    \State Debias the model prediction using the global prior:
    \[
    P_{\text{deb}}(d_i \mid T) = \operatorname{softmax} \left( \log P_o(d_i \mid T) - \log \tilde{P}_{\text{prior}}(d_i) \right)
    \]
    \State Add the predicted answer to $\mathcal{Y}$
\EndFor
\State \Return $\mathcal{Y}$
\end{algorithmic}
\end{algorithm*}

\begin{table*}[ht]
\centering
\caption{Comparison of BOLD and Weighted\_BOLD bias mitigation approaches across models and datasets for performance and bias monitoring metrics with $k$ = 0.25. Green arrows indicate improvements: upward for Accuracy and F1\_mean, and downward for standard deviation metrics. Red arrows represent deterioration, respectively.}
\footnotesize{
\resizebox{\textwidth}{!}{
\begin{tabular}{l l p{2.7cm} r p{2.2cm} p{2.2cm} p{2.2cm} p{2.2cm}}
\hline
\multirow{2}{*}{\textbf{Model}} & \multirow{2}{*}{\textbf{Dataset}} & \multirow{2}{*}{\textbf{Configuration}} & \multicolumn{2}{c}{\textbf{Performance Metrics}} & \multicolumn{3}{c}{\textbf{Bias Monitoring Metrics}} \\ 
\cline{4-5} \cline{6-8}
& & & \textbf{Accuracy} & \textbf{F1\_mean} & \textbf{Recall\_std} & \textbf{F1\_std} & \textbf{JS\_std} \\ 
\hline
\multirow{4}{*}{} & NExT-QA & \textbf{Default} & 49.95 & 49.81  & 16.31  & 3.27 & 5.4\\
& & \textbf{BOLD} & 51.81 (\textcolor[rgb]{0,0.5,0}{↑}3.72\%) & 51.71 (\textcolor[rgb]{0,0.5,0}{↑}3.82\%) & 13.29 (\textcolor[rgb]{0,0.5,0}{↓}18.54\%) & 2.66 (\textcolor[rgb]{0,0.5,0}{↓}18.59\%) & 4.58 (\textcolor[rgb]{0,0.5,0}{↓}15.12\%)\\
& & \textbf{+Weighted\_BOLD} & 51.89 (\textcolor[rgb]{0,0.5,0}{↑}3.88\%) & 51.77 (\textcolor[rgb]{0,0.5,0}{↑}3.95\%) & 13.28 (\textcolor[rgb]{0,0.5,0}{↓}18.57\%) & 2.73 (\textcolor[rgb]{0,0.5,0}{↓}16.57\%) & 4.56 (\textcolor[rgb]{0,0.5,0}{↓}15.39\%)\\
& & \textbf{-/+Weighted\_BOLD} & 52.28 (\textcolor[rgb]{0,0.5,0}{↑}4.65\%) & 52.25 (\textcolor[rgb]{0,0.5,0}{↑}4.9\%) & 10.9 (\textcolor[rgb]{0,0.5,0}{↓}33.18\%) & 2.23 (\textcolor[rgb]{0,0.5,0}{↓}31.89\%) & 3.9 (\textcolor[rgb]{0,0.5,0}{↓}27.63\%)\\
\cline{2-8}
\multirow{4}{*}{Video-LLaVA} & STAR & \textbf{Default} & 34.77 & 31.83  & 24.99  & 6.82 & 5.78\\
& & \textbf{BOLD} & 37.45 (\textcolor[rgb]{0,0.5,0}{↑}7.7\%) & 36.14 (\textcolor[rgb]{0,0.5,0}{↑}13.54\%) & 17.43 (\textcolor[rgb]{0,0.5,0}{↓}30.23\%) & 4.84 (\textcolor[rgb]{0,0.5,0}{↓}29.02\%) & 4.52 (\textcolor[rgb]{0,0.5,0}{↓}21.74\%)\\
& & \textbf{+Weighted\_BOLD} & 37.81 (\textcolor[rgb]{0,0.5,0}{↑}8.75\%) & 36.67 (\textcolor[rgb]{0,0.5,0}{↑}15.22\%) & 16.15 (\textcolor[rgb]{0,0.5,0}{↓}35.38\%) & 4.71 (\textcolor[rgb]{0,0.5,0}{↓}30.94\%) & 4.15 (\textcolor[rgb]{0,0.5,0}{↓}28.15\%)\\
& & \textbf{-/+Weighted\_BOLD} & 39.87 (\textcolor[rgb]{0,0.5,0}{↑}14.67\%) & 39.65 (\textcolor[rgb]{0,0.5,0}{↑}24.57\%) & 6.85 (\textcolor[rgb]{0,0.5,0}{↓}72.56\%) & 2.34 (\textcolor[rgb]{0,0.5,0}{↓}65.67\%) & 2.08 (\textcolor[rgb]{0,0.5,0}{↓}63.98\%)\\
\cline{2-8}
\multirow{4}{*}{} & Perception Test & \textbf{Default} & 40.72 & 35.69  & 27.5  & 14.75 & 5.28\\
& & \textbf{BOLD} & 41.62 (\textcolor[rgb]{0,0.5,0}{↑}2.22\%) & 38.68 (\textcolor[rgb]{0,0.5,0}{↑}8.38\%) & 21.77 (\textcolor[rgb]{0,0.5,0}{↓}20.85\%) & 10.93 (\textcolor[rgb]{0,0.5,0}{↓}25.88\%) & 3.77 (\textcolor[rgb]{0,0.5,0}{↓}28.54\%)\\
& & \textbf{+Weighted\_BOLD} & 41.95 (\textcolor[rgb]{0,0.5,0}{↑}3.02\%) & 39.46 (\textcolor[rgb]{0,0.5,0}{↑}10.58\%) & 19.73 (\textcolor[rgb]{0,0.5,0}{↓}28.26\%) & 10.24 (\textcolor[rgb]{0,0.5,0}{↓}30.53\%) & 3.39 (\textcolor[rgb]{0,0.5,0}{↓}35.79\%)\\
& & \textbf{-/+Weighted\_BOLD} & 42.01 (\textcolor[rgb]{0,0.5,0}{↑}3.18\%) & 40.48 (\textcolor[rgb]{0,0.5,0}{↑}13.43\%) & 14.69 (\textcolor[rgb]{0,0.5,0}{↓}46.59\%) & 8.48 (\textcolor[rgb]{0,0.5,0}{↓}42.52\%) & 2.91 (\textcolor[rgb]{0,0.5,0}{↓}44.9\%)\\
\cline{2-8}
\multirow{4}{*}{} & Video-MME & \textbf{Default} & 34.3 & 30.99  & 24.09  & 8.06 & 4.79\\
& & \textbf{BOLD} & 34.63 (\textcolor[rgb]{0,0.5,0}{↑}0.97\%) & 32.71 (\textcolor[rgb]{0,0.5,0}{↑}5.54\%) & 18.21 (\textcolor[rgb]{0,0.5,0}{↓}24.4\%) & 6.18 (\textcolor[rgb]{0,0.5,0}{↓}23.36\%) & 3.86 (\textcolor[rgb]{0,0.5,0}{↓}19.47\%)\\
& & \textbf{+Weighted\_BOLD} & 34.59 (\textcolor[rgb]{0,0.5,0}{↑}0.86\%) & 33.01 (\textcolor[rgb]{0,0.5,0}{↑}6.51\%) & 15.92 (\textcolor[rgb]{0,0.5,0}{↓}33.93\%) & 5.91 (\textcolor[rgb]{0,0.5,0}{↓}26.68\%) & 3.38 (\textcolor[rgb]{0,0.5,0}{↓}29.56\%)\\
& & \textbf{-/+Weighted\_BOLD} & 33.93 (\textcolor{red}{↓}1.08\%) & 33.21 (\textcolor[rgb]{0,0.5,0}{↑}7.16\%) & 10.1 (\textcolor[rgb]{0,0.5,0}{↓}58.07\%) & 4.09 (\textcolor[rgb]{0,0.5,0}{↓}49.26\%) & 2.33 (\textcolor[rgb]{0,0.5,0}{↓}51.39\%)\\
\cline{1-8}
\multirow{4}{*}{} & NExT-QA & \textbf{Default} & 44.79 & 40.85  & 23.83  & 15.86 & 16.09\\
& & \textbf{BOLD} & 45.88 (\textcolor[rgb]{0,0.5,0}{↑}2.43\%) & 42.15 (\textcolor[rgb]{0,0.5,0}{↑}3.18\%) & 22.98 (\textcolor[rgb]{0,0.5,0}{↓}3.53\%) & 15.53 (\textcolor[rgb]{0,0.5,0}{↓}2.11\%) & 15.56 (\textcolor[rgb]{0,0.5,0}{↓}3.3\%)\\
& & \textbf{+Weighted\_BOLD} & 45.96 (\textcolor[rgb]{0,0.5,0}{↑}2.61\%) & 42.27 (\textcolor[rgb]{0,0.5,0}{↑}3.46\%) & 22.78 (\textcolor[rgb]{0,0.5,0}{↓}4.4\%) & 15.46 (\textcolor[rgb]{0,0.5,0}{↓}2.55\%) & 15.52 (\textcolor[rgb]{0,0.5,0}{↓}3.55\%)\\
& & \textbf{-/+Weighted\_BOLD} & 47.56 (\textcolor[rgb]{0,0.5,0}{↑}6.19\%) & 46.63 (\textcolor[rgb]{0,0.5,0}{↑}14.14\%) & 13.79 (\textcolor[rgb]{0,0.5,0}{↓}42.11\%) & 8.47 (\textcolor[rgb]{0,0.5,0}{↓}46.62\%) & 12.14 (\textcolor[rgb]{0,0.5,0}{↓}24.53\%)\\
\cline{2-8}
\multirow{4}{*}{Video-LLaMA} & STAR & \textbf{Default} & 36.52 & 31.86  & 24.1  & 15.44 & 15.29\\
& & \textbf{BOLD} & 37.19 (\textcolor[rgb]{0,0.5,0}{↑}1.82\%) & 33.02 (\textcolor[rgb]{0,0.5,0}{↑}3.65\%) & 22.29 (\textcolor[rgb]{0,0.5,0}{↓}7.5\%) & 14.55 (\textcolor[rgb]{0,0.5,0}{↓}5.75\%) & 14.58 (\textcolor[rgb]{0,0.5,0}{↓}4.6\%)\\
& & \textbf{+Weighted\_BOLD} & 37.21 (\textcolor[rgb]{0,0.5,0}{↑}1.89\%) & 33.32 (\textcolor[rgb]{0,0.5,0}{↑}4.59\%) & 21.45 (\textcolor[rgb]{0,0.5,0}{↓}10.97\%) & 14.03 (\textcolor[rgb]{0,0.5,0}{↓}9.1\%) & 14.22 (\textcolor[rgb]{0,0.5,0}{↓}7.0\%)\\
& & \textbf{-/+Weighted\_BOLD} & 38.63 (\textcolor[rgb]{0,0.5,0}{↑}5.76\%) & 38.07 (\textcolor[rgb]{0,0.5,0}{↑}19.52\%) & 10.16 (\textcolor[rgb]{0,0.5,0}{↓}57.84\%) & 4.78 (\textcolor[rgb]{0,0.5,0}{↓}69.05\%) & 8.93 (\textcolor[rgb]{0,0.5,0}{↓}41.61\%)\\
\cline{2-8}
\multirow{4}{*}{} & Perception Test & \textbf{Default} & 41.39 & 37.19  & 27.06  & 11.4 & 16.06\\
& & \textbf{BOLD} & 41.92 (\textcolor[rgb]{0,0.5,0}{↑}1.27\%) & 38.7 (\textcolor[rgb]{0,0.5,0}{↑}4.06\%) & 24.04 (\textcolor[rgb]{0,0.5,0}{↓}11.16\%) & 9.85 (\textcolor[rgb]{0,0.5,0}{↓}13.61\%) & 14.44 (\textcolor[rgb]{0,0.5,0}{↓}10.12\%)\\
& & \textbf{+Weighted\_BOLD} & 42.06 (\textcolor[rgb]{0,0.5,0}{↑}1.62\%) & 39.36 (\textcolor[rgb]{0,0.5,0}{↑}5.86\%) & 21.8 (\textcolor[rgb]{0,0.5,0}{↓}19.45\%) & 9.11 (\textcolor[rgb]{0,0.5,0}{↓}20.06\%) & 13.05 (\textcolor[rgb]{0,0.5,0}{↓}18.75\%)\\
& & \textbf{-/+Weighted\_BOLD} & 42.05 (\textcolor[rgb]{0,0.5,0}{↑}1.59\%) & 41.35 (\textcolor[rgb]{0,0.5,0}{↑}11.2\%) & 11.13 (\textcolor[rgb]{0,0.5,0}{↓}58.86\%) & 4.71 (\textcolor[rgb]{0,0.5,0}{↓}58.65\%) & 8.43 (\textcolor[rgb]{0,0.5,0}{↓}47.51\%)\\
\cline{2-8}
\multirow{4}{*}{} & Video-MME & \textbf{Default} & 32.05 & 28.15  & 20.62  & 12.52 & 13.75\\
& & \textbf{BOLD} & 32.69 (\textcolor[rgb]{0,0.5,0}{↑}2.01\%) & 29.2 (\textcolor[rgb]{0,0.5,0}{↑}3.73\%) & 19.28 (\textcolor[rgb]{0,0.5,0}{↓}6.5\%) & 11.95 (\textcolor[rgb]{0,0.5,0}{↓}4.54\%) & 13.13 (\textcolor[rgb]{0,0.5,0}{↓}4.53\%)\\
& & \textbf{+Weighted\_BOLD} & 32.2 (\textcolor[rgb]{0,0.5,0}{↑}0.47\%) & 28.98 (\textcolor[rgb]{0,0.5,0}{↑}2.94\%) & 17.65 (\textcolor[rgb]{0,0.5,0}{↓}14.38\%) & 11.69 (\textcolor[rgb]{0,0.5,0}{↓}6.56\%) & 12.6 (\textcolor[rgb]{0,0.5,0}{↓}8.39\%)\\
& & \textbf{-/+Weighted\_BOLD} & 32.92 (\textcolor[rgb]{0,0.5,0}{↑}2.72\%) & 30.39 (\textcolor[rgb]{0,0.5,0}{↑}7.97\%) & 15.65 (\textcolor[rgb]{0,0.5,0}{↓}24.08\%) & 10.21 (\textcolor[rgb]{0,0.5,0}{↓}18.39\%) & 11.68 (\textcolor[rgb]{0,0.5,0}{↓}15.04\%)\\
\cline{1-8}
\multirow{4}{*}{} & NExT-QA & \textbf{Default} & 63.91 & 63.88  & 2.15  & 1.32 & 1.99\\
& & \textbf{BOLD} & 63.92 (\textcolor[rgb]{0,0.5,0}{↑}0.02\%) & 63.89 (\textcolor[rgb]{0,0.5,0}{↑}0.03\%) & 2.0 (\textcolor[rgb]{0,0.5,0}{↓}6.81\%) & 1.18 (\textcolor[rgb]{0,0.5,0}{↓}10.43\%) & 1.97 (\textcolor[rgb]{0,0.5,0}{↓}0.91\%)\\
& & \textbf{+Weighted\_BOLD} & 63.88 (\textcolor{red}{↓}0.04\%) & 63.86 (\textcolor{red}{↓}0.03\%) & 2.08 (\textcolor[rgb]{0,0.5,0}{↓}3.47\%) & 1.28 (\textcolor[rgb]{0,0.5,0}{↓}2.87\%) & 1.9 (\textcolor[rgb]{0,0.5,0}{↓}4.64\%)\\
& & \textbf{-/+Weighted\_BOLD} & 63.95 (\textcolor[rgb]{0,0.5,0}{↑}0.07\%) & 63.95 (\textcolor[rgb]{0,0.5,0}{↑}0.11\%) & 1.44 (\textcolor[rgb]{0,0.5,0}{↓}32.8\%) & 1.16 (\textcolor[rgb]{0,0.5,0}{↓}12.17\%) & 1.63 (\textcolor[rgb]{0,0.5,0}{↓}18.3\%)\\
\cline{2-8}
\multirow{4}{*}{SeViLA} & STAR & \textbf{Default} & 46.28 & 46.14  & 4.47  & 2.3 & 2.25\\
& & \textbf{BOLD} & 46.17 (\textcolor{red}{↓}0.24\%) & 46.05 (\textcolor{red}{↓}0.2\%) & 4.09 (\textcolor[rgb]{0,0.5,0}{↓}8.42\%) & 2.24 (\textcolor[rgb]{0,0.5,0}{↓}2.93\%) & 2.11 (\textcolor[rgb]{0,0.5,0}{↓}6.25\%)\\
& & \textbf{+Weighted\_BOLD} & 46.07 (\textcolor{red}{↓}0.46\%) & 45.95 (\textcolor{red}{↓}0.42\%) & 4.12 (\textcolor[rgb]{0,0.5,0}{↓}7.63\%) & 2.28 (\textcolor[rgb]{0,0.5,0}{↓}1.24\%) & 2.13 (\textcolor[rgb]{0,0.5,0}{↓}5.17\%)\\
& & \textbf{-/+Weighted\_BOLD} & 46.2 (\textcolor{red}{↓}0.18\%) & 46.08 (\textcolor{red}{↓}0.13\%) & 4.04 (\textcolor[rgb]{0,0.5,0}{↓}9.49\%) & 2.23 (\textcolor[rgb]{0,0.5,0}{↓}3.15\%) & 2.16 (\textcolor[rgb]{0,0.5,0}{↓}3.9\%)\\
\cline{2-8}
\multirow{4}{*}{} & Perception Test & \textbf{Default} & 45.3 & 45.11  & 6.21  & 3.02 & 1.59\\
& & \textbf{BOLD} & 45.28 (\textcolor{red}{↓}0.03\%) & 45.15 (\textcolor[rgb]{0,0.5,0}{↑}0.09\%) & 5.1 (\textcolor[rgb]{0,0.5,0}{↓}17.85\%) & 2.93 (\textcolor[rgb]{0,0.5,0}{↓}3.09\%) & 1.26 (\textcolor[rgb]{0,0.5,0}{↓}21.04\%)\\
& & \textbf{+Weighted\_BOLD} & 45.34 (\textcolor[rgb]{0,0.5,0}{↑}0.09\%) & 45.24 (\textcolor[rgb]{0,0.5,0}{↑}0.29\%) & 4.26 (\textcolor[rgb]{0,0.5,0}{↓}31.39\%) & 2.65 (\textcolor[rgb]{0,0.5,0}{↓}12.43\%) & 1.03 (\textcolor[rgb]{0,0.5,0}{↓}35.41\%)\\
& & \textbf{-/+Weighted\_BOLD} & 45.22 (\textcolor{red}{↓}0.17\%) & 45.15 (\textcolor[rgb]{0,0.5,0}{↑}0.09\%) & 3.69 (\textcolor[rgb]{0,0.5,0}{↓}40.51\%) & 2.38 (\textcolor[rgb]{0,0.5,0}{↓}21.06\%) & 0.85 (\textcolor[rgb]{0,0.5,0}{↓}46.67\%)\\
\cline{2-8}
\multirow{4}{*}{} & Video-MME & \textbf{Default} & 39.85 & 39.82  & 4.67  & 1.68 & 0.84\\
& & \textbf{BOLD} & 40.04 (\textcolor[rgb]{0,0.5,0}{↑}0.46\%) & 40.04 (\textcolor[rgb]{0,0.5,0}{↑}0.56\%) & 3.79 (\textcolor[rgb]{0,0.5,0}{↓}18.89\%) & 1.48 (\textcolor[rgb]{0,0.5,0}{↓}12.03\%) & 0.74 (\textcolor[rgb]{0,0.5,0}{↓}11.46\%)\\
& & \textbf{+Weighted\_BOLD} & 40.07 (\textcolor[rgb]{0,0.5,0}{↑}0.56\%) & 40.08 (\textcolor[rgb]{0,0.5,0}{↑}0.66\%) & 3.64 (\textcolor[rgb]{0,0.5,0}{↓}22.05\%) & 1.48 (\textcolor[rgb]{0,0.5,0}{↓}11.82\%) & 0.67 (\textcolor[rgb]{0,0.5,0}{↓}20.11\%)\\
& & \textbf{-/+Weighted\_BOLD} & 40.04 (\textcolor[rgb]{0,0.5,0}{↑}0.46\%) & 40.06 (\textcolor[rgb]{0,0.5,0}{↑}0.63\%) & 2.31 (\textcolor[rgb]{0,0.5,0}{↓}50.5\%) & 1.36 (\textcolor[rgb]{0,0.5,0}{↓}18.83\%) & 0.3 (\textcolor[rgb]{0,0.5,0}{↓}63.83\%)\\
\cline{2-8}
\hline
\end{tabular}
}}
\label{tab:25}
\end{table*}

\begin{table*}[ht]
\centering
\caption{Comparison of BOLD and Weighted\_BOLD bias mitigation approaches across models and datasets for performance and bias monitoring metrics with $k$ = 0.5. Green arrows indicate improvements: upward for Accuracy and F1\_mean, and downward for standard deviation metrics. Red arrows represent deterioration, respectively.}
\footnotesize{
\resizebox{\textwidth}{!}{
\begin{tabular}{l l p{2.7cm} r p{2.2cm} p{2.2cm} p{2.2cm} p{2.2cm}}
\hline
\multirow{2}{*}{\textbf{Model}} & \multirow{2}{*}{\textbf{Dataset}} & \multirow{2}{*}{\textbf{Configuration}} & \multicolumn{2}{c}{\textbf{Performance Metrics}} & \multicolumn{3}{c}{\textbf{Bias Monitoring Metrics}} \\ 
\cline{4-5} \cline{6-8}
& & & \textbf{Accuracy} & \textbf{F1\_mean} & \textbf{Recall\_std} & \textbf{F1\_std} & \textbf{JS\_std} \\ 
\hline
\multirow{4}{*}{} & NExT-QA & \textbf{Default} & 44.79 & 40.85  & 23.83  & 15.86 & 16.09\\
& & \textbf{BOLD} & 45.88 (\textcolor[rgb]{0,0.5,0}{↑}2.43\%) & 42.15 (\textcolor[rgb]{0,0.5,0}{↑}3.18\%) & 22.98 (\textcolor[rgb]{0,0.5,0}{↓}3.53\%) & 15.53 (\textcolor[rgb]{0,0.5,0}{↓}2.11\%) & 15.55 (\textcolor[rgb]{0,0.5,0}{↓}3.31\%)\\
& & \textbf{+Weighted\_BOLD} & 45.91 (\textcolor[rgb]{0,0.5,0}{↑}2.51\%) & 42.2 (\textcolor[rgb]{0,0.5,0}{↑}3.29\%) & 22.83 (\textcolor[rgb]{0,0.5,0}{↓}4.19\%) & 15.51 (\textcolor[rgb]{0,0.5,0}{↓}2.2\%) & 15.56 (\textcolor[rgb]{0,0.5,0}{↓}3.27\%)\\
& & \textbf{-/+Weighted\_BOLD} & 48.25 (\textcolor[rgb]{0,0.5,0}{↑}7.73\%) & 46.52 (\textcolor[rgb]{0,0.5,0}{↑}13.86\%) & 16.6 (\textcolor[rgb]{0,0.5,0}{↓}30.32\%) & 10.6 (\textcolor[rgb]{0,0.5,0}{↓}33.17\%) & 13.27 (\textcolor[rgb]{0,0.5,0}{↓}17.48\%)\\
\cline{2-8}
\multirow{4}{*}{Video-LLaMA} & STAR & \textbf{Default} & 36.52 & 31.86  & 24.1  & 15.44 & 15.29\\
& & \textbf{BOLD} & 37.19 (\textcolor[rgb]{0,0.5,0}{↑}1.82\%) & 33.02 (\textcolor[rgb]{0,0.5,0}{↑}3.65\%) & 22.29 (\textcolor[rgb]{0,0.5,0}{↓}7.5\%) & 14.55 (\textcolor[rgb]{0,0.5,0}{↓}5.75\%) & 14.58 (\textcolor[rgb]{0,0.5,0}{↓}4.66\%)\\
& & \textbf{+Weighted\_BOLD} & 37.34 (\textcolor[rgb]{0,0.5,0}{↑}2.24\%) & 33.5 (\textcolor[rgb]{0,0.5,0}{↑}5.16\%) & 21.13 (\textcolor[rgb]{0,0.5,0}{↓}12.3\%) & 14.05 (\textcolor[rgb]{0,0.5,0}{↓}8.98\%) & 14.14 (\textcolor[rgb]{0,0.5,0}{↓}7.54\%)\\
& & \textbf{-/+Weighted\_BOLD} & 38.61 (\textcolor[rgb]{0,0.5,0}{↑}5.72\%) & 38.06 (\textcolor[rgb]{0,0.5,0}{↑}19.47\%) & 10.18 (\textcolor[rgb]{0,0.5,0}{↓}57.75\%) & 4.79 (\textcolor[rgb]{0,0.5,0}{↓}68.99\%) & 8.58 (\textcolor[rgb]{0,0.5,0}{↓}43.89\%)\\
\cline{2-8}
\multirow{4}{*}{} & Perception Test & \textbf{Default} & 41.39 & 37.19  & 27.06  & 11.4 & 16.06\\
& & \textbf{BOLD} & 41.9 (\textcolor[rgb]{0,0.5,0}{↑}1.24\%) & 38.68 (\textcolor[rgb]{0,0.5,0}{↑}4.01\%) & 24.05 (\textcolor[rgb]{0,0.5,0}{↓}11.11\%) & 9.87 (\textcolor[rgb]{0,0.5,0}{↓}13.45\%) & 14.45 (\textcolor[rgb]{0,0.5,0}{↓}10.01\%)\\
& & \textbf{+Weighted\_BOLD} & 42.06 (\textcolor[rgb]{0,0.5,0}{↑}1.62\%) & 39.36 (\textcolor[rgb]{0,0.5,0}{↑}5.86\%) & 21.8 (\textcolor[rgb]{0,0.5,0}{↓}19.45\%) & 9.11 (\textcolor[rgb]{0,0.5,0}{↓}20.06\%) & 13.02 (\textcolor[rgb]{0,0.5,0}{↓}18.94\%)\\
& & \textbf{-/+Weighted\_BOLD} & 42.27 (\textcolor[rgb]{0,0.5,0}{↑}2.13\%) & 41.24 (\textcolor[rgb]{0,0.5,0}{↑}10.9\%) & 13.55 (\textcolor[rgb]{0,0.5,0}{↓}49.93\%) & 5.64 (\textcolor[rgb]{0,0.5,0}{↓}50.54\%) & 9.73 (\textcolor[rgb]{0,0.5,0}{↓}39.41\%)\\
\cline{2-8}
\multirow{4}{*}{} & Video-MME & \textbf{Default} & 32.05 & 28.15  & 20.62  & 12.52 & 13.75\\
& & \textbf{BOLD} & 32.73 (\textcolor[rgb]{0,0.5,0}{↑}2.13\%) & 29.23 (\textcolor[rgb]{0,0.5,0}{↑}3.85\%) & 19.29 (\textcolor[rgb]{0,0.5,0}{↓}6.45\%) & 11.96 (\textcolor[rgb]{0,0.5,0}{↓}4.46\%) & 13.1 (\textcolor[rgb]{0,0.5,0}{↓}4.73\%)\\
& & \textbf{+Weighted\_BOLD} & 32.2 (\textcolor[rgb]{0,0.5,0}{↑}0.47\%) & 28.98 (\textcolor[rgb]{0,0.5,0}{↑}2.94\%) & 17.65 (\textcolor[rgb]{0,0.5,0}{↓}14.38\%) & 11.69 (\textcolor[rgb]{0,0.5,0}{↓}6.56\%) & 12.64 (\textcolor[rgb]{0,0.5,0}{↓}8.04\%)\\
& & \textbf{-/+Weighted\_BOLD} & 32.88 (\textcolor[rgb]{0,0.5,0}{↑}2.6\%) & 31.21 (\textcolor[rgb]{0,0.5,0}{↑}10.87\%) & 12.77 (\textcolor[rgb]{0,0.5,0}{↓}38.06\%) & 8.02 (\textcolor[rgb]{0,0.5,0}{↓}35.91\%) & 10.42 (\textcolor[rgb]{0,0.5,0}{↓}24.25\%)\\
\cline{1-8}
\multirow{4}{*}{} & NExT-QA & \textbf{Default} & 49.95 & 49.81  & 16.31  & 3.27 & 5.4\\
& & \textbf{BOLD} & 51.81 (\textcolor[rgb]{0,0.5,0}{↑}3.72\%) & 51.71 (\textcolor[rgb]{0,0.5,0}{↑}3.82\%) & 13.27 (\textcolor[rgb]{0,0.5,0}{↓}18.63\%) & 2.67 (\textcolor[rgb]{0,0.5,0}{↓}18.42\%) & 4.58 (\textcolor[rgb]{0,0.5,0}{↓}15.06\%)\\
& & \textbf{+Weighted\_BOLD} & 52.15 (\textcolor[rgb]{0,0.5,0}{↑}4.39\%) & 52.04 (\textcolor[rgb]{0,0.5,0}{↑}4.49\%) & 12.72 (\textcolor[rgb]{0,0.5,0}{↓}22.02\%) & 2.62 (\textcolor[rgb]{0,0.5,0}{↓}19.83\%) & 4.49 (\textcolor[rgb]{0,0.5,0}{↓}16.83\%)\\
& & \textbf{-/+Weighted\_BOLD} & 53.65 (\textcolor[rgb]{0,0.5,0}{↑}7.41\%) & 53.53 (\textcolor[rgb]{0,0.5,0}{↑}7.48\%) & 8.49 (\textcolor[rgb]{0,0.5,0}{↓}47.97\%) & 3.15 (\textcolor[rgb]{0,0.5,0}{↓}3.61\%) & 3.78 (\textcolor[rgb]{0,0.5,0}{↓}29.87\%)\\
\cline{2-8}
\multirow{4}{*}{Video-LLaVA} & STAR & \textbf{Default} & 34.77 & 31.83  & 24.99  & 6.82 & 5.78\\
& & \textbf{BOLD} & 37.21 (\textcolor[rgb]{0,0.5,0}{↑}7.01\%) & 35.76 (\textcolor[rgb]{0,0.5,0}{↑}12.37\%) & 18.28 (\textcolor[rgb]{0,0.5,0}{↓}26.85\%) & 4.97 (\textcolor[rgb]{0,0.5,0}{↓}27.15\%) & 4.54 (\textcolor[rgb]{0,0.5,0}{↓}21.39\%)\\
& & \textbf{+Weighted\_BOLD} & 37.53 (\textcolor[rgb]{0,0.5,0}{↑}7.94\%) & 36.18 (\textcolor[rgb]{0,0.5,0}{↑}13.69\%) & 17.45 (\textcolor[rgb]{0,0.5,0}{↓}30.15\%) & 4.91 (\textcolor[rgb]{0,0.5,0}{↓}28.03\%) & 4.26 (\textcolor[rgb]{0,0.5,0}{↓}26.24\%)\\
& & \textbf{-/+Weighted\_BOLD} & 38.31 (\textcolor[rgb]{0,0.5,0}{↑}10.17\%) & 37.39 (\textcolor[rgb]{0,0.5,0}{↑}17.47\%) & 13.26 (\textcolor[rgb]{0,0.5,0}{↓}46.92\%) & 4.59 (\textcolor[rgb]{0,0.5,0}{↓}32.59\%) & 3.25 (\textcolor[rgb]{0,0.5,0}{↓}43.67\%)\\
\cline{2-8}
\multirow{4}{*}{} & Perception Test & \textbf{Default} & 40.72 & 35.69  & 27.5  & 14.75 & 5.28\\
& & \textbf{BOLD} & 41.62 (\textcolor[rgb]{0,0.5,0}{↑}2.22\%) & 38.69 (\textcolor[rgb]{0,0.5,0}{↑}8.4\%) & 21.75 (\textcolor[rgb]{0,0.5,0}{↓}20.9\%) & 10.92 (\textcolor[rgb]{0,0.5,0}{↓}25.97\%) & 3.78 (\textcolor[rgb]{0,0.5,0}{↓}28.4\%)\\
& & \textbf{+Weighted\_BOLD} & 41.95 (\textcolor[rgb]{0,0.5,0}{↑}3.02\%) & 39.46 (\textcolor[rgb]{0,0.5,0}{↑}10.58\%) & 19.73 (\textcolor[rgb]{0,0.5,0}{↓}28.26\%) & 10.24 (\textcolor[rgb]{0,0.5,0}{↓}30.53\%) & 3.4 (\textcolor[rgb]{0,0.5,0}{↓}35.51\%)\\
& & \textbf{-/+Weighted\_BOLD} & 41.62 (\textcolor[rgb]{0,0.5,0}{↑}2.22\%) & 40.88 (\textcolor[rgb]{0,0.5,0}{↑}14.56\%) & 11.01 (\textcolor[rgb]{0,0.5,0}{↓}59.96\%) & 5.44 (\textcolor[rgb]{0,0.5,0}{↓}63.11\%) & 2.59 (\textcolor[rgb]{0,0.5,0}{↓}50.97\%)\\
\cline{2-8}
\multirow{4}{*}{} & Video-MME & \textbf{Default} & 34.3 & 30.99  & 24.09  & 8.06 & 4.79\\
& & \textbf{BOLD} & 34.7 (\textcolor[rgb]{0,0.5,0}{↑}1.19\%) & 32.79 (\textcolor[rgb]{0,0.5,0}{↑}5.81\%) & 18.19 (\textcolor[rgb]{0,0.5,0}{↓}24.51\%) & 6.16 (\textcolor[rgb]{0,0.5,0}{↓}23.63\%) & 3.84 (\textcolor[rgb]{0,0.5,0}{↓}19.93\%)\\
& & \textbf{+Weighted\_BOLD} & 34.63 (\textcolor[rgb]{0,0.5,0}{↑}0.97\%) & 32.97 (\textcolor[rgb]{0,0.5,0}{↑}6.38\%) & 16.36 (\textcolor[rgb]{0,0.5,0}{↓}32.07\%) & 6.01 (\textcolor[rgb]{0,0.5,0}{↓}25.43\%) & 3.43 (\textcolor[rgb]{0,0.5,0}{↓}28.5\%)\\
& & \textbf{-/+Weighted\_BOLD} & 34.0 (\textcolor{red}{↓}0.86\%) & 32.6 (\textcolor[rgb]{0,0.5,0}{↑}5.21\%) & 14.11 (\textcolor[rgb]{0,0.5,0}{↓}41.43\%) & 5.79 (\textcolor[rgb]{0,0.5,0}{↓}28.18\%) & 2.92 (\textcolor[rgb]{0,0.5,0}{↓}39.01\%)\\
\cline{1-8}
\multirow{4}{*} & NExT-QA & \textbf{Default} & 63.91 & 63.88  & 2.15  & 1.32 & 1.99\\
& & \textbf{BOLD} & 63.92 (\textcolor[rgb]{0,0.5,0}{↑}0.02\%) & 63.89 (\textcolor[rgb]{0,0.5,0}{↑}0.02\%) & 2.03 (\textcolor[rgb]{0,0.5,0}{↓}5.47\%) & 1.18 (\textcolor[rgb]{0,0.5,0}{↓}10.4\%) & 1.99 (\textcolor[rgb]{0,0.5,0}{↓}0.17\%)\\
& & \textbf{+Weighted\_BOLD} & 63.93 (\textcolor[rgb]{0,0.5,0}{↑}0.04\%) & 63.91 (\textcolor[rgb]{0,0.5,0}{↑}0.04\%) & 1.99 (\textcolor[rgb]{0,0.5,0}{↓}7.64\%) & 1.19 (\textcolor[rgb]{0,0.5,0}{↓}9.83\%) & 1.99 (\textcolor[rgb]{0,0.5,0}{↓}0.04\%)\\
& & \textbf{-/+Weighted\_BOLD} & 64.01 (\textcolor[rgb]{0,0.5,0}{↑}0.16\%) & 64.0 (\textcolor[rgb]{0,0.5,0}{↑}0.19\%) & 1.14 (\textcolor[rgb]{0,0.5,0}{↓}47.18\%) & 1.17 (\textcolor[rgb]{0,0.5,0}{↓}11.43\%) & 2.01 (\textcolor{red}{↑}1.1\%)\\
\cline{2-8}
\multirow{4}{*}{SeViLA} & STAR & \textbf{Default} & 46.28 & 46.14  & 4.47  & 2.3 & 2.25\\
& & \textbf{BOLD} & 46.22 (\textcolor{red}{↓}0.12\%) & 46.1 (\textcolor{red}{↓}0.08\%) & 4.13 (\textcolor[rgb]{0,0.5,0}{↓}7.46\%) & 2.26 (\textcolor[rgb]{0,0.5,0}{↓}1.78\%) & 2.1 (\textcolor[rgb]{0,0.5,0}{↓}6.64\%)\\
& & \textbf{+Weighted\_BOLD} & 46.2 (\textcolor{red}{↓}0.18\%) & 46.08 (\textcolor{red}{↓}0.13\%) & 4.01 (\textcolor[rgb]{0,0.5,0}{↓}10.17\%) & 2.2 (\textcolor[rgb]{0,0.5,0}{↓}4.36\%) & 2.07 (\textcolor[rgb]{0,0.5,0}{↓}8.2\%)\\
& & \textbf{-/+Weighted\_BOLD} & 46.38 (\textcolor[rgb]{0,0.5,0}{↑}0.21\%) & 46.3 (\textcolor[rgb]{0,0.5,0}{↑}0.34\%) & 3.41 (\textcolor[rgb]{0,0.5,0}{↓}23.66\%) & 1.94 (\textcolor[rgb]{0,0.5,0}{↓}15.89\%) & 1.93 (\textcolor[rgb]{0,0.5,0}{↓}14.33\%)\\
\cline{2-8}
\multirow{4}{*}{} & Perception Test & \textbf{Default} & 45.3 & 45.11  & 6.21  & 3.02 & 1.59\\
& & \textbf{BOLD} & 45.32 (\textcolor[rgb]{0,0.5,0}{↑}0.06\%) & 45.18 (\textcolor[rgb]{0,0.5,0}{↑}0.16\%) & 5.18 (\textcolor[rgb]{0,0.5,0}{↓}16.61\%) & 2.95 (\textcolor[rgb]{0,0.5,0}{↓}2.43\%) & 1.3 (\textcolor[rgb]{0,0.5,0}{↓}18.42\%)\\
& & \textbf{+Weighted\_BOLD} & 45.31 (\textcolor[rgb]{0,0.5,0}{↑}0.03\%) & 45.2 (\textcolor[rgb]{0,0.5,0}{↑}0.21\%) & 4.56 (\textcolor[rgb]{0,0.5,0}{↓}26.58\%) & 2.83 (\textcolor[rgb]{0,0.5,0}{↓}6.43\%) & 1.08 (\textcolor[rgb]{0,0.5,0}{↓}31.92\%)\\
& & \textbf{-/+Weighted\_BOLD} & 45.25 (\textcolor{red}{↓}0.12\%) & 45.2 (\textcolor[rgb]{0,0.5,0}{↑}0.2\%) & 3.25 (\textcolor[rgb]{0,0.5,0}{↓}47.69\%) & 2.28 (\textcolor[rgb]{0,0.5,0}{↓}24.52\%) & 0.71 (\textcolor[rgb]{0,0.5,0}{↓}55.13\%)\\
\cline{2-8}
\multirow{4}{*}{} & Video-MME & \textbf{Default} & 39.85 & 39.82  & 4.67  & 1.68 & 0.84\\
& & \textbf{BOLD} & 40.19 (\textcolor[rgb]{0,0.5,0}{↑}0.84\%) & 40.17 (\textcolor[rgb]{0,0.5,0}{↑}0.88\%) & 4.11 (\textcolor[rgb]{0,0.5,0}{↓}12.03\%) & 1.41 (\textcolor[rgb]{0,0.5,0}{↓}16.19\%) & 0.73 (\textcolor[rgb]{0,0.5,0}{↓}12.52\%)\\
& & \textbf{+Weighted\_BOLD} & 40.04 (\textcolor[rgb]{0,0.5,0}{↑}0.46\%) & 40.03 (\textcolor[rgb]{0,0.5,0}{↑}0.54\%) & 3.78 (\textcolor[rgb]{0,0.5,0}{↓}19.14\%) & 1.44 (\textcolor[rgb]{0,0.5,0}{↓}14.6\%) & 0.69 (\textcolor[rgb]{0,0.5,0}{↓}17.22\%)\\
& & \textbf{-/+Weighted\_BOLD} & 39.81 (\textcolor{red}{↓}0.09\%) & 39.83 (\textcolor[rgb]{0,0.5,0}{↑}0.04\%) & 2.99 (\textcolor[rgb]{0,0.5,0}{↓}35.93\%) & 1.43 (\textcolor[rgb]{0,0.5,0}{↓}14.97\%) & 0.45 (\textcolor[rgb]{0,0.5,0}{↓}45.74\%)\\
\cline{2-8}
\hline
\end{tabular}
}}
\end{table*}

\begin{table*}[ht]
\centering
\caption{Comparison of BOLD and Weighted\_BOLD bias mitigation approaches across models and datasets for performance and bias monitoring metrics with $k$ = 0.75. Green arrows indicate improvements: upward for Accuracy and F1\_mean, and downward for standard deviation metrics. Red arrows represent deterioration for Accuracy and F1\_mean.}
\footnotesize{
\resizebox{\textwidth}{!}{
\begin{tabular}{l l p{2.7cm} r p{2.2cm} p{2.2cm} p{2.2cm} p{2.2cm}}
\hline
\multirow{2}{*}{\textbf{Model}} & \multirow{2}{*}{\textbf{Dataset}} & \multirow{2}{*}{\textbf{Configuration}} & \multicolumn{2}{c}{\textbf{Performance Metrics}} & \multicolumn{3}{c}{\textbf{Bias Monitoring Metrics}} \\ 
\cline{4-5} \cline{6-8}
& & & \textbf{Accuracy} & \textbf{F1\_mean} & \textbf{Recall\_std} & \textbf{F1\_std} & \textbf{JS\_std} \\ 
\hline
\multirow{4}{*}{} & NExT-QA & \textbf{Default} & 44.79 & 40.85  & 23.83  & 15.86 & 16.09\\
& & \textbf{BOLD} & 45.87 (\textcolor[rgb]{0,0.5,0}{↑}2.4\%) & 42.16 (\textcolor[rgb]{0,0.5,0}{↑}3.19\%) & 22.96 (\textcolor[rgb]{0,0.5,0}{↓}3.63\%) & 15.49 (\textcolor[rgb]{0,0.5,0}{↓}2.37\%) & 15.56 (\textcolor[rgb]{0,0.5,0}{↓}3.28\%)\\
& & \textbf{+Weighted\_BOLD} & 45.91 (\textcolor[rgb]{0,0.5,0}{↑}2.51\%) & 42.2 (\textcolor[rgb]{0,0.5,0}{↑}3.29\%) & 22.83 (\textcolor[rgb]{0,0.5,0}{↓}4.19\%) & 15.51 (\textcolor[rgb]{0,0.5,0}{↓}2.2\%) & 15.57 (\textcolor[rgb]{0,0.5,0}{↓}3.23\%)\\
& & \textbf{-/+Weighted\_BOLD} & 47.56 (\textcolor[rgb]{0,0.5,0}{↑}6.19\%) & 46.63 (\textcolor[rgb]{0,0.5,0}{↑}14.14\%) & 13.36 (\textcolor[rgb]{0,0.5,0}{↓}43.91\%) & 8.42 (\textcolor[rgb]{0,0.5,0}{↓}46.91\%) & 11.79 (\textcolor[rgb]{0,0.5,0}{↓}26.73\%)\\
\cline{2-8}
\multirow{4}{*}{Video-LLaMA} & STAR & \textbf{Default} & 36.52 & 31.86  & 24.1  & 15.44 & 15.29\\
& & \textbf{BOLD} & 37.19 (\textcolor[rgb]{0,0.5,0}{↑}1.82\%) & 33.02 (\textcolor[rgb]{0,0.5,0}{↑}3.65\%) & 22.29 (\textcolor[rgb]{0,0.5,0}{↓}7.5\%) & 14.55 (\textcolor[rgb]{0,0.5,0}{↓}5.75\%) & 14.59 (\textcolor[rgb]{0,0.5,0}{↓}4.59\%)\\
& & \textbf{+Weighted\_BOLD} & 37.34 (\textcolor[rgb]{0,0.5,0}{↑}2.24\%) & 33.5 (\textcolor[rgb]{0,0.5,0}{↑}5.16\%) & 21.13 (\textcolor[rgb]{0,0.5,0}{↓}12.3\%) & 14.05 (\textcolor[rgb]{0,0.5,0}{↓}8.98\%) & 14.13 (\textcolor[rgb]{0,0.5,0}{↓}7.56\%)\\
& & \textbf{-/+Weighted\_BOLD} & 38.74 (\textcolor[rgb]{0,0.5,0}{↑}6.07\%) & 38.14 (\textcolor[rgb]{0,0.5,0}{↑}19.74\%) & 10.58 (\textcolor[rgb]{0,0.5,0}{↓}56.11\%) & 4.94 (\textcolor[rgb]{0,0.5,0}{↓}68.03\%) & 9.64 (\textcolor[rgb]{0,0.5,0}{↓}36.94\%)\\
\cline{2-8}
\multirow{4}{*}{} & Perception Test & \textbf{Default} & 41.39 & 37.19  & 27.06  & 11.4 & 16.06\\
& & \textbf{BOLD} & 41.9 (\textcolor[rgb]{0,0.5,0}{↑}1.24\%) & 38.67 (\textcolor[rgb]{0,0.5,0}{↑}3.98\%) & 24.09 (\textcolor[rgb]{0,0.5,0}{↓}11.0\%) & 9.88 (\textcolor[rgb]{0,0.5,0}{↓}13.29\%) & 14.46 (\textcolor[rgb]{0,0.5,0}{↓}9.97\%)\\
& & \textbf{+Weighted\_BOLD} & 42.06 (\textcolor[rgb]{0,0.5,0}{↑}1.62\%) & 39.36 (\textcolor[rgb]{0,0.5,0}{↑}5.86\%) & 21.8 (\textcolor[rgb]{0,0.5,0}{↓}19.45\%) & 9.11 (\textcolor[rgb]{0,0.5,0}{↓}20.06\%) & 13.12 (\textcolor[rgb]{0,0.5,0}{↓}18.33\%)\\
& & \textbf{-/+Weighted\_BOLD} & 42.29 (\textcolor[rgb]{0,0.5,0}{↑}2.16\%) & 41.32 (\textcolor[rgb]{0,0.5,0}{↑}11.13\%) & 12.96 (\textcolor[rgb]{0,0.5,0}{↓}52.13\%) & 5.53 (\textcolor[rgb]{0,0.5,0}{↓}51.52\%) & 9.56 (\textcolor[rgb]{0,0.5,0}{↓}40.45\%)\\
\cline{2-8}
\multirow{4}{*}{} & Video-MME & \textbf{Default} & 32.05 & 28.15  & 20.62  & 12.52 & 13.75\\
& & \textbf{BOLD} & 32.69 (\textcolor[rgb]{0,0.5,0}{↑}2.01\%) & 29.2 (\textcolor[rgb]{0,0.5,0}{↑}3.75\%) & 19.24 (\textcolor[rgb]{0,0.5,0}{↓}6.7\%) & 11.94 (\textcolor[rgb]{0,0.5,0}{↓}4.63\%) & 13.09 (\textcolor[rgb]{0,0.5,0}{↓}4.8\%)\\
& & \textbf{+Weighted\_BOLD} & 32.2 (\textcolor[rgb]{0,0.5,0}{↑}0.47\%) & 28.97 (\textcolor[rgb]{0,0.5,0}{↑}2.94\%) & 17.65 (\textcolor[rgb]{0,0.5,0}{↓}14.38\%) & 11.7 (\textcolor[rgb]{0,0.5,0}{↓}6.54\%) & 12.56 (\textcolor[rgb]{0,0.5,0}{↓}8.62\%)\\
& & \textbf{-/+Weighted\_BOLD} & 32.65 (\textcolor[rgb]{0,0.5,0}{↑}1.89\%) & 30.55 (\textcolor[rgb]{0,0.5,0}{↑}8.55\%) & 13.94 (\textcolor[rgb]{0,0.5,0}{↓}32.38\%) & 9.24 (\textcolor[rgb]{0,0.5,0}{↓}26.17\%) & 11.04 (\textcolor[rgb]{0,0.5,0}{↓}19.73\%)\\
\cline{1-8}
\multirow{4}{*}{} & NExT-QA & \textbf{Default} & 49.95 & 49.81  & 16.31  & 3.27 & 5.4\\
& & \textbf{BOLD} & 51.81 (\textcolor[rgb]{0,0.5,0}{↑}3.72\%) & 51.71 (\textcolor[rgb]{0,0.5,0}{↑}3.82\%) & 13.27 (\textcolor[rgb]{0,0.5,0}{↓}18.63\%) & 2.67 (\textcolor[rgb]{0,0.5,0}{↓}18.42\%) & 4.58 (\textcolor[rgb]{0,0.5,0}{↓}15.14\%)\\
& & \textbf{+Weighted\_BOLD} & 51.96 (\textcolor[rgb]{0,0.5,0}{↑}4.02\%) & 51.86 (\textcolor[rgb]{0,0.5,0}{↑}4.12\%) & 12.92 (\textcolor[rgb]{0,0.5,0}{↓}20.76\%) & 2.64 (\textcolor[rgb]{0,0.5,0}{↓}19.44\%) & 4.48 (\textcolor[rgb]{0,0.5,0}{↓}17.02\%)\\
& & \textbf{-/+Weighted\_BOLD} & 53.75 (\textcolor[rgb]{0,0.5,0}{↑}7.6\%) & 53.66 (\textcolor[rgb]{0,0.5,0}{↑}7.73\%) & 8.51 (\textcolor[rgb]{0,0.5,0}{↓}47.82\%) & 2.58 (\textcolor[rgb]{0,0.5,0}{↓}21.07\%) & 3.6 (\textcolor[rgb]{0,0.5,0}{↓}33.28\%)\\
\cline{2-8}
\multirow{4}{*}{Video-LLaVA} & STAR & \textbf{Default} & 34.77 & 31.83  & 24.99  & 6.82 & 5.78\\
& & \textbf{BOLD} & 37.21 (\textcolor[rgb]{0,0.5,0}{↑}7.01\%) & 35.76 (\textcolor[rgb]{0,0.5,0}{↑}12.37\%) & 18.28 (\textcolor[rgb]{0,0.5,0}{↓}26.85\%) & 4.97 (\textcolor[rgb]{0,0.5,0}{↓}27.15\%) & 4.62 (\textcolor[rgb]{0,0.5,0}{↓}20.03\%)\\
& & \textbf{+Weighted\_BOLD} & 37.52 (\textcolor[rgb]{0,0.5,0}{↑}7.9\%) & 36.17 (\textcolor[rgb]{0,0.5,0}{↑}13.64\%) & 17.45 (\textcolor[rgb]{0,0.5,0}{↓}30.14\%) & 4.91 (\textcolor[rgb]{0,0.5,0}{↓}28.02\%) & 4.27 (\textcolor[rgb]{0,0.5,0}{↓}26.1\%)\\
& & \textbf{-/+Weighted\_BOLD} & 37.86 (\textcolor[rgb]{0,0.5,0}{↑}8.87\%) & 36.87 (\textcolor[rgb]{0,0.5,0}{↑}15.85\%) & 13.12 (\textcolor[rgb]{0,0.5,0}{↓}47.51\%) & 4.68 (\textcolor[rgb]{0,0.5,0}{↓}31.37\%) & 3.09 (\textcolor[rgb]{0,0.5,0}{↓}46.45\%)\\
\cline{2-8}
\multirow{4}{*}{} & Perception Test & \textbf{Default} & 40.72 & 35.69  & 27.5  & 14.75 & 5.28\\
& & \textbf{BOLD} & 41.62 (\textcolor[rgb]{0,0.5,0}{↑}2.22\%) & 38.68 (\textcolor[rgb]{0,0.5,0}{↑}8.38\%) & 21.77 (\textcolor[rgb]{0,0.5,0}{↓}20.85\%) & 10.93 (\textcolor[rgb]{0,0.5,0}{↓}25.88\%) & 3.79 (\textcolor[rgb]{0,0.5,0}{↓}28.21\%)\\
& & \textbf{+Weighted\_BOLD} & 41.95 (\textcolor[rgb]{0,0.5,0}{↑}3.02\%) & 39.46 (\textcolor[rgb]{0,0.5,0}{↑}10.58\%) & 19.73 (\textcolor[rgb]{0,0.5,0}{↓}28.26\%) & 10.24 (\textcolor[rgb]{0,0.5,0}{↓}30.53\%) & 3.41 (\textcolor[rgb]{0,0.5,0}{↓}35.4\%)\\
& & \textbf{-/+Weighted\_BOLD} & 42.21 (\textcolor[rgb]{0,0.5,0}{↑}3.66\%) & 41.64 (\textcolor[rgb]{0,0.5,0}{↑}16.67\%) & 9.47 (\textcolor[rgb]{0,0.5,0}{↓}65.56\%) & 5.14 (\textcolor[rgb]{0,0.5,0}{↓}65.14\%) & 1.8 (\textcolor[rgb]{0,0.5,0}{↓}65.84\%)\\
\cline{2-8}
\multirow{4}{*}{} & Video-MME & \textbf{Default} & 34.3 & 30.99  & 24.09  & 8.06 & 4.79\\
& & \textbf{BOLD} & 34.63 (\textcolor[rgb]{0,0.5,0}{↑}0.97\%) & 32.64 (\textcolor[rgb]{0,0.5,0}{↑}5.33\%) & 18.59 (\textcolor[rgb]{0,0.5,0}{↓}22.84\%) & 6.19 (\textcolor[rgb]{0,0.5,0}{↓}23.2\%) & 3.85 (\textcolor[rgb]{0,0.5,0}{↓}19.65\%)\\
& & \textbf{+Weighted\_BOLD} & 34.48 (\textcolor[rgb]{0,0.5,0}{↑}0.54\%) & 32.89 (\textcolor[rgb]{0,0.5,0}{↑}6.13\%) & 15.94 (\textcolor[rgb]{0,0.5,0}{↓}33.83\%) & 5.86 (\textcolor[rgb]{0,0.5,0}{↓}27.32\%) & 3.37 (\textcolor[rgb]{0,0.5,0}{↓}29.66\%)\\
& & \textbf{-/+Weighted\_BOLD} & 33.78 (\textcolor{red}{↓}1.51\%) & 33.2 (\textcolor[rgb]{0,0.5,0}{↑}7.12\%) & 9.23 (\textcolor[rgb]{0,0.5,0}{↓}61.7\%) & 3.35 (\textcolor[rgb]{0,0.5,0}{↓}58.46\%) & 1.42 (\textcolor[rgb]{0,0.5,0}{↓}70.35\%)\\
\cline{1-8}
\multirow{4}{*}{} & NExT-QA & \textbf{Default} & 63.91 & 63.88  & 2.15  & 1.32 & 1.99\\
& & \textbf{BOLD} & 63.91 (\textcolor[rgb]{0,0.5,0}{↑}0.0\%) & 63.88 (\textcolor[rgb]{0,0.5,0}{↑}0.01\%) & 2.03 (\textcolor[rgb]{0,0.5,0}{↓}5.45\%) & 1.18 (\textcolor[rgb]{0,0.5,0}{↓}10.77\%) & 1.98 (\textcolor[rgb]{0,0.5,0}{↓}0.34\%)\\
& & \textbf{+Weighted\_BOLD} & 63.93 (\textcolor[rgb]{0,0.5,0}{↑}0.04\%) & 63.91 (\textcolor[rgb]{0,0.5,0}{↑}0.04\%) & 2.0 (\textcolor[rgb]{0,0.5,0}{↓}6.83\%) & 1.2 (\textcolor[rgb]{0,0.5,0}{↓}8.97\%) & 1.98 (\textcolor[rgb]{0,0.5,0}{↓}0.23\%)\\
& & \textbf{-/+Weighted\_BOLD} & 64.11 (\textcolor[rgb]{0,0.5,0}{↑}0.31\%) & 64.09 (\textcolor[rgb]{0,0.5,0}{↑}0.34\%) & 1.2 (\textcolor[rgb]{0,0.5,0}{↓}44.36\%) & 1.09 (\textcolor[rgb]{0,0.5,0}{↓}17.41\%) & 2.15 (\textcolor{red}{↑}8.2\%)\\
\cline{2-8}
\multirow{4}{*}{SeViLA} & STAR & \textbf{Default} & 46.28 & 46.14  & 4.47  & 2.3 & 2.25\\
& & \textbf{BOLD} & 46.27 (\textcolor{red}{↓}0.03\%) & 46.15 (\textcolor[rgb]{0,0.5,0}{↑}0.02\%) & 4.07 (\textcolor[rgb]{0,0.5,0}{↓}8.91\%) & 2.22 (\textcolor[rgb]{0,0.5,0}{↓}3.53\%) & 2.08 (\textcolor[rgb]{0,0.5,0}{↓}7.72\%)\\
& & \textbf{+Weighted\_BOLD} & 46.17 (\textcolor{red}{↓}0.24\%) & 46.05 (\textcolor{red}{↓}0.19\%) & 4.01 (\textcolor[rgb]{0,0.5,0}{↓}10.17\%) & 2.2 (\textcolor[rgb]{0,0.5,0}{↓}4.43\%) & 2.04 (\textcolor[rgb]{0,0.5,0}{↓}9.14\%)\\
& & \textbf{-/+Weighted\_BOLD} & 46.17 (\textcolor{red}{↓}0.24\%) & 46.1 (\textcolor{red}{↓}0.09\%) & 2.96 (\textcolor[rgb]{0,0.5,0}{↓}33.63\%) & 1.78 (\textcolor[rgb]{0,0.5,0}{↓}22.67\%) & 1.64 (\textcolor[rgb]{0,0.5,0}{↓}27.25\%)\\
\cline{2-8}
\multirow{4}{*}{} & Perception Test & \textbf{Default} & 45.3 & 45.11  & 6.21  & 3.02 & 1.59\\
& & \textbf{BOLD} & 45.31 (\textcolor[rgb]{0,0.5,0}{↑}0.03\%) & 45.16 (\textcolor[rgb]{0,0.5,0}{↑}0.12\%) & 5.25 (\textcolor[rgb]{0,0.5,0}{↓}15.42\%) & 2.98 (\textcolor[rgb]{0,0.5,0}{↓}1.26\%) & 1.3 (\textcolor[rgb]{0,0.5,0}{↓}18.44\%)\\
& & \textbf{+Weighted\_BOLD} & 45.35 (\textcolor[rgb]{0,0.5,0}{↑}0.12\%) & 45.25 (\textcolor[rgb]{0,0.5,0}{↑}0.32\%) & 4.27 (\textcolor[rgb]{0,0.5,0}{↓}31.19\%) & 2.65 (\textcolor[rgb]{0,0.5,0}{↓}12.18\%) & 1.03 (\textcolor[rgb]{0,0.5,0}{↓}34.92\%)\\
& & \textbf{-/+Weighted\_BOLD} & 45.15 (\textcolor{red}{↓}0.32\%) & 45.09 (\textcolor{red}{↓}0.03\%) & 3.48 (\textcolor[rgb]{0,0.5,0}{↓}43.88\%) & 2.3 (\textcolor[rgb]{0,0.5,0}{↓}24.02\%) & 0.78 (\textcolor[rgb]{0,0.5,0}{↓}51.02\%)\\
\cline{2-8}
\multirow{4}{*}{} & Video-MME & \textbf{Default} & 39.85 & 39.82  & 4.67  & 1.68 & 0.84\\
& & \textbf{BOLD} & 40.04 (\textcolor[rgb]{0,0.5,0}{↑}0.46\%) & 40.03 (\textcolor[rgb]{0,0.5,0}{↑}0.53\%) & 3.84 (\textcolor[rgb]{0,0.5,0}{↓}17.85\%) & 1.46 (\textcolor[rgb]{0,0.5,0}{↓}13.42\%) & 0.69 (\textcolor[rgb]{0,0.5,0}{↓}18.0\%)\\
& & \textbf{+Weighted\_BOLD} & 39.89 (\textcolor[rgb]{0,0.5,0}{↑}0.09\%) & 39.87 (\textcolor[rgb]{0,0.5,0}{↑}0.15\%) & 3.7 (\textcolor[rgb]{0,0.5,0}{↓}20.9\%) & 1.43 (\textcolor[rgb]{0,0.5,0}{↓}14.96\%) & 0.58 (\textcolor[rgb]{0,0.5,0}{↓}31.13\%)\\
& & \textbf{-/+Weighted\_BOLD} & 39.7 (\textcolor{red}{↓}0.37\%) & 39.71 (\textcolor{red}{↓}0.26\%) & 2.69 (\textcolor[rgb]{0,0.5,0}{↓}42.53\%) & 1.42 (\textcolor[rgb]{0,0.5,0}{↓}15.75\%) & 0.15 (\textcolor[rgb]{0,0.5,0}{↓}81.72\%)\\
\cline{2-8}
\hline
\end{tabular}
}}
\end{table*}

\begin{table*}[ht]
\centering
\caption{Comparison of BOLD and Weighted\_BOLD bias mitigation approaches across models and datasets for performance and bias monitoring metrics $k$ = 1 (equal to the entire dataset). Green arrows indicate improvements: upward for Accuracy and F1\_mean, and downward for standard deviation metrics. Red arrows represent deterioration for Accuracy and F1\_mean.}
\footnotesize{
\resizebox{\textwidth}{!}{
\begin{tabular}{l l p{2.7cm} r p{2.2cm} p{2.2cm} p{2.2cm} p{2.2cm}}
\hline
\multirow{2}{*}{\textbf{Model}} & \multirow{2}{*}{\textbf{Dataset}} & \multirow{2}{*}{\textbf{Configuration}} & \multicolumn{2}{c}{\textbf{Performance Metrics}} & \multicolumn{3}{c}{\textbf{Bias Monitoring Metrics}} \\ 
\cline{4-5} \cline{6-8}
& & & \textbf{Accuracy} & \textbf{F1\_mean} & \textbf{Recall\_std} & \textbf{F1\_std} & \textbf{JS\_std} \\ 
\hline
\multirow{4}{*}{} & NExT-QA & \textbf{Default} & 44.79 & 40.85  & 23.83  & 15.86 & 16.09\\
& & \textbf{BOLD} & 45.87 (\textcolor[rgb]{0,0.5,0}{↑}2.4\%) & 42.16 (\textcolor[rgb]{0,0.5,0}{↑}3.19\%) & 22.96 (\textcolor[rgb]{0,0.5,0}{↓}3.63\%) & 15.49 (\textcolor[rgb]{0,0.5,0}{↓}2.37\%) & 15.56 (\textcolor[rgb]{0,0.5,0}{↓}3.29\%)\\
& & \textbf{+Weighted\_BOLD} & 45.91 (\textcolor[rgb]{0,0.5,0}{↑}2.51\%) & 42.2 (\textcolor[rgb]{0,0.5,0}{↑}3.29\%) & 22.83 (\textcolor[rgb]{0,0.5,0}{↓}4.19\%) & 15.51 (\textcolor[rgb]{0,0.5,0}{↓}2.2\%) & 15.57 (\textcolor[rgb]{0,0.5,0}{↓}3.23\%)\\
& & \textbf{-/+Weighted\_BOLD} & 47.53 (\textcolor[rgb]{0,0.5,0}{↑}6.11\%) & 46.62 (\textcolor[rgb]{0,0.5,0}{↑}14.1\%) & 13.25 (\textcolor[rgb]{0,0.5,0}{↓}44.39\%) & 8.46 (\textcolor[rgb]{0,0.5,0}{↓}46.68\%) & 11.78 (\textcolor[rgb]{0,0.5,0}{↓}26.76\%)\\
\cline{2-8}
\multirow{4}{*}{Video-LLaMA} & STAR & \textbf{Default} & 36.52 & 31.86  & 24.1  & 15.44 & 15.29\\
& & \textbf{BOLD} & 37.19 (\textcolor[rgb]{0,0.5,0}{↑}1.82\%) & 33.02 (\textcolor[rgb]{0,0.5,0}{↑}3.65\%) & 22.29 (\textcolor[rgb]{0,0.5,0}{↓}7.5\%) & 14.55 (\textcolor[rgb]{0,0.5,0}{↓}5.75\%) & 14.59 (\textcolor[rgb]{0,0.5,0}{↓}4.58\%)\\
& & \textbf{+Weighted\_BOLD} & 37.34 (\textcolor[rgb]{0,0.5,0}{↑}2.24\%) & 33.5 (\textcolor[rgb]{0,0.5,0}{↑}5.16\%) & 21.13 (\textcolor[rgb]{0,0.5,0}{↓}12.3\%) & 14.05 (\textcolor[rgb]{0,0.5,0}{↓}8.98\%) & 14.13 (\textcolor[rgb]{0,0.5,0}{↓}7.55\%)\\
& & \textbf{-/+Weighted\_BOLD} & 38.72 (\textcolor[rgb]{0,0.5,0}{↑}6.03\%) & 38.18 (\textcolor[rgb]{0,0.5,0}{↑}19.85\%) & 10.0 (\textcolor[rgb]{0,0.5,0}{↓}58.49\%) & 4.85 (\textcolor[rgb]{0,0.5,0}{↓}68.6\%) & 8.53 (\textcolor[rgb]{0,0.5,0}{↓}44.23\%)\\
\cline{2-8}
\multirow{4}{*}{} & Perception Test & \textbf{Default} & 41.39 & 37.19  & 27.06  & 11.4 & 16.06\\
& & \textbf{BOLD} & 41.89 (\textcolor[rgb]{0,0.5,0}{↑}1.21\%) & 38.67 (\textcolor[rgb]{0,0.5,0}{↑}3.98\%) & 24.04 (\textcolor[rgb]{0,0.5,0}{↓}11.18\%) & 9.86 (\textcolor[rgb]{0,0.5,0}{↓}13.52\%) & 14.45 (\textcolor[rgb]{0,0.5,0}{↓}10.04\%)\\
& & \textbf{+Weighted\_BOLD} & 42.08 (\textcolor[rgb]{0,0.5,0}{↑}1.65\%) & 39.38 (\textcolor[rgb]{0,0.5,0}{↑}5.89\%) & 21.8 (\textcolor[rgb]{0,0.5,0}{↓}19.46\%) & 9.11 (\textcolor[rgb]{0,0.5,0}{↓}20.03\%) & 12.93 (\textcolor[rgb]{0,0.5,0}{↓}19.5\%)\\
& & \textbf{-/+Weighted\_BOLD} & 42.25 (\textcolor[rgb]{0,0.5,0}{↑}2.07\%) & 41.53 (\textcolor[rgb]{0,0.5,0}{↑}11.69\%) & 11.21 (\textcolor[rgb]{0,0.5,0}{↓}58.59\%) & 4.76 (\textcolor[rgb]{0,0.5,0}{↓}58.2\%) & 8.83 (\textcolor[rgb]{0,0.5,0}{↓}45.01\%)\\
\cline{2-8}
\multirow{4}{*}{} & Video-MME & \textbf{Default} & 32.05 & 28.15  & 20.62  & 12.52 & 13.75\\
& & \textbf{BOLD} & 32.65 (\textcolor[rgb]{0,0.5,0}{↑}1.89\%) & 29.13 (\textcolor[rgb]{0,0.5,0}{↑}3.49\%) & 19.29 (\textcolor[rgb]{0,0.5,0}{↓}6.44\%) & 12.03 (\textcolor[rgb]{0,0.5,0}{↓}3.86\%) & 13.09 (\textcolor[rgb]{0,0.5,0}{↓}4.82\%)\\
& & \textbf{+Weighted\_BOLD} & 32.2 (\textcolor[rgb]{0,0.5,0}{↑}0.47\%) & 28.97 (\textcolor[rgb]{0,0.5,0}{↑}2.94\%) & 17.65 (\textcolor[rgb]{0,0.5,0}{↓}14.38\%) & 11.7 (\textcolor[rgb]{0,0.5,0}{↓}6.54\%) & 12.57 (\textcolor[rgb]{0,0.5,0}{↓}8.56\%)\\
& & \textbf{-/+Weighted\_BOLD} & 32.92 (\textcolor[rgb]{0,0.5,0}{↑}2.72\%) & 31.32 (\textcolor[rgb]{0,0.5,0}{↑}11.27\%) & 12.6 (\textcolor[rgb]{0,0.5,0}{↓}38.89\%) & 7.85 (\textcolor[rgb]{0,0.5,0}{↓}37.25\%) & 10.36 (\textcolor[rgb]{0,0.5,0}{↓}24.67\%)\\
\cline{1-8}
\multirow{4}{*}{} & NExT-QA & \textbf{Default} & 49.95 & 49.81  & 16.31  & 3.27 & 5.4\\
& & \textbf{BOLD} & 51.81 (\textcolor[rgb]{0,0.5,0}{↑}3.72\%) & 51.71 (\textcolor[rgb]{0,0.5,0}{↑}3.82\%) & 13.27 (\textcolor[rgb]{0,0.5,0}{↓}18.63\%) & 2.67 (\textcolor[rgb]{0,0.5,0}{↓}18.42\%) & 4.58 (\textcolor[rgb]{0,0.5,0}{↓}15.18\%)\\
& & \textbf{+Weighted\_BOLD} & 51.83 (\textcolor[rgb]{0,0.5,0}{↑}3.76\%) & 51.77 (\textcolor[rgb]{0,0.5,0}{↑}3.93\%) & 12.79 (\textcolor[rgb]{0,0.5,0}{↓}21.58\%) & 2.5 (\textcolor[rgb]{0,0.5,0}{↓}23.66\%) & 4.43 (\textcolor[rgb]{0,0.5,0}{↓}17.81\%)\\
& & \textbf{-/+Weighted\_BOLD} & 52.94 (\textcolor[rgb]{0,0.5,0}{↑}5.98\%) & 53.0 (\textcolor[rgb]{0,0.5,0}{↑}6.41\%) & 8.49 (\textcolor[rgb]{0,0.5,0}{↓}47.93\%) & 1.98 (\textcolor[rgb]{0,0.5,0}{↓}39.48\%) & 3.02 (\textcolor[rgb]{0,0.5,0}{↓}43.96\%)\\
\cline{2-8}
\multirow{4}{*}{Video-LLaVA} & STAR & \textbf{Default} & 34.77 & 31.83  & 24.99  & 6.82 & 5.78\\
& & \textbf{BOLD} & 37.14 (\textcolor[rgb]{0,0.5,0}{↑}6.81\%) & 35.65 (\textcolor[rgb]{0,0.5,0}{↑}12.02\%) & 18.53 (\textcolor[rgb]{0,0.5,0}{↓}25.84\%) & 5.01 (\textcolor[rgb]{0,0.5,0}{↓}26.55\%) & 4.6 (\textcolor[rgb]{0,0.5,0}{↓}20.38\%)\\
& & \textbf{+Weighted\_BOLD} & 37.33 (\textcolor[rgb]{0,0.5,0}{↑}7.37\%) & 35.98 (\textcolor[rgb]{0,0.5,0}{↑}13.04\%) & 17.62 (\textcolor[rgb]{0,0.5,0}{↓}29.49\%) & 4.83 (\textcolor[rgb]{0,0.5,0}{↓}29.09\%) & 4.35 (\textcolor[rgb]{0,0.5,0}{↓}24.73\%)\\
& & \textbf{-/+Weighted\_BOLD} & 38.56 (\textcolor[rgb]{0,0.5,0}{↑}10.9\%) & 37.86 (\textcolor[rgb]{0,0.5,0}{↑}18.96\%) & 11.35 (\textcolor[rgb]{0,0.5,0}{↓}54.56\%) & 3.69 (\textcolor[rgb]{0,0.5,0}{↓}45.92\%) & 2.91 (\textcolor[rgb]{0,0.5,0}{↓}49.67\%)\\
\cline{2-8}
\multirow{4}{*}{} & Perception Test & \textbf{Default} & 40.72 & 35.69  & 27.5  & 14.75 & 5.28\\
& & \textbf{BOLD} & 41.62 (\textcolor[rgb]{0,0.5,0}{↑}2.22\%) & 38.69 (\textcolor[rgb]{0,0.5,0}{↑}8.4\%) & 21.75 (\textcolor[rgb]{0,0.5,0}{↓}20.9\%) & 10.92 (\textcolor[rgb]{0,0.5,0}{↓}25.97\%) & 3.79 (\textcolor[rgb]{0,0.5,0}{↓}28.14\%)\\
& & \textbf{+Weighted\_BOLD} & 41.95 (\textcolor[rgb]{0,0.5,0}{↑}3.02\%) & 39.46 (\textcolor[rgb]{0,0.5,0}{↑}10.58\%) & 19.73 (\textcolor[rgb]{0,0.5,0}{↓}28.26\%) & 10.24 (\textcolor[rgb]{0,0.5,0}{↓}30.53\%) & 3.41 (\textcolor[rgb]{0,0.5,0}{↓}35.37\%)\\
& & \textbf{-/+Weighted\_BOLD} & 41.57 (\textcolor[rgb]{0,0.5,0}{↑}2.09\%) & 41.11 (\textcolor[rgb]{0,0.5,0}{↑}15.2\%) & 8.95 (\textcolor[rgb]{0,0.5,0}{↓}67.47\%) & 4.3 (\textcolor[rgb]{0,0.5,0}{↓}70.81\%) & 2.11 (\textcolor[rgb]{0,0.5,0}{↓}59.94\%)\\
\cline{2-8}
\multirow{4}{*}{} & Video-MME & \textbf{Default} & 34.3 & 30.99  & 24.09  & 8.06 & 4.79\\
& & \textbf{BOLD} & 34.56 (\textcolor[rgb]{0,0.5,0}{↑}0.76\%) & 32.63 (\textcolor[rgb]{0,0.5,0}{↑}5.3\%) & 18.24 (\textcolor[rgb]{0,0.5,0}{↓}24.28\%) & 6.08 (\textcolor[rgb]{0,0.5,0}{↓}24.59\%) & 3.86 (\textcolor[rgb]{0,0.5,0}{↓}19.41\%)\\
& & \textbf{+Weighted\_BOLD} & 34.41 (\textcolor[rgb]{0,0.5,0}{↑}0.32\%) & 32.78 (\textcolor[rgb]{0,0.5,0}{↑}5.77\%) & 16.1 (\textcolor[rgb]{0,0.5,0}{↓}33.17\%) & 5.89 (\textcolor[rgb]{0,0.5,0}{↓}26.93\%) & 3.37 (\textcolor[rgb]{0,0.5,0}{↓}29.69\%)\\
& & \textbf{-/+Weighted\_BOLD} & 34.22 (\textcolor{red}{↓}0.22\%) & 33.56 (\textcolor[rgb]{0,0.5,0}{↑}8.31\%) & 9.8 (\textcolor[rgb]{0,0.5,0}{↓}59.31\%) & 3.71 (\textcolor[rgb]{0,0.5,0}{↓}54.0\%) & 1.84 (\textcolor[rgb]{0,0.5,0}{↓}61.61\%)\\
\cline{1-8}
\multirow{4}{*}{} & NExT-QA & \textbf{Default} & 63.91 & 63.88  & 2.15  & 1.32 & 1.99\\
& & \textbf{BOLD} & 63.91 (\textcolor[rgb]{0,0.5,0}{↑}0.0\%) & 63.88 (\textcolor[rgb]{0,0.5,0}{↑}0.01\%) & 2.03 (\textcolor[rgb]{0,0.5,0}{↓}5.45\%) & 1.18 (\textcolor[rgb]{0,0.5,0}{↓}10.77\%) & 1.98 (\textcolor[rgb]{0,0.5,0}{↓}0.66\%)\\
& & \textbf{+Weighted\_BOLD} & 64.0 (\textcolor[rgb]{0,0.5,0}{↑}0.15\%) & 63.98 (\textcolor[rgb]{0,0.5,0}{↑}0.15\%) & 1.95 (\textcolor[rgb]{0,0.5,0}{↓}9.2\%) & 1.24 (\textcolor[rgb]{0,0.5,0}{↓}6.37\%) & 1.97 (\textcolor[rgb]{0,0.5,0}{↓}0.96\%)\\
& & \textbf{-/+Weighted\_BOLD} & 64.07 (\textcolor[rgb]{0,0.5,0}{↑}0.26\%) & 64.06 (\textcolor[rgb]{0,0.5,0}{↑}0.29\%) & 0.84 (\textcolor[rgb]{0,0.5,0}{↓}60.74\%) & 1.07 (\textcolor[rgb]{0,0.5,0}{↓}18.78\%) & 2.08 (\textcolor{red}{↑}4.64\%)\\
\cline{2-8}
\multirow{4}{*}{SeViLA} & STAR & \textbf{Default} & 46.28 & 46.14  & 4.47  & 2.3 & 2.25\\
& & \textbf{BOLD} & 46.25 (\textcolor{red}{↓}0.06\%) & 46.14 (\textcolor{red}{↓}0.01\%) & 4.05 (\textcolor[rgb]{0,0.5,0}{↓}9.25\%) & 2.22 (\textcolor[rgb]{0,0.5,0}{↓}3.75\%) & 2.06 (\textcolor[rgb]{0,0.5,0}{↓}8.29\%)\\
& & \textbf{+Weighted\_BOLD} & 46.15 (\textcolor{red}{↓}0.27\%) & 46.04 (\textcolor{red}{↓}0.22\%) & 4.01 (\textcolor[rgb]{0,0.5,0}{↓}10.23\%) & 2.2 (\textcolor[rgb]{0,0.5,0}{↓}4.52\%) & 2.05 (\textcolor[rgb]{0,0.5,0}{↓}9.07\%)\\
& & \textbf{-/+Weighted\_BOLD} & 46.18 (\textcolor{red}{↓}0.21\%) & 46.1 (\textcolor{red}{↓}0.1\%) & 3.48 (\textcolor[rgb]{0,0.5,0}{↓}22.16\%) & 2.04 (\textcolor[rgb]{0,0.5,0}{↓}11.64\%) & 1.85 (\textcolor[rgb]{0,0.5,0}{↓}17.64\%)\\
\cline{2-8}
\multirow{4}{*}{} & Perception Test & \textbf{Default} & 45.3 & 45.11  & 6.21  & 3.02 & 1.59\\
& & \textbf{BOLD} & 45.4 (\textcolor[rgb]{0,0.5,0}{↑}0.23\%) & 45.25 (\textcolor[rgb]{0,0.5,0}{↑}0.33\%) & 5.24 (\textcolor[rgb]{0,0.5,0}{↓}15.53\%) & 3.04 (\textcolor{red}{↑}0.72\%) & 1.3 (\textcolor[rgb]{0,0.5,0}{↓}18.45\%)\\
& & \textbf{+Weighted\_BOLD} & 45.34 (\textcolor[rgb]{0,0.5,0}{↑}0.09\%) & 45.24 (\textcolor[rgb]{0,0.5,0}{↑}0.3\%) & 4.16 (\textcolor[rgb]{0,0.5,0}{↓}33.06\%) & 2.62 (\textcolor[rgb]{0,0.5,0}{↓}13.42\%) & 1.03 (\textcolor[rgb]{0,0.5,0}{↓}34.99\%)\\
& & \textbf{-/+Weighted\_BOLD} & 45.22 (\textcolor{red}{↓}0.17\%) & 45.14 (\textcolor[rgb]{0,0.5,0}{↑}0.08\%) & 3.85 (\textcolor[rgb]{0,0.5,0}{↓}38.0\%) & 2.43 (\textcolor[rgb]{0,0.5,0}{↓}19.68\%) & 0.9 (\textcolor[rgb]{0,0.5,0}{↓}43.13\%)\\
\cline{2-8}
\multirow{4}{*}{} & Video-MME & \textbf{Default} & 39.85 & 39.82  & 4.67  & 1.68 & 0.84\\
& & \textbf{BOLD} & 40.0 (\textcolor[rgb]{0,0.5,0}{↑}0.37\%) & 39.99 (\textcolor[rgb]{0,0.5,0}{↑}0.44\%) & 3.81 (\textcolor[rgb]{0,0.5,0}{↓}18.5\%) & 1.48 (\textcolor[rgb]{0,0.5,0}{↓}11.8\%) & 0.66 (\textcolor[rgb]{0,0.5,0}{↓}21.49\%)\\
& & \textbf{+Weighted\_BOLD} & 39.81 (\textcolor{red}{↓}0.09\%) & 39.81 (\textcolor{red}{↓}0.02\%) & 3.57 (\textcolor[rgb]{0,0.5,0}{↓}23.49\%) & 1.45 (\textcolor[rgb]{0,0.5,0}{↓}13.53\%) & 0.53 (\textcolor[rgb]{0,0.5,0}{↓}36.44\%)\\
& & \textbf{-/+Weighted\_BOLD} & 39.85 (\textcolor[rgb]{0,0.5,0}{↑}0.0\%) & 39.85 (\textcolor[rgb]{0,0.5,0}{↑}0.08\%) & 2.66 (\textcolor[rgb]{0,0.5,0}{↓}43.14\%) & 1.47 (\textcolor[rgb]{0,0.5,0}{↓}12.42\%) & 0.16 (\textcolor[rgb]{0,0.5,0}{↓}80.65\%)\\
\hline
\end{tabular}
}}
\label{tab:100}
\end{table*}

\end{document}